\PassOptionsToPackage{svgnames}{xcolor}

\documentclass[preprint,journal]{vgtc}            

\preprinttext{}

\vgtccategory{Research}

\vgtcpapertype{algorithm/technique}

\title{ENS-t-SNE: Embedding Neighborhoods Simultaneously t-SNE}

\author{%
  Jacob Miller, Vahan Huroyan, Raymundo Navarrete, Md Iqbal Hossain, Stephen Kobourov
}

\authorfooter{
    \item
  	Authors are at the University of Arizona, Department of Computer Science
    \item Jacob Miller is the corresponding author. Email: jacobmiller1@arizona.edu
}

\abstract{%
    When visualizing a high-dimensional dataset, dimension reduction techniques are commonly employed which provide a single 2 dimensional view of the data. 
    We describe ENS-t-SNE: an algorithm for Embedding Neighborhoods Simultaneously that generalizes the t-Stochastic Neighborhood Embedding 
     approach. By using different viewpoints in ENS-t-SNE's 3D embedding, one can visualize different types of clusters within the same high-dimensional dataset. This enables the viewer to see and keep track of the different types of clusters, which is harder to do  when providing multiple 2D embeddings, where corresponding points cannot be easily identified. We illustrate the utility of ENS-t-SNE with real-world applications and  provide an extensive quantitative evaluation with datasets of different types and sizes.
     
}

\keywords{Dimension Reduction, Joint Optimization, Simultaneous Embedding, t-SNE}

\teaser{
    \begin{centering}
    \includegraphics[width=0.32\linewidth]{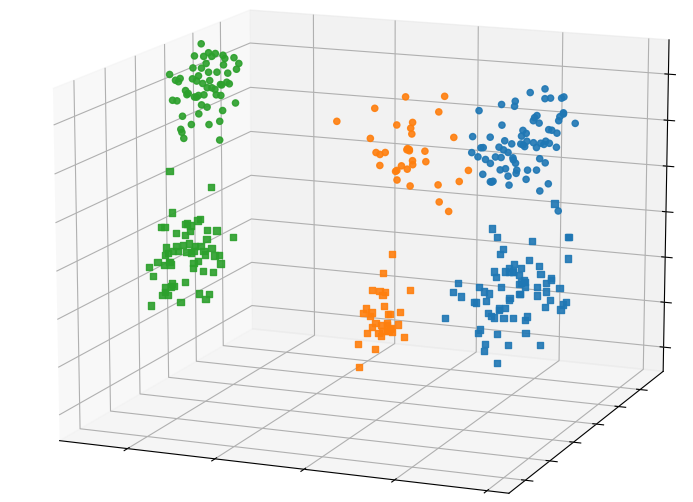}
    \includegraphics[width=0.32\linewidth, height=5cm]{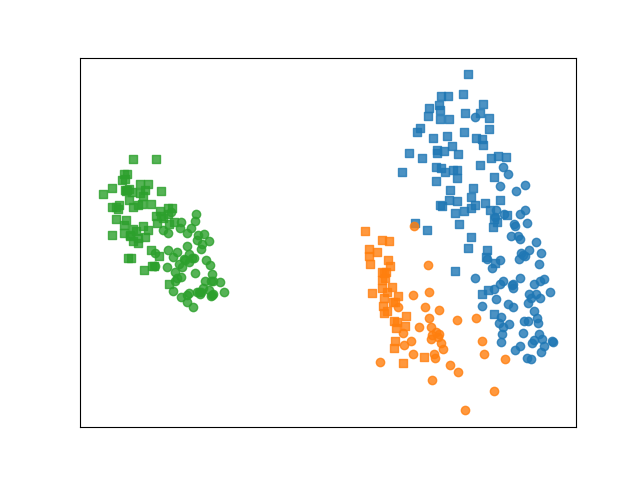}
    \includegraphics[width=0.32\linewidth, height=5cm]{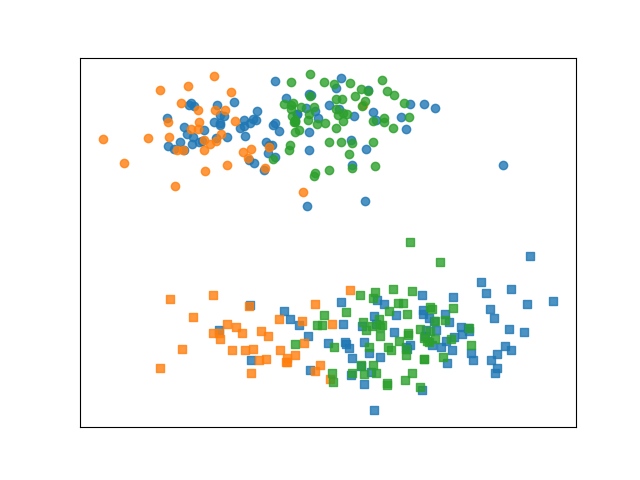}
    \end{centering}
    
    \parbox[c]{0.32\linewidth}{\centering (a)}
    \parbox[c]{0.32\linewidth}{\centering (b)}
    \parbox[c]{0.32\linewidth}{\centering (c)}
    
    \caption{The Palmer's Penguins dataset captured by ENS-t-SNE (perplexity=40). (a): The full 3D embedding. (b): The projection capturing physical characteristics, encoded by color (c): The embedding capturing penguin sex, encoded by shape. ENS-t-SNE captures all 6 distinct clusters in 3d, while finding suitable projections that show the two types of clusters. 
    }
    \label{fig:penguins_ens_t_sne}
}

\graphicspath{{figs/}{figures/}{pictures/}{images/}{./}} 

\usepackage{tabu}                      
\usepackage{booktabs}                  
\usepackage{lipsum}                    
\usepackage{mwe}                       

\usepackage{mathptmx}                  

\usepackage{amsmath}
\usepackage{amsfonts}
\usepackage{algpseudocode}
\usepackage{algorithm}
\usepackage{todonotes}

\DeclareMathOperator{\vx}{\mathbf{x}}
\DeclareMathOperator{\vy}{\mathbf{y}}

\DeclareMathOperator{\mD}{\mathbf{D}}

\DeclareMathOperator{\mX}{\mathbf{X}}
\DeclareMathOperator{\mY}{\mathbf{Y}}

\usepackage[T1]{fontenc}

\usepackage{cite}  

\usepackage{caption}
\usepackage{subcaption}

\begin{document}


\firstsection{Introduction}

\maketitle

Dimension reduction has become one of the primary techniques to visualize high dimensional data. 
These techniques place data points from a high dimensional space into the (typically) two-dimensional plane of the computer screen while maintaining some characteristics of the dataset. Well known algorithms include PCA~\cite{jolliffe1986principal}, t-SNE~\cite{maaten2008visualizing} and UMAP~\cite{mcionnes2018umap}.  
These algorithms assume either a high dimensional dataset as an input or a distance matrix between its instances.
However, many such datasets contain complex phenomena which cannot be captured in a single, two dimensional, static view.
Often there are several subspaces of interest to compare, so small multiple plots are employed~\cite{DBLP:journals/tvcg/LiuMWBP17, DBLP:conf/ieeevast/TatuMFBSSK12}. Even with coordinated views, it is difficult to track where groups of points go from one projection to the next, in effect offloading the mental effort of comparison to the user~\cite{DBLP:journals/ivs/GleicherAWJHR11}. 
We present a technique to capture multiple subspaces of interest in a single 3D embedding. 
We also utilize the 2D linear projections from the computed 3D embedding, to visualize specific cluster relationships within the corresponding subspaces. Finally, we provide a seamless transition between the subspace views, which reduces the movement of the points from one view to the next, as the views are obtained by simple rotation of the 3D embedding.

\subsection{Motivation}

Our proposed algorithm, named ENS-t-SNE, facilitates a challenging task within datasets: understanding the differences among individual data points, groups of points, or the dataset as a whole.
In the typology of  Brehmer and Munzner ~\cite{DBLP:journals/tvcg/BrehmerM13}, this is a ``compare" task. There are two visualization comparison designs applicable to subspaces, as described by Gleicher et al.~\cite{DBLP:journals/ivs/GleicherAWJHR11}: \textit{juxtaposition} and \textit{superposition}. Juxtaposition places points separately, as in a small multiple plot, but requires scanning to detect differences and similarities. 
Superposition uses the same space to show two or more subspaces, resulting in more complex stimulus, which is closer to a direct comparison. ENS-t-SNE allows us access to both these comparison designs. The 3D embedding produced by ENS-t-SNE can be seen as a superposition, by encoding each subspace from a different view of the object. 
ENS-t-SNE can provide juxtaposition with small multiple plots corresponding to projections for each subspace, replacing standard independent projections of subspaces; see Fig.~\ref{fig:penguins_ens_t_sne}.

Real-world multi-perspective embeddings can be found in art, e.g.,  the ``1, 2, 3'' sculpture by J.~Hopkins which looks like the different numbers depending on the viewpoint\footnote{\url{https://www.jameshopkinsworks.com/commissions4.html}} and ``Squaring the circle'' by Troika\footnote{\url{https://trendland.com/troika-squaring-the-circle/}}.
Motivated by such 3D physicalizations, as well as by work on Multi-Perspective Simultaneous Embedding~\cite{hossain2021multi}, our proposed algorithm enables viewers to virtually ``walk around'' an ENS-t-SNE embedding and see different aspects of the same dataset. The concept of data physicalization (data exists in a physical space) has been shown to be effective for information retrieval when the visualization can be realized~\cite{DBLP:conf/chi/JansenDF13,DBLP:conf/chi/JansenDIAKKSH15}.

We present a technique to capture multiple subspaces of interest in a single 3D embedding. We also utilize the 2D linear projections from the computed 3D embedding, to visualize specific cluster relationships within the corresponding subspaces. Finally, we provide a seamless
transition between the subspace views, which reduces the movement of the points from one view to the next, as the views are obtained by simple rotation of the 3D embedding.

\subsection{Our Contributions}
We propose, describe and provide an implementation for ENS-t-SNE: a technique to perform dimension reduction on a high-dimensional dataset that captures multiple subspaces of interest in a single 3D embedding.
Unlike the only prior work in this domain, which optimized global distance preservation (distances between all pairs of points)~\cite{hossain2021multi}, we focus on preserving local relationships (clusters). 
We also visualize specific cluster relationships within the corresponding subspaces, with the help of 2D linear projections from the computed 3D embedding. Finally, we provide a seamless
transition between the subspace views, which reduces the movement of the points from one view to the next, as the views are obtained by simple rotation of the 3D embedding; see Fig.~\ref{fig:teaser}.
This is accomplished by generalizing the t-SNE algorithm~\cite{maaten2008visualizing} to simultaneously optimize multiple distance matrices, while also simultaneously optimizing projection views. 
In addition to the implementation, we provide a publicly available web demonstration~\footnote{\url{https://jacoblmiller.github.io/ENS-t-SNE-web/}}. 
We remark that, unlike t-SNE, where the relative positions of clusters are usually meaningless, ENS-t-SNE does better in preserving these positions. It embeds clusters in 3D in a manner that ensures the corresponding 2D projections contain the corresponding clusters, compelling the 3D clusters to have meaningful positions. We experimentally verify this and observe it visually for several datasets.

\section{Related Work}
\label{sec:prev_work}

The related work section is organized as follows. First, we review several dimensionality reduction algorithms that are widely used in visualization. Second, we delve into the fundamentals of t-SNE, to provide the needed background information needed for its generalization, ENS-t-SNE. Third, we review algorithms for subspace clustering, a domain that shares a common goal with ENS-t-SNE: finding multiple embeddings, each capturing a distinct aspect of the data. Fourth, we consider related prior approaches to simultaneous or multi-view embeddings. Finally, 
we review Multi-Perspective Simultaneous Embedding (MPSE), an algorithm that preserves global distances.

\smallskip\noindent \textbf{Dimension Reduction.}
A wide variety of dimension reduction techniques abound:  Principal Component Analysis (PCA)~\cite{jolliffe1986principal}, Multi-Dimensional Scaling (MDS)~\cite{shepard1962analysis}, Laplacian Eigenmaps~\cite{belkin2001advances}, t-Distributed Stochastic Neighbor Embedding (t-SNE)~\cite{maaten2008visualizing}, Uniform Manifold Approximation and Projection (UMAP)~\cite{mcionnes2018umap}. These techniques attempt to capture the variance in the data (PCA), the global distances in the data (MDS), the local distances in the data (t-SNE), manifolds in the data (Laplacian Eigenmaps, UMAP).
For a survey of other single projection dimension reduction methods, see~\cite{DBLP:journals/tvcg/EspadotoMKHT21}. Note that none of these allows the viewer the opportunity to compare and contrast different aspects (subspaces).

\smallskip\noindent \textbf{T-Distributed Stochastic Neighbor Embedding (t-SNE)}
~\cite{maaten2008visualizing, wattenberg2016use}. 
creates a low dimensional embedding from a high dimensional dataset, based on the short distances between points in the data. Unlike stress based methods such as MDS, t-SNE converts these distances into a probability distribution which tells us the likelihood that two data are `neighbors' and should appear near each other. 
The mathematical formulation of the problem is the following: Given an $N \times N$ distance matrix $\mD$, 
and a perplexity parameter, $\sigma$, t-SNE seeks to minimize the following cost function:
\begin{equation}
\label{eq:t_sne_objective}
    C(Y) = \sum_{i,j} p_{i j} \log \frac{p_{i j}}{q_{i j}}
\end{equation}
    Here, $P = [p_{i j}]$ is determined by $D$ and the perplexity parameter, where $p_{ij} = \frac{p_{j | i} + p_{i | j}}{2N}$ and
    \begin{equation}
    p_{j | i} = \frac{e^{-(\mD_{ij})^2/2\sigma_i^2}}{\sum_{k \neq i} e^{-(\mD_{ik})^2/2\sigma_i^2}}.
\end{equation}
    Furthermore, $Q=[q_{i j}]$ is a function of $Y$ and is defined as
\begin{equation}
    q_{i j} = \frac{\left( 1 + \Vert (\vy_i - \vy_j)\Vert^2 \right)^{-1}}{\sum_{k \neq l} \left( 1 + \Vert (\vy_k - \vy_l) \Vert^2 \right)^{-1}}.
\end{equation}
The computational complexity of t-SNE is high and speed improvements have been proposed~\cite{maaten2014accelerating}.
Although the original paper proposes default values and ranges for the t-SNE hyperparameters (perplexity, learning rate, etc.), automatically selecting these parameters is also a topic of interest~\cite{cao2017automatic,belkina2019automated}.
recent paper reviews t-SNE and applications thereof~\cite{ghojogh2020stochastic}. 
In our ENS-t-SNE algorithm we optimize a generalization of the cost function of t-SNE (eq.~\ref{eq:t_sne_objective}) per projection, as detailed in section~\ref{sec:ens_t_sne_description}.

\smallskip\noindent \textbf{Subspace Clustering}.
A single projection or perspective may not be sufficient to understand diverse patterns in high-dimensional data. The goal of subspace clustering is to find multiple embeddings, each capturing a different aspect of the data~\cite{DBLP:journals/tvcg/LiuMWBP17}. Indeed, two different sets of dimensions may hold different, or even conflicting, patterns. 
The Pattern Trails tool orders axis-aligned subspaces, plotting them as a series of 2D embeddings and overlaying parallel coordinates to track the overall pattern of data position changes~\cite{jackle2017pattern}. 
Tatu et al.~\cite{DBLP:conf/ieeevast/TatuMFBSSK12} use the SURFING~\cite{DBLP:conf/icdm/BaumgartnerPKKK04} algorithm to prune away uninteresting subspaces, while the interesting subspaces are  embedded with MDS and incorporated into a visual analytics tool for further filtering and exploration. 
Fujiwara et al. provide a feature learning method and visualization tool to explore non-axis aligned subspaces using a series of UMAP projections to embed the data~\cite{DBLP:journals/corr/abs-2206-13891}. For more detail, see the subspace clustering section from the recent survey by Liu et al.~\cite{DBLP:journals/tvcg/LiuMWBP17}.

Combining subspace clustering techniques with ENS-t-SNE seems like a promising idea.  First, subspace clustering algorithms are the natural way to select interesting subspaces as input into ENS-t-SNE, especially for truly large and high-dimensional datasets where domain knowledge and expertise might not be enough. Second, ENS-t-SNE offers a powerful tool to perform comparison tasks on interesting subspaces, something that is typically done with small multiple plots (which do not support the full range of comparison tasks)~\cite{DBLP:journals/ivs/GleicherAWJHR11}. 
We illustrate this by example in Section~\ref{sec:food-comp}, with a dataset used in several subspace clustering papers~\cite{DBLP:conf/ieeevast/TatuMFBSSK12,yuan2012dimension}.

\smallskip\noindent \textbf{Simultaneous Embedding}. Some recent algorithms for simultaneous embedding/multiview embedding include 
Multiview Stochastic Neighbor Embedding (m-SNE)~\cite{bo2010msne, xie2011msne}, based on a probabilistic framework that integrates heterogeneous features of the dataset into one combined embedding and Multiview Spectral Embedding (MSE)~\cite{xia2010multiview}, which encodes features in such a way to achieve a physically meaningful embedding.
Multi-view Data Visualization via Manifold Learning~\cite{rodosthenous2021multi}, proposes extensions of t-SNE, LLE and ISOMAP, for dimensionality reduction and visualization of multiview data by computing and summing together the gradient descent for each data-view. 
Multi-view clustering for multi-omics data using unified embedding~\cite{mitra2020multi} uses the sum of the Kullback-Leibler divergence over the datapoints, which leads to a simple gradient adjusting the position of the samples in the embedded space.

The approaches above seek to achieve one of two objectives: either generating an embedding that encompasses multiple subspaces~\cite{bo2010msne,xie2011msne,xia2010multiview}, thereby potentially blending the information and lacking a guarantee of preserving individual subspaces of interest, or solely creating 2D projections where each corresponds to a distinct subspace~\cite{rodosthenous2021multi,mitra2020multi}. In the latter case, however, the challenge lies in establishing clear correspondences between these views.
The advantage of ENS-t-SNE is that it first creates a 3D embedding, where all the information is encoded and specific projections contain more fine-grained information based on subspaces of interest.
Thus, there are smooth transitions between the subspace views; consequently, ENS-t-SNE ensures coherence across all 2D perspectives, fostering a more nuanced and comprehensive representation of the underlying data.

\smallskip\noindent \textbf{Multi-Perspective Simultaneous Embedding (MPSE)}. 
Motivated by multi-view MDS, introduced in~\cite{bai2017multidimensional, kanaan2018multiview},
MPSE~\cite{hossain2021multi} computes a 3D embedding and set of projections from the 3D embedding much like ENS-t-SNE. 
The MPSE algorithm can be seen as a generalization of MDS that is able to visualize multiple distance matrices simultaneously, by producing a three-dimensional embedding, so that the different distance matrices are preserved after projecting the 3D coordinates to 2D ones using specified projections. Given a set of $M$ distance matrices of size $N\times N$:  $\mD^1,\mD^2,\dots,\mD^M$, MPSE aims to find 3D coordinates $\vx_1, \vx_2, \dots, \vx_N$ for the samples $i = 1, \dots, N$ and a set of 3D to 2D projection matrices $\Pi^1,\Pi^2,\dots,\Pi^M$ so that
\begin{equation}
    \sigma^2_M(\mX,\Pi) = \sum_{m=1}^M \sigma^2(\Pi^m \mX)
\end{equation}
where $\sigma^2$ is the MDS stress function.  
MPSE aims to preserve global distances: all pairwise distances in the input high-dimensional space must match all pairwise distances in low-dimensional space. In contrast, ENS-t-SNE aims to preserve in low dimensional space only local neighborhoods.


\section{Embedding Neighborhoods Simultaneously}
\label{sec:ens_t_sne_description}
Our proposed ENS-t-SNE algorithm is a generalization of the standard t-SNE algorithm. 
For ENS-t-SNE, we assume a set of distance matrices for the same set of objects. 
Similar to Multi-Perspective Simultaneous Embedding~\cite{hossain2021multi} with MDS, we generalize t-SNE. 
For this purpose we generalize the objective function of t-SNE onto the one that would take multiple distance matrices and have one projection for each on which the desired distances would locally be preserved.
We generalize the objective function of t-SNE as follows: Assume we are given $M$ distance matrices between n objects. We define
\begin{equation}
\label{eq:ens_t_sne_objectivs}
    \widetilde{C} = \sum_{m = 1}^M \sum_{i} \sum_{j} p^{m}_{ij} \log \frac{p^{m}_{ij}}{q^{m}_{ij}}
\end{equation}
where $ p^{m}_{j | i}$ and $q_{ij}^m$ for $1 \le i, j \le N$ and $1 \le m \le M$ are defined as

\begin{equation}
    p^{m}_{j | i} = \frac{e^{-(D_{ij}^m)^2/2(\sigma_i^m)^2}}{\sum_{k \neq i} e^{-(D_{ik}^m)^2/2(\sigma_i^m)^2}}, \text{ and }p_{ij}^m
    = \frac {p^{m}_{j | i} + p^{m}_{i | j}}{2N},
\end{equation}

and
\begin{equation}
    q_{ij}^m = \frac{\left( 1 + \Vert \Pi^{m} (\vy_i - \vy_j)\Vert^2 \right)^{-1}}{\sum_{k \neq l} \left( 1 + \Vert \Pi^{m} (\vy_k - \vy_l) \Vert^2 \right)^{-1}}.
\end{equation}
Here $D_{ij}^m$ for $1 \le i, j \le n$ corresponds to the $m$-th distance matrix between objects $i$ and $j$, $\Pi^m$ corresponds to the $m$-th projection, which depending on the problem might be given or not, and $y_1, \dots, y_N$ correspond to the desired embedding. 
The objective function $\widetilde{C}$ is a function of the embedding $\vy_1, \dots, \vy_N $ and projections $\Pi^1, \dots, \Pi^M$. The goal is to find $\vy_1, \dots, \vy_N $ and $\Pi^1, \dots, \Pi^M$ that would minimize $\widetilde{C}$.
Here, the $\widetilde{C}$ is a non-convex function. 
The only parameter that we have not discussed yet are the $\sigma_i^m$. These parameters are application specific and can vary depending on the dataset. Depending on the density of the dataset around each high dimensional point, the values of $\sigma_i^m$ can vary. The denser the dataset is, the smaller $\sigma_i^m$ can be chosen to be.

In order to pick appropriate values for $\sigma_i^m$ for all $1 \le i \le N$ and $1 \le m \le M$, we follow the steps of~\cite{maaten2008visualizing} and use the notion of perplexity. 
Perplexity can be interpreted as roughly the effective number of neighbors we want to capture around each point.
Each value of $\sigma_i^m$ creates a new distribution $P_i^m$ over the datapoints. 
Similar to t-SNE we perform a binary search to compute the value of $\sigma_i^m$ that produces a distribution $P_i^m$ with a fixed perplexity that is specified by the user. The perplexity is defined as
\begin{equation}
    \text{Perp}(P_i^m) = 2^{H(P_i^m)},
\end{equation}
where $H(P_i^m)$ is the Shannon entropy function, defined as 
\begin{equation}
    H(P_i^m) = - \sum_{j} p_{j|i}^m \log_2 (p_{j|i}^m).
\end{equation}
In order to optimize the objective function of ENS-T-SNE~\eqref{eq:ens_t_sne_objectivs} we use stochastic gradient descent discussed in Sec.~\ref{sec:ens_t_sne_algorithm}. 
We would like to mention that one can compute the exact gradients of $\widetilde{C}$.

There are two natural variants of this problem. The first version assumes that we are given a set of projections $\Pi^1, \dots, \Pi^M$ and a set of distance matrices $\mD^1, \dots, \mD^M \in \mathbb{R}^{N \times N}$  between $N$ objects, where each projection matrix corresponds to one distance matrix. 
The goal is to find an embedding of the dataset in $3D$ $Y = \{ \vy_1, \dots, \vy_N \}$such that on each of these projections the corresponding distances are locally preserved. 
The second version assumes only a set of distance matrices $\mD^1, \dots, \mD^M \in \mathbb{R}^{N \times N}$ and finds the best embedding $\mY = \{ \vy_1, \dots, \vy_N \}$ as well as the projections $\Pi^1, \dots, \Pi^M$ onto which the local distances would be preserved.

\subsection{ENS-T-SNE Algorithm}
\label{sec:ens_t_sne_algorithm}

In this section we summarize the ENS-T-SNE algorithm and provide a practical implementation.
Given a list of $N \times N$ distance matrices $\mD^1, \mD^2,\dots, \mD^M$ and a perplexity parameter $\mathrm{Perp}$, ENS-T-SNE algorithm aims to find a three-dimensional embedding $\vy_1, \vy_2, \dots, \vy_N \in \mathbb{R}^3$ and a set of projection matrices $\Pi^1,\Pi^2,\dots,\Pi^M \in \mathbb{R}^{2\times 3}$ so that for each perspective $m = 1, 2, \dots, M$, the corresponding projected dataset $\Pi^m \vy_1, \Pi^m \vy_2, \dots, \Pi^m \vy_M$ minimizes the t-SNE cost function $C(\mY^m)$ on that particular 2D Euclidean space as much as possible.

\begin{figure*}
    \centering
    \includegraphics[width=0.24\linewidth]{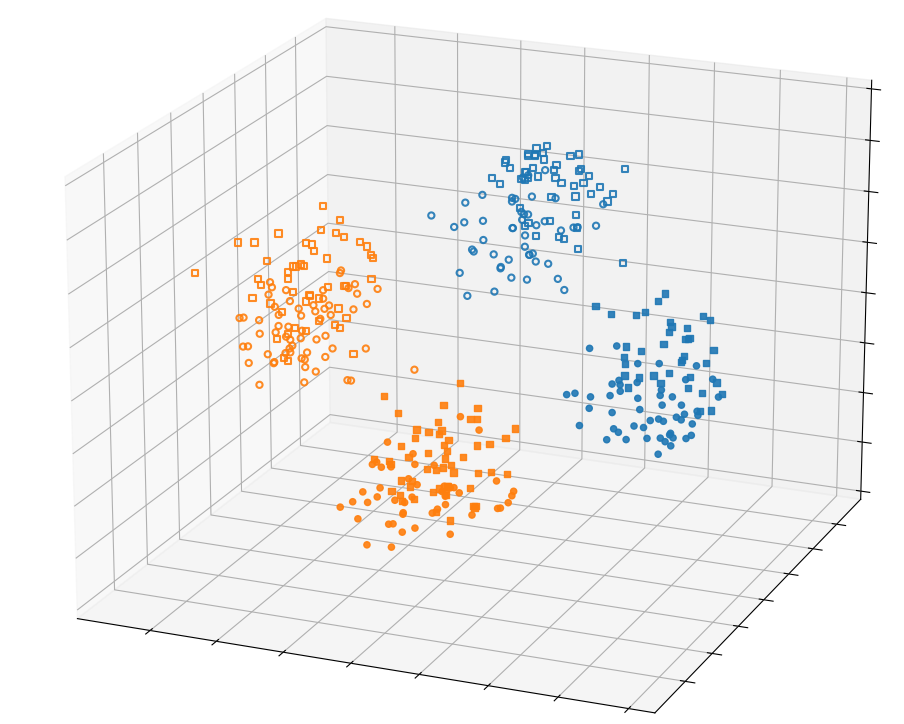}
  \includegraphics[width=0.24\linewidth]{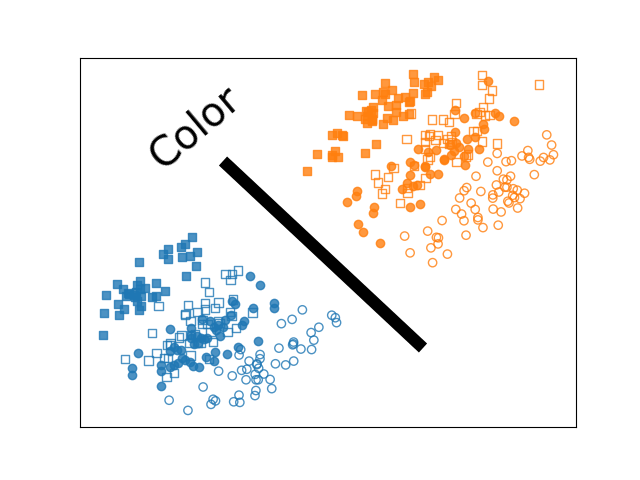}
  \includegraphics[width=0.24\linewidth]{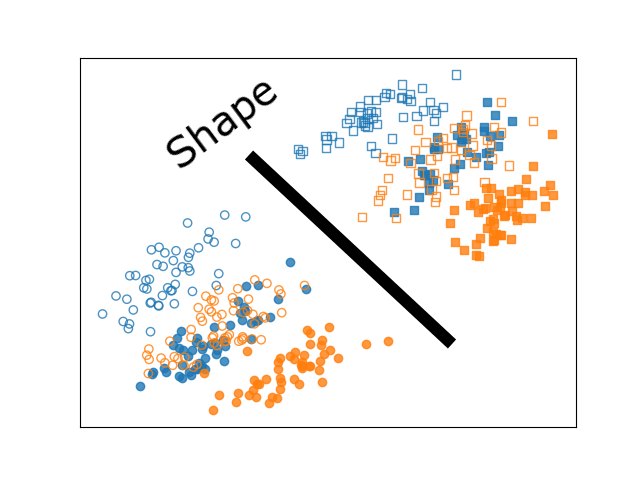}
  \includegraphics[width=0.24\linewidth]{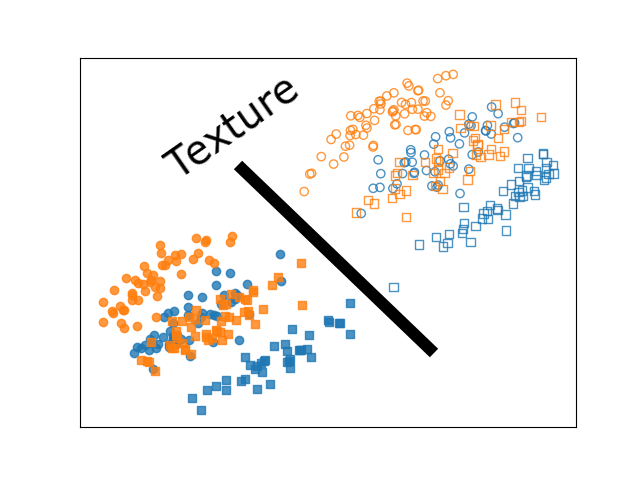}  

  \parbox[c]{0.24\linewidth}{\centering (a)}
  \parbox[c]{0.24\linewidth}{\centering (b)}
  \parbox[c]{0.24\linewidth}{\centering (c)}
  \parbox[c]{0.24\linewidth}{\centering (d)}

  \caption{The ENS-t-SNE embedding of a $400$-point dataset with three perspectives: in each perspective there are two different clusters. We have encoded each perspective's clusters using visual channels: color (orange or blue), shape (square or circle), and texture (filled or not filled). 
    (a) shows the three dimensional embedding of the dataset. (b) shows the first view where data points are clustered by color, (c) shows the second view where points are clustered by shape, and (d) where points are clustered by texture. 
  }
  \label{fig:teaser}
\end{figure*}

We write $\mY = [\vy_1, \vy_2, \dots, \vy_N]$ and $\Pi=[\Pi^1, \Pi^2,\dots,\Pi^M]$. Similar to t-SNE, we accomplish this using a gradient descent type algorithm.
The ENS-T-SNE cost function \eqref{eq:ens_t_sne_objectivs} is defined as the sum of the t-SNE cost function evaluated at each of the 2D projections $\Pi^1 \mY, \Pi^2 \mY, \dots, \Pi^M \mY$.
\begin{equation}
    \widetilde{C}(\mY,\Pi)  = \sum_{m=1}^M C(\Pi^m \mY) = \sum_{m = 1}^M \sum_{i<j} p^{m}_{ij} \log \frac{p^{m}_{ij}}{q^{m}_{ij}},
\end{equation}
where $C$ is the t-sne cost function. The gradients of $\widetilde{C}$ with respect to $Y$ and $\Pi^1,\Pi^2,\dots,\Pi^M$ are then
\begin{equation}
    \nabla_{\mY} \widetilde{C} (\mY, \Pi^1, \Pi^2, \dots, \Pi^M) = \sum_{m=1}^M (\Pi^m)^T \nabla C(\Pi^m \mY) 
\end{equation}
and
\begin{equation}
    \nabla_{\Pi^m} \widetilde{C} (\mY, \Pi^1, \Pi^2, \dots, \Pi^M) = \nabla C(\Pi^m \mY) \mY^T.
\end{equation}
As derived in~\cite{maaten2008visualizing}, the gradient $\nabla C = \left[\frac{\partial C}{\partial \vy_1}, \frac{\partial C}{\partial \vy_2}, \dots, \frac{\partial C}{\partial \vy_N} \right] $ is given by
$$ \frac{\partial C}{\partial \vy_i} = 4 \sum_j (p_{ij}-q_{ij})(\vy_i - \vy_j). $$
In its simplest form (full gradient and projected gradient descent), the update rules are
$$ \mY \mapsto \mY + \mu  \nabla_{\mY} \widetilde{C} (\mY, \Pi^1, \Pi^2, \dots, \Pi^M)$$
and 
$$ \Pi^m \mapsto Q \left( \nu \nabla_{\Pi^m} \widetilde{C} (\mY, \Pi^1, \Pi^2, \dots, \Pi^M) \right),$$
where $\mu, \nu > 0$ are learning rates and $Q$ maps a $2\times 3$ matrix to its nearest orthogonal matrix.

In practice, we found that a combination of adaptive learning rate and stochastic gradient descent works the best in consistently avoiding local minima. 
To avoid flat solutions, we first optimize for the embedding $Y$ while keeping the projections fixed (which are randomly chosen among the set of 2x3 orthogonal matrices). 
This algorithm is described in supplemental material.

\begin{figure*}[t]
    \begin{centering}
    \includegraphics[height=4cm,width=0.24\linewidth]{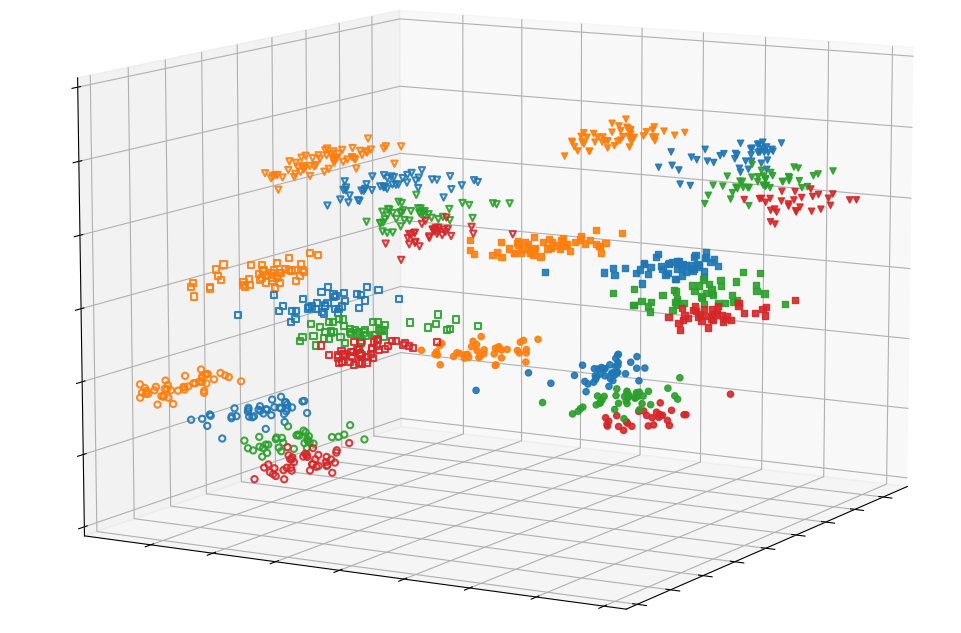}
    \includegraphics[height=4cm,width=0.24\linewidth]{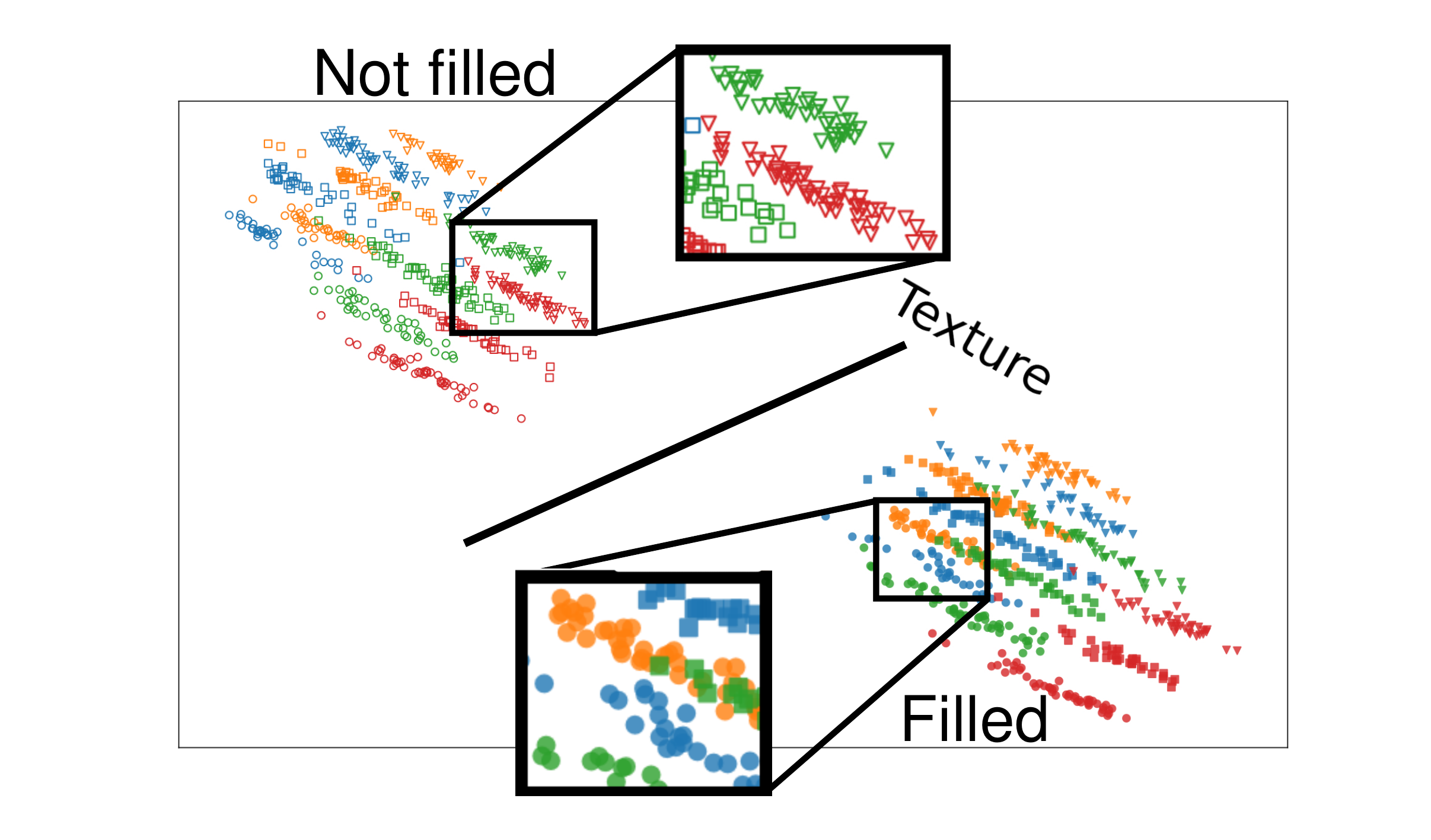}
    \includegraphics[height=4cm,width=0.24\linewidth]{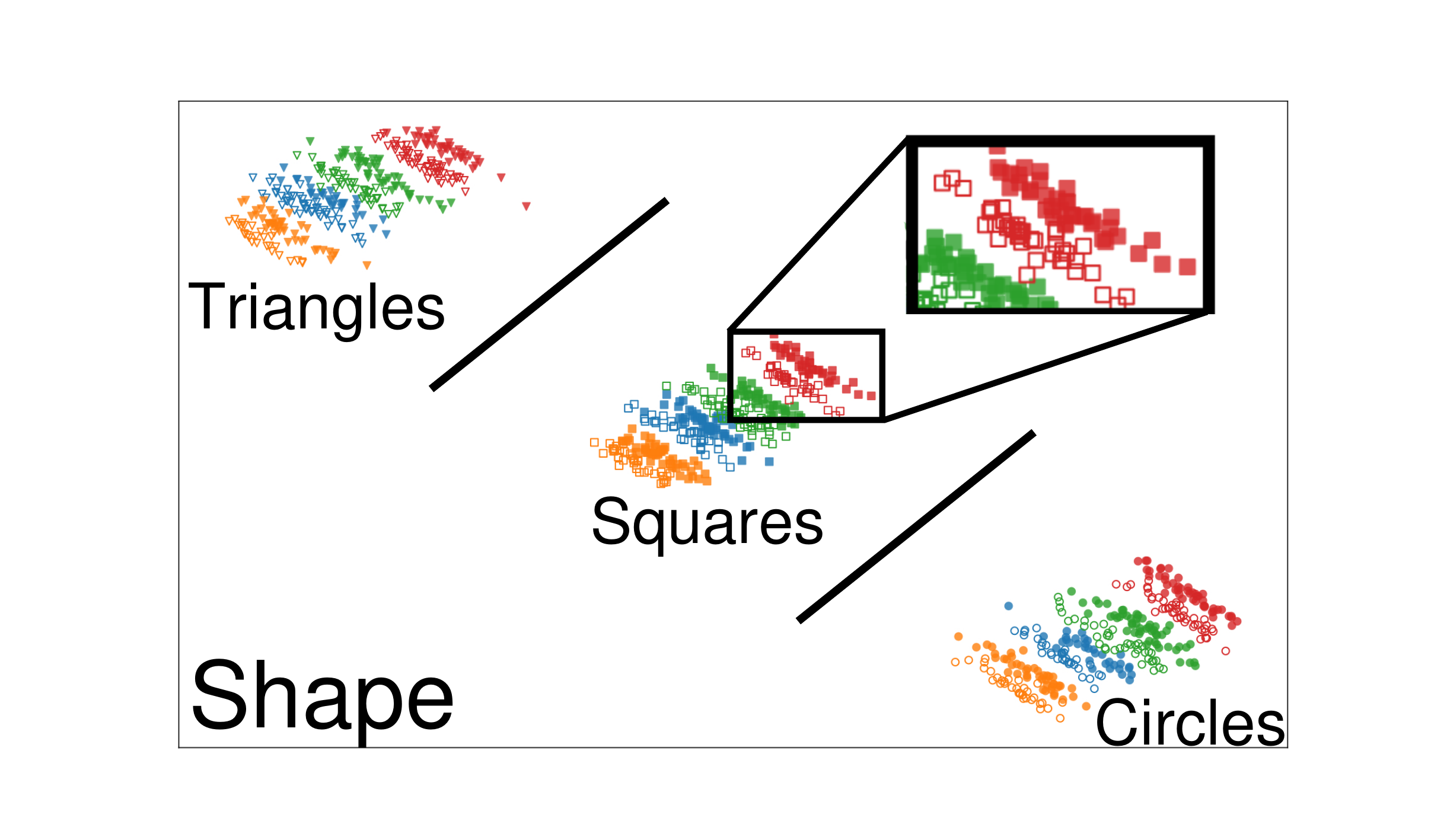}
    \includegraphics[height=4cm,width=0.24\linewidth]{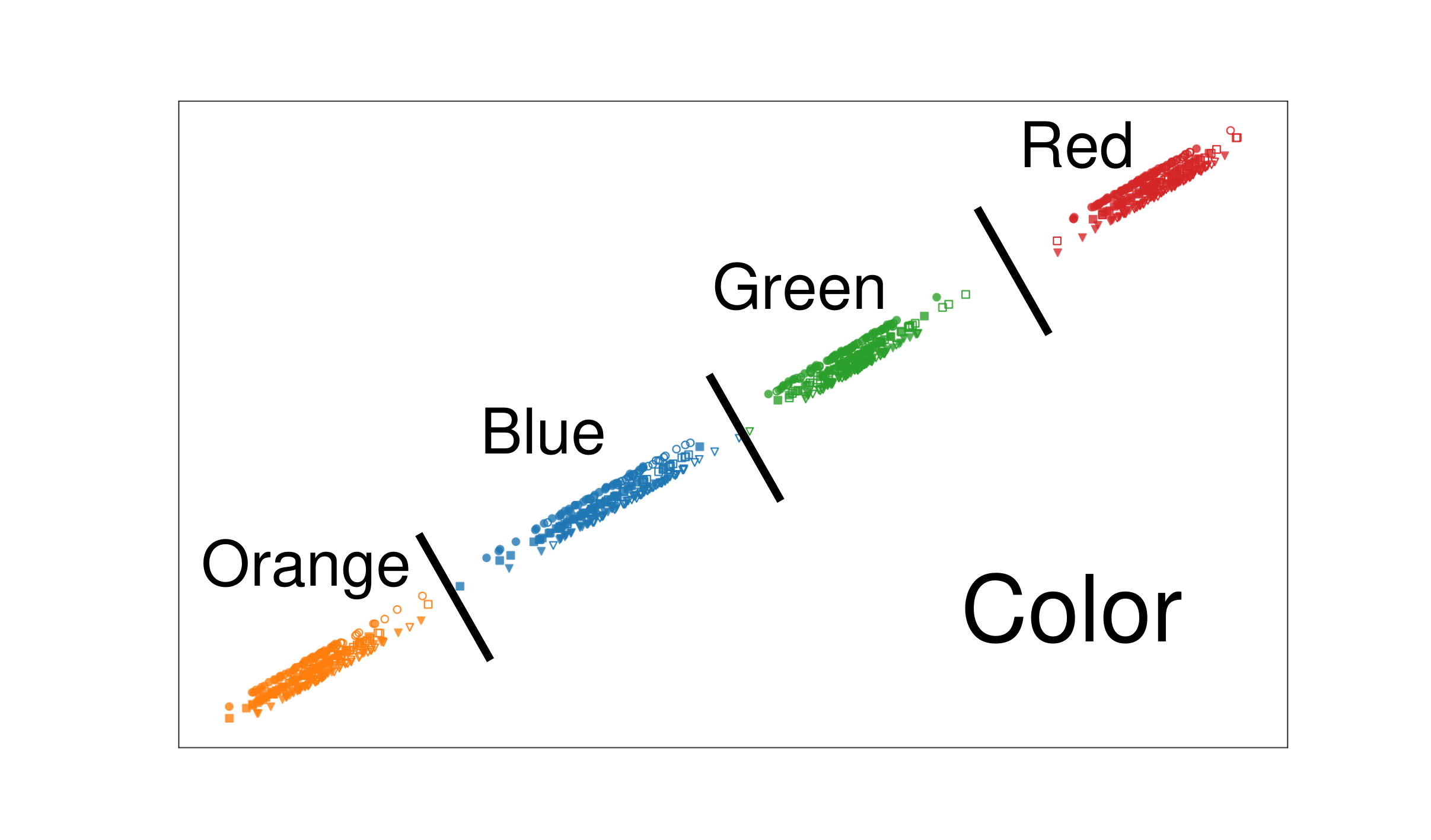}
    \end{centering}

    \parbox[c]{0.24\linewidth}{\centering (a)}
    \parbox[c]{0.24\linewidth}{\centering (b)}
    \parbox[c]{0.24\linewidth}{\centering (c)}
    \parbox[c]{0.24\linewidth}{\centering (d)}    
  
    \caption{ENS-t-SNE embedding of a clustered dataset, created according to Section~\ref{sec:cluster_construction} where the number of perspectives is $M = 3$, 
    the number of datapoints is $N = 1000$,
    and the number of clusters per perspective is $NC_1 = 2$, $NC_2 = 3$ and $NC_3 = 4$.
    Fig.~\ref{fig:clustering_234_400}(a) shows a snapshot of the 3D ENS-t-SNE embedding and Figures~\ref{fig:clustering_234_400}(b)-(d) show the 2D projections of the 3D ENS-t-SNE embedding. The original clusters are shown in texture (filled or not), shape (square, triangle and circle), and color (blue, orange, green, red). ENS-t-SNE  is able to recover the clusters and create an embedding which respects all the different types of clusters.
    }
    \label{fig:clustering_234_400}
\end{figure*}

\begin{figure*}[t]
    \centering 
    \includegraphics[width=0.23\linewidth]{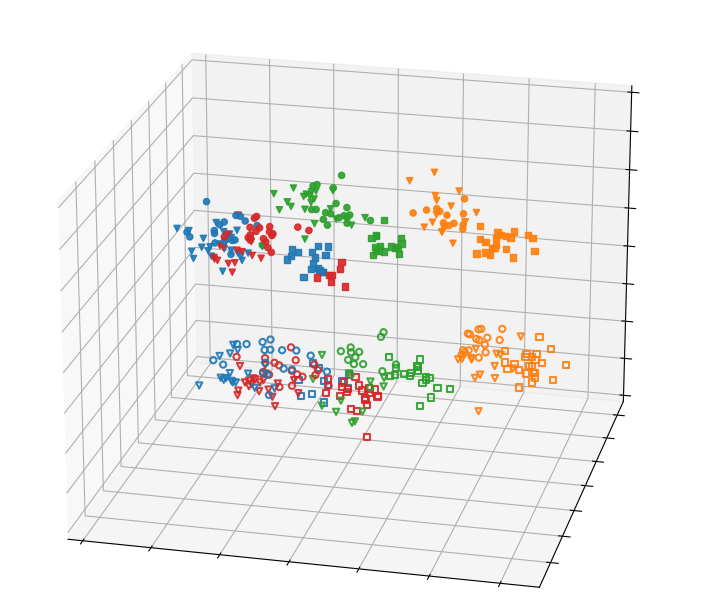}
    \raisebox{5cm}{\includegraphics[width=0.69\linewidth,height=5cm,angle=180]{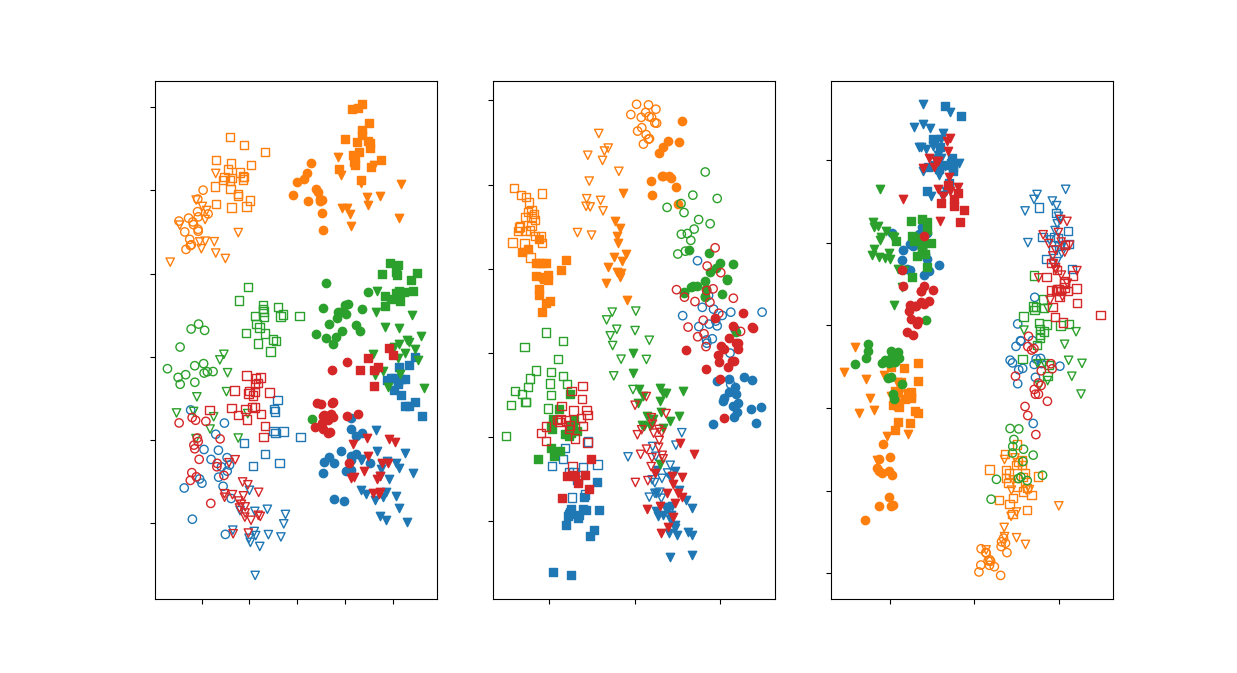}}
    
    \caption{MPSE applied to the synthetic data from Section~\ref{sec:cluster_construction}. Recall that we should see clusters based on texture, shape, and color in the three views. MPSE fails in capturing this information by mixing clusters in the color and shape views. In general, these clusters are better separated in Fig~\ref{fig:clustering_234_400}.}
    \label{fig:mpse-432}
\end{figure*}

\textbf{Initialization: }
\label{sec:smart-init}
Similar to t-SNE, the objective function of ENS-t-SNE in \eqref{eq:ens_t_sne_objectivs} is non-convex. 
Thus,  gradient-based methods are not guaranteed to find global optima; e.g., 
it is known that t-SNE is vulnerable to being caught in local minima when randomly initialized~\cite{kobak2021initialization}. 
Further, different ENS-t-SNE runs with random initialization, produce different embeddings. 
A good initialization for ENS-t-SNE can help reach better solutions and provide  deterministic results.
In the case of standard t-SNE, a `smart initilization' based on a 2D PCA projection tends to be better than random initialization.  
This addresses both the poor local minima problem and the non-determinism.

Since we are given multiple pairwise dissimilarity matrices in ENS-t-SNE,
we cannot simply use a single PCA projection
for initialization. 
Therefore, we devise a `smart initialization' strategy by first taking the average over all 
pairwise dissimilarity matrices and 
then applying dimensionality reduction to 3D via 
classical (Torgerson) MDS~\cite{Torgerson1952MultidimensionalSI}.
When using this initialization scheme we also deterministically compute a set of 2x3 orthogonal matrices, ensuring that the ENS-t-SNE optimization is deterministic.


\section{Experiments on Synthetic Data}

We first verify the ENS-t-SNE algorithm on constructed synthetic data.

\label{sec:cluster_construction}
\noindent\textbf{Construction of Synthetic Data:}
In order to create a dataset with multiple perspectives, each containing multiple clusters, we propose the following procedure.
Fix the number of points and the number of projections; for each projection fix the number of clusters, and the number of points corresponding to each cluster. 
For each perspective, randomly split the points between all the groups and define distances by assigning smaller within-cluster distances and larger between-cluster distances. The procedure is formally defined below.

Let the number of points be $N$ and the number of perspectives be $M$. 
For each perspective 
set
the number of clusters to $NC_m$, for $1 \le m \le M$, and
the number of points corresponding to each cluster to $N_{c, m}$, where $1 \le m \le M$ and $1 \le c \le NC_m$. Note, that the total number $N$ of points per perspective is a fixed constant and these points are in correspondence with each other. That is, $N = \sum_{c = 1}^{NC_m} N_{c, m}$ for all $1 \le m \le M$.

For each perspective $1 \le m \le M$, the sample points with labels $1, 2,\dots, N$ are randomly assigned to one of $NC_m$ clusters. 
That is, the sample point $1 \le i \le N$ will have labels $l_i^1, l_i^2, \dots,l_i^M$. 
Hence, two samples $l_1$ and $l_2$ may share the same label in some of the $M$ perspectives, but are unlikely to share the same labels in all of them. Next, we create the distance matrix between the $N$ datapoints as follows: The observed distance between points $i, j$ for $1 \le i, j, \le N$ in each perspective $m$, where $1 \le m \le M$ is given by
\begin{equation}
\label{eq:cluster_dist}
    \mD_{ij}^m = \left\{
        \begin{array}{ll}
            d_{in}^m + \epsilon & \quad \text{ if } l_i^m = l_j^m  \\
            d_{out}^m + \epsilon & \quad \mathrm{otherwise}
        \end{array}
    \right.
\end{equation}
where $\epsilon \sim N(0, \sigma^2)$ is a normal random variable with mean 0 and standard deviation $\sigma^2$, $d_{in}^m$ corresponds to within cluster distance for the $m$-th perspective and $d_{out}^m$ corresponds to the outside cluster distances.

\begin{figure*}[t]
    \begin{centering}
    \vspace{-.5cm}\includegraphics[height=4cm,width=0.32\linewidth]{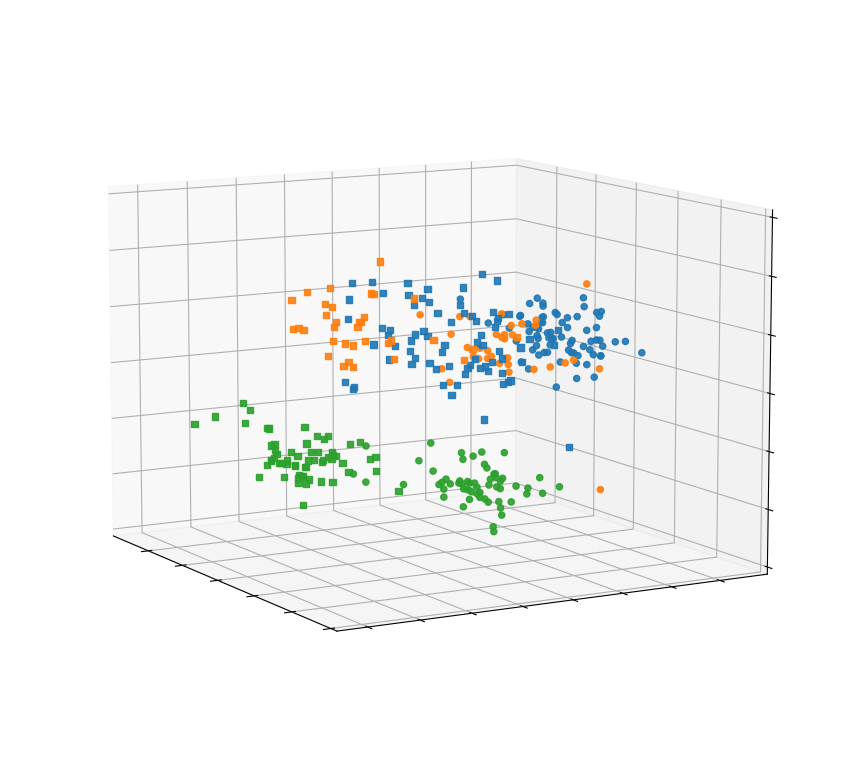}
    \includegraphics[height=4cm,width=0.32\linewidth]{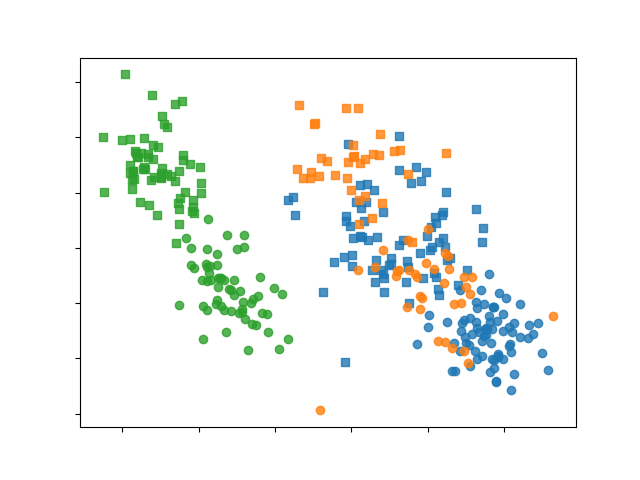}
    \includegraphics[height=4cm,width=0.32\linewidth]{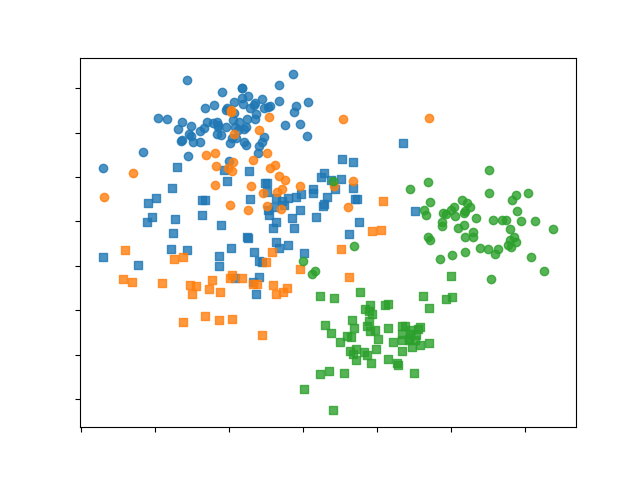}
    \end{centering}
    
    \parbox[c]{0.32\linewidth}{\centering (a)}
    \parbox[c]{0.32\linewidth}{\centering (b)}
    \parbox[c]{0.32\linewidth}{\centering (c)}
    \vspace{-.1cm}
    \caption{The Palmer's Penguins dataset captured by MPSE. (a): The full 3D embedding. (b): The projection capturing physical characteristics, encoded by color. (c): The embedding capturing penguin sex, encoded by shape.
    MPSE mixes the blue and orange clusters 
    also squared and circled clusters.
    }
    \label{fig:penguins_mpse}
\end{figure*}

\begin{figure*}
    \centering
    \includegraphics[width=0.32\linewidth]{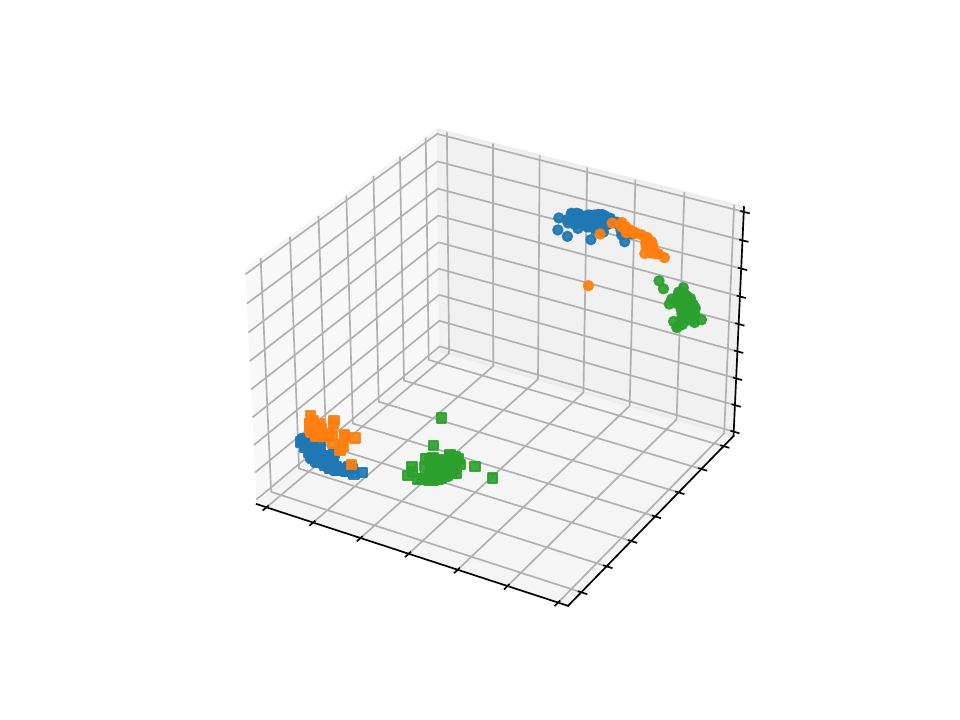}
    \includegraphics[width=0.32\linewidth]{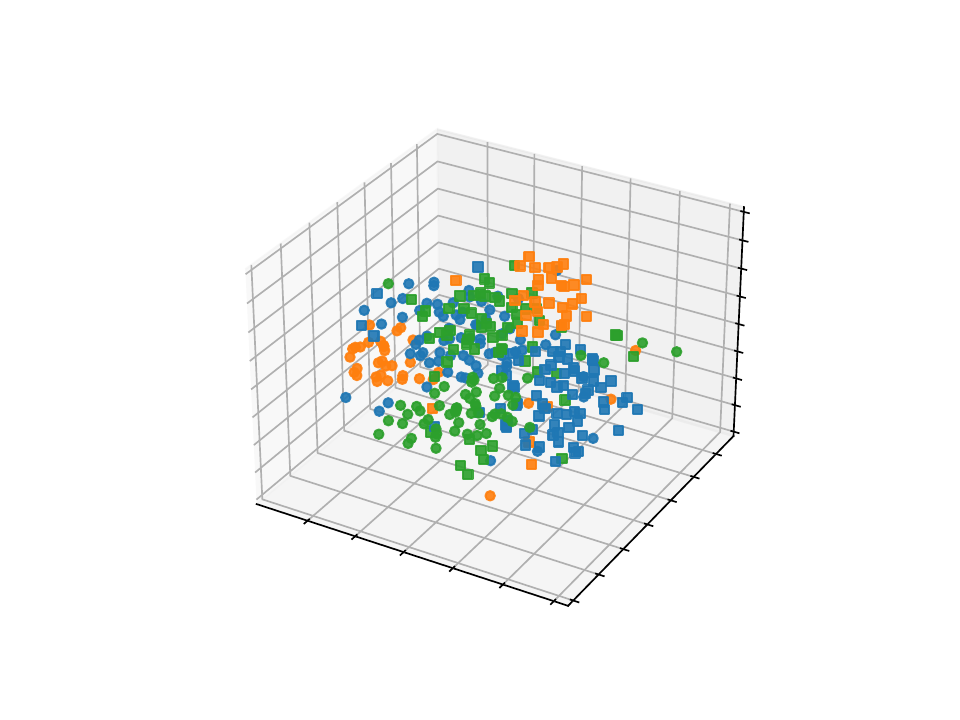}
    \includegraphics[width=0.32\linewidth]{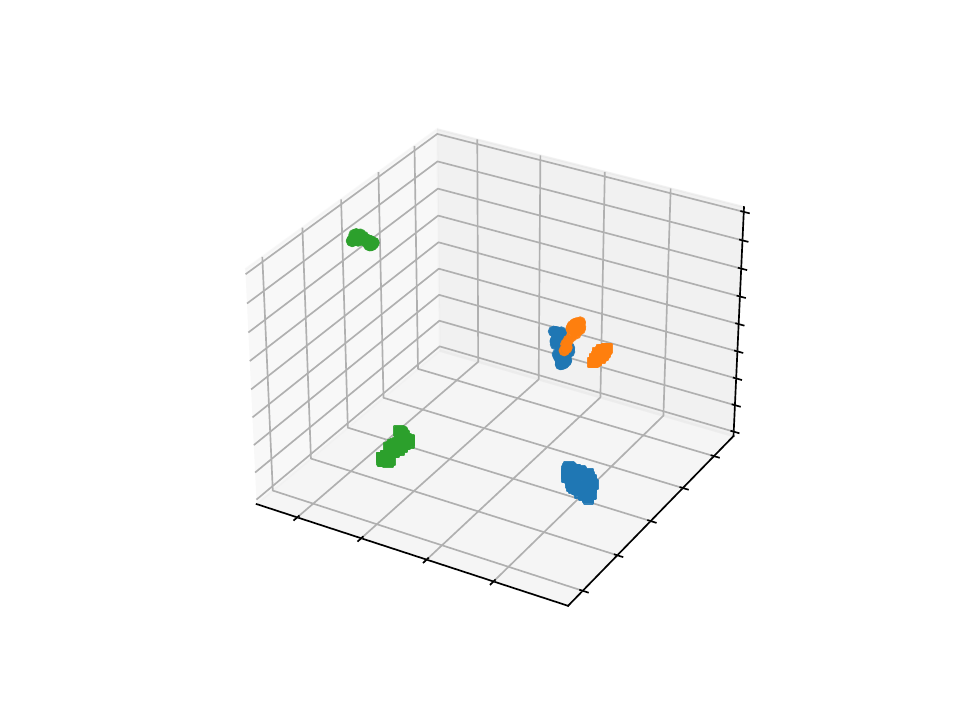}  

    \parbox[c]{0.32\linewidth}{\centering MDS}
    \parbox[c]{0.32\linewidth}{\centering t-SNE}
    \parbox[c]{0.32\linewidth}{\centering UMAP}

    \caption{The Palmer's Penguins dataset embedded by MDS, t-SNE, and UMAP in 3D. While these 
    algorithms are not directly comparable to MPSE and ENS-t-SNE, they are provided for visual comparison. Embeddings of the remaining datasets by these algorithms are in supplemental materials.}
    \label{fig:penguins-others}    
\end{figure*}

\noindent\textbf{Experiments:} Since the goal of ENS-t-SNE is to preserve local structures, this dataset suits its purpose and serves as a good example to show experimentally how the proposed algorithm works.
The aim is to show that 
ENS-t-SNE preserves the local structures of the dataset, that is, each of the perspectives recovers the corresponding original clusters.
We conduct experiments on two synthetic datasets: the first with $400$ datapoints, three perspectives, and two clusters each; the second with $1000$ datapoints, three perspectives, and two, three, and four clusters.

We run ENS-t-SNE
for datasets generated according to Section~\ref{sec:cluster_construction}. 
For the first experiment, we fix the number of perspectives to be $M = 3$, the number of datapoints $N = 400$, the number of clusters per perspective $NC_m = 2$ for $1 \le m \le 3$. We create the distance matrices for each perspective by formula~\eqref{eq:cluster_dist} described in Section~\ref{sec:cluster_construction} with $d_{in} = 1$ and $d_{out} = 2$ and $\sigma^2 = 0.1$. We run the ENS-t-SNE algorithm, summarized in 
Section~\ref{sec:ens_t_sne_algorithm}, for the distance matrices and a fixed perplexity value of $40$.
The result is demonstrated in Figure~\ref{fig:teaser}.

We use separate visual channels to encode the different types of clusters. Specifically, to show the original clusters for the first perspective, we use colors (blue and orange), for the second perspective, we use the shape (circles and squares), and for the third perspective, we use texture, filled and not filled; see Figure~\ref{fig:teaser}. We observe that ENS-t-SNE did a good job finding an embedding in 3D with 3 perspectives, such that for the first perspective the blue and orange datapoints are separated, for the second perspective, datapoints shown in circles and squares are separated, and for the third perspective, filled and not filled datapoints are separated. As expected, ENS-t-SNE split the data into multiple small clusters in 3D, but in such a way that for each 2D perspective, the ones with similar features in 3D overlap and create bigger clusters.

For the second experiment, we fix the number of perspectives to be $M = 3$, the number of datapoints $N = 400$, the number of clusters per perspective $NC_1 = 2$, $NC_2 = 3$, and $NC_3 = 4$. We create the distance matrices for each perspective by formula~\eqref{eq:cluster_dist} described in Section~\ref{sec:cluster_construction} with $d_{in} = 1$ and $d_{out} = 2$ and $\sigma^2 = 0.1$. We run the ENS-t-SNE algorithm
for these distance matrices with perplexity value $40$. The results are demonstrated in Figure~\ref{fig:clustering_234_400}.

We again use separate visual channels to encode the different types of cluster. For the first type of clustering with $NC_1 = 2$ we use  texture: either filled or not filled. The second type of clustering has $NC_2 = 3$ separate clusters, and we use shape: square, triangle, or circle. The third clustering has $NC_3 = 4$ and  we use the color channel: blue, orange, green, and red.  Note how the computed 3D embedding allows us to see in each view 2, 3 and 4 clusters, and that they are well defined and separated in the corresponding projections.

Compare this to how MPSE performs; see Fig.~\ref{fig:mpse-432}. Clusters are more mixed and not clearly separable. For example, the projections that is supposed to separate the data by color, mixes up the blue and red clusters, and both of them are too close to the green. If the cluster identities were not given as part of the input, we would not get as clear an idea about the structure of the data from the MPSE embedding and projections as with ENS-t-SNE.


\section{Experiments on Real-World Data}
\label{sec:applications_real_world}

In this section we demonstrate the application of ENS-t-SNE algorithm on real-world datasets.
The quality of ENS-t-SNE embeddings is affected by choice of subspaces, so determining how to select them is important. We describe two approaches for  subspace selection.

There might be some clear semantic grouping of features, or some grouping of features that is of interest. If this is the case, ENS-t-SNE can be readily applied  and we show an example of such as dataset with the Palmer's Penguins in section~\ref{sec:penguins_sec}.

However, feature grouping is not always clear. The data might have hundreds of features or come from where the meaning of features is unclear. Subspace clustering algorithms can efficiently find subspaces of interest. The USDA food composition dataset is frequently analyzed in subspace clustering literature; we use two interesting subspaces identified by Tatu et al.~\cite{DBLP:conf/ieeevast/TatuMFBSSK12} and show how ENS-t-SNE can provide insights about the different groups in the data. Finally, we apply the CLIQUE subspace clustering algorithm~\cite{DBLP:conf/sigmod/AgrawalGGR98} to the auto-mpg dataset to identify interesting subspaces. The resulting ENS-t-SNE embedding shows patterns that are missing from the standard t-SNE embedding. Further real-world data examples are in supplemental materials.

\subsection{Palmer's Penguins}
\label{sec:penguins_sec}
The Palmer's Penguins dataset~\cite{palmerpenguins} is a collection of $344$ penguins with documented bill length, bill depth, body mass, flipper length, and sex. 
The dataset contains $3$ different species of penguins. 

\begin{figure*}[t]
    \begin{centering}
    \raisebox{-0.5\height}{\includegraphics[width=0.3\linewidth]{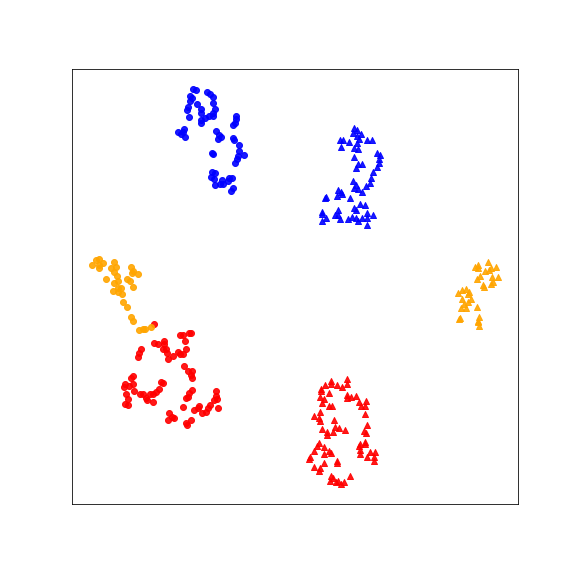}}
    \raisebox{-0.5\height}{\includegraphics[width=0.32\linewidth]{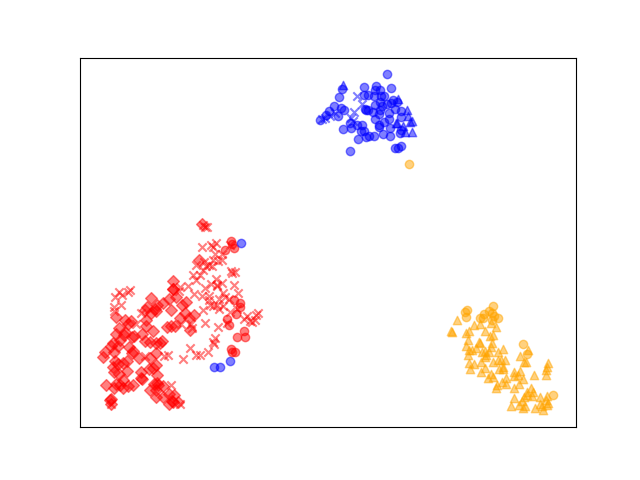}}  
    \raisebox{-0.5\height}{\includegraphics[width=0.32\linewidth]{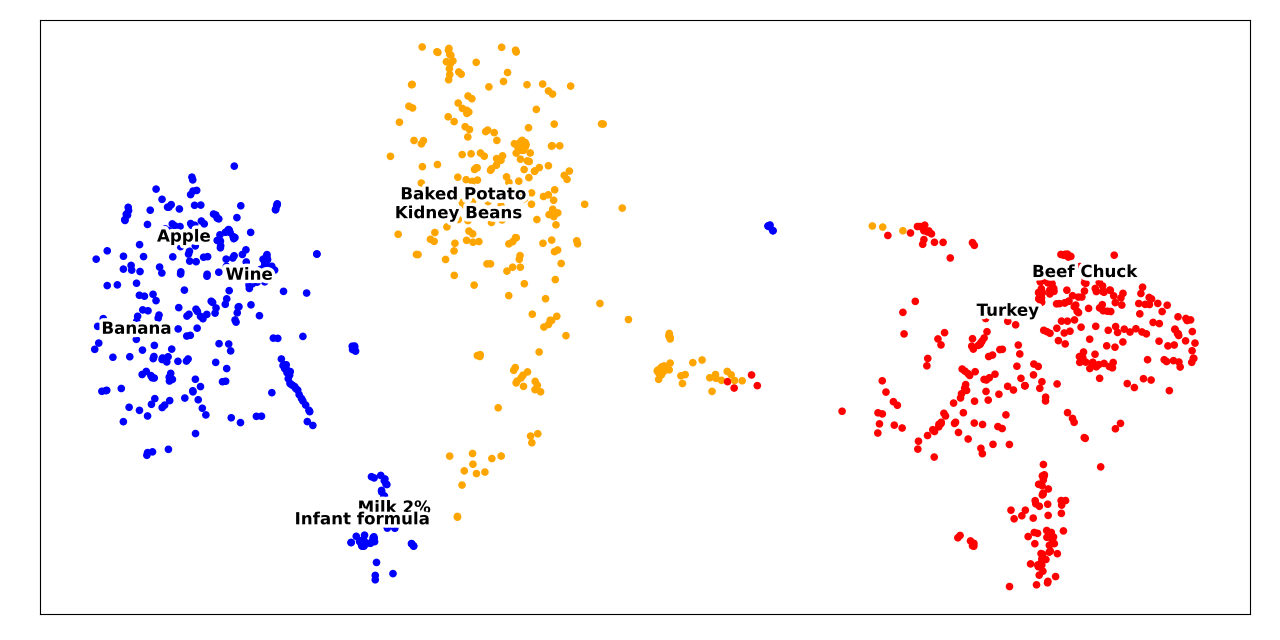}\vspace{-.2cm}}
\end{centering}
 \parbox[c]{0.32\linewidth}{\centering (a)}
    \parbox[c]{0.32\linewidth}{\centering (b)}
    \parbox[c]{0.32\linewidth}{\centering (c)}
\caption{Standard 2D t-SNE embeddings for datasets (a) Palmer's Penguins, (b) auto-mpg, and (c) USDA Food Composition. Note how in each projection, t-SNE misses something that ENS-t-SNE can capture. In (a), the two yellow clusters are split far from each other. In (b), the shapes are mixed with colors. In (c), we cannot see the alternative clustering found in Fig.~\ref{fig:food-comp-ens-tsne}.}
    \label{fig:tsne-default}
\end{figure*}

Applying standard t-SNE on this dataset produces six distinct clusters, one for each sex-species pair; see Fig.~\ref{fig:tsne-default}(a).
However, we can obtain
a more fine-grained information by utilizing ENS-t-SNE and two subspaces: the 4-dimensional physical attributes and the 1-dimensional sex attribute. 
Specifically, we create two pairwise distance matrices, the first one based on the 4D physical attributes which have numerical values. The second one is based on penguin's sex, where the distance between same sex penguins is 0 and the distance between different sex penguins is 1. Running ENS-t-SNE with these distance matrices 
produces the embedding in Fig.~\ref{fig:penguins_ens_t_sne}(a).
The first view (middle) aims to capture the physical attributes and the second view (right) aims to capture the sex information.
While ENS-t-SNE also clusters the data into 6 distinct clusters in 3D, the positions of these clusters are more meaningful.
In the first view (Fig.~\ref{fig:penguins_ens_t_sne}b) the algorithm has captured physical attributes which are largely correlated with the 3 penguin species colored blue for Adelie, orange for Gentoo, and green for Chinstrap. Meanwhile, the second view (Fig.~\ref{fig:penguins_ens_t_sne}c) shows two large clusters based on sex: circles for female penguins and rectangles for male penguins.

While one could run t-SNE on each subdimension separately and plot them as small multiples, such a visualization does not capture the correspondence between datapoints in the two independent embeddings. 
In the ENS-t-SNE embedding, each point belongs to \textit{two} clusters; one for its species and one for its sex. In an interactive environment, one can follow a datapoint from one projection to the other. In other words, there is a transition between the two views in three dimensions that is missing when using small multiples. 
Comparing the t-SNE and ENS-t-SNE embeddings of the dataset we can see that unlike in t-SNE, the positions of the clusters in ENS-t-SNE are meaningful. For example, the two orange clusters in t-SNE embedding in Figure~\ref{fig:tsne-default} are far from each other, while they are close in the ENS-t-SNE embedding in Figure~\ref{fig:penguins_ens_t_sne}.
Thus both datapoint positions and cluster positions are more meaningful in the ENS-t-SNE embedding.

In Fig.~\ref{fig:penguins_mpse}, we compare MPSE embedding of Palmer's Penguins dataset to ENS-t-SNE (Fig.~\ref{fig:penguins_ens_t_sne}) using the same variables. In the first view (Fig.~\ref{fig:penguins_mpse}(b)), blue and orange points are mixed, and in the second view (Fig.~\ref{fig:penguins_mpse}(c)), squared and circled shapes are mixed. ENS-t-SNE, however, offers superior visualization with clear separation of components in each view.
In Fig.~\ref{fig:penguins-others}, we demonstrate the 3D embeddings of the same dataset using MDS, t-SNE, and UMAP. While these methods are not directly comparable to ENS-t-SNE and MPSE, the intent is to highlight the benefits of simultaneous embedding when features are somewhat separated. In all embeddings, points roughly form six clusters, but there is no distinct separation by both color and shape.

\subsection{Food Composition Dataset}
\label{sec:food-comp}
The USDA Food Composition Dataset~\cite{usda-food-comp}
is a collection foods specified by  nutrient components (e.g., calories, proteins, fats, iron, vitamins). 
The full dataset contains over $7000$ entries with $46$ dimensions; after removing entries with missing values we have 948 unique foods, which we use for our experiments.
We walk through how one might use ENS-t-SNE for exploratory data analysis. We first utilize a standard t-SNE projection to get a sense of the data; see Fig.~\ref{fig:tsne-default}(c). There are roughly three large clusters (indicated by color), and we confirm by running a $k$-means clustering on the high dimensional data and see that it matches the clusters in the t-SNE projection. 

Manually examining the clusters for human-interpretable meaning shows that 
the first cluster (red) contains almost entirely meats, while the second and third clusters (blue and orange) appear to have a lot in common, which is unexpected given the $k$-means results and the t-SNE plot.
While the orange cluster contains many grains and vegetables and the blue cluster contains many fruits and beverages, both clusters contain many dairy products such as milks, cheeses, and yogurts. 

We suspect though, that there are more interesting insights to be gained by looking at subspaces of the data since there is seemingly an overlap in the blue and yellow clusters. In fact, this dataset has been used in several subspace clustering papers~\cite{DBLP:conf/ieeevast/TatuMFBSSK12,yuan2012dimension}.
We apply the method from~\cite{DBLP:conf/ieeevast/TatuMFBSSK12} and obtain two  subspaces that we then pass on to ENS-t-SNE to visually investigate further.  
\begin{figure*}
    \begin{centering}
    \includegraphics[width=0.31\linewidth]{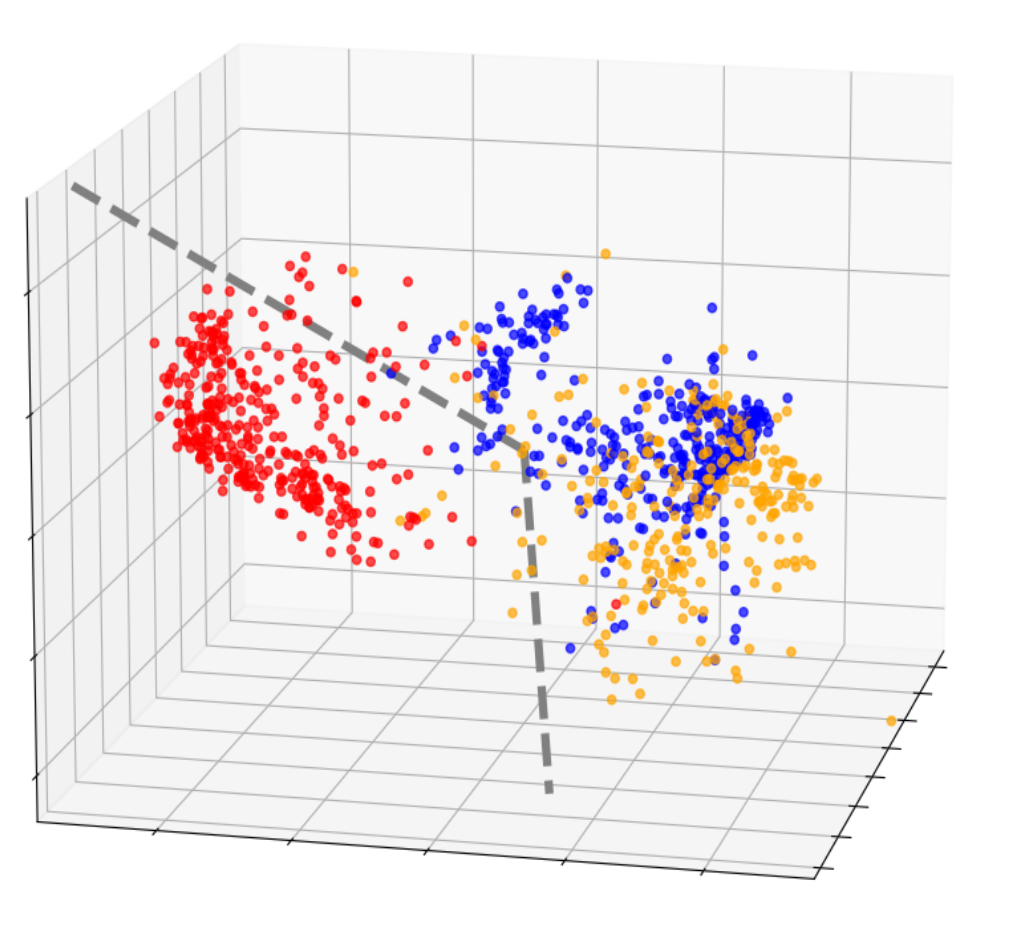}
    \includegraphics[width=0.62\linewidth]{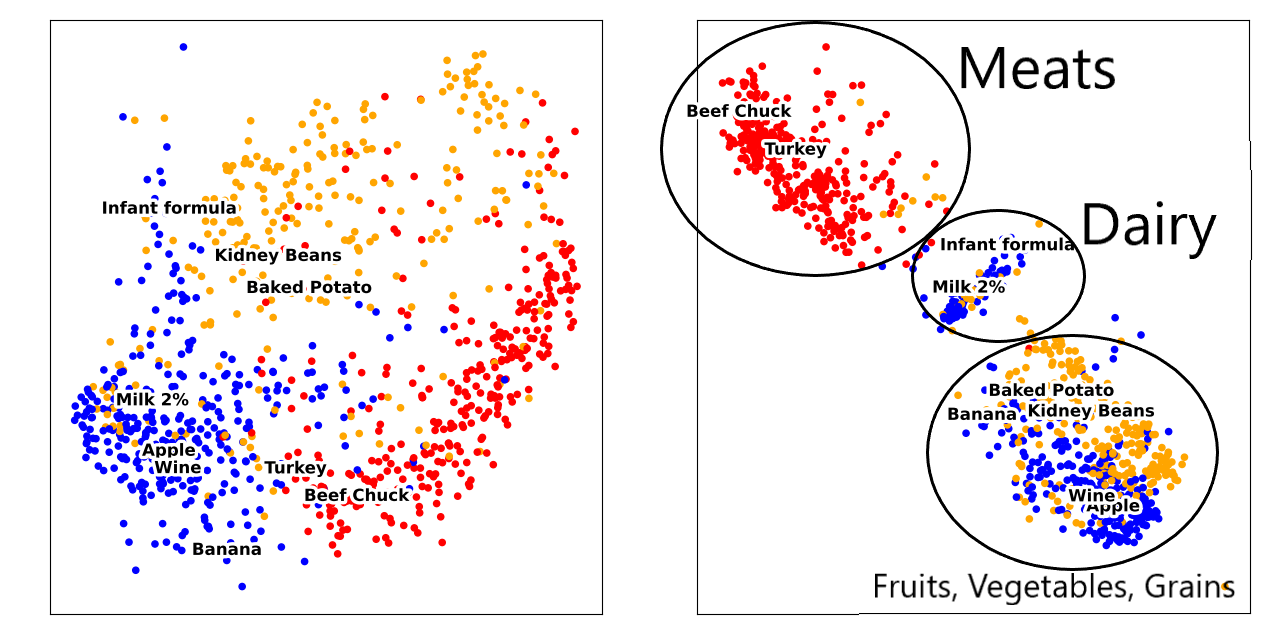} 

        \parbox[c]{0.32\linewidth}{\centering (a)}
        \parbox[c]{0.32\linewidth}{\centering (b)}
        \parbox[c]{0.32\linewidth}{\centering (c)}
    
    \end{centering}

    \caption{ENS-t-SNE applied to the USDA Food Composition dataset. (a): The 3D embedding found by ENS-t-SNE, where each of the three classes have been separated. (b): The first projection corresponding to the water+lipids subspace. The blue and orange clusters (grains and vegetables, fruits and drinks) have been separated, with several blue points mixed in both the orange and red clusters. (c): The second projection corresponding to the proteins+B vitamins subspace. Here the red cluster (meats) has been well separated but the blue and orange cluster contain similar protein and B vitamin components so have been mixed into two clusters, the smaller of which corresponds to dairy and the larger to other meatless foods. }
    \label{fig:food-comp-ens-tsne}
\end{figure*}

We call the first subspace `waters+lipids', as it contains the following features: water, vitamin E, sodium, total lipids, and calories. We call the second subspace `proteins+vitamins' as it contains    protein, vitamin B6, vitamin B12, and vitamin D. 
We computed the corresponding pairwise distance matrices and applied ENS-t-SNE to these pairwise distance matrices; see Fig.~\ref{fig:food-comp-ens-tsne}.

From the 3D embedding obtained by ENS-t-SNE (Figure~\ref{fig:food-comp-ens-tsne}a) we can make the following observations:
In 3D there are strong orange, red, and blue clusters, though there is a portion of the embedding where the colors are mixed, mostly between orange and blue. 
In the first projection 
(Figure~\ref{fig:food-comp-ens-tsne}b),
corresponding to `water+lipids', each cluster has been separated though there are many points that fall between clusters; notably between blue and orange. 
Our implementation provides a hover popout when mousing over datapoints with labels, so we use this to confirm that these are blue points that are particularly `watery' such as lettuce, cucumber, baby-food. 

In the second view 
(Fig.~\ref{fig:food-comp-ens-tsne}c)
which corresponds to the `proteins+vitamins' subspace, we see a much different picture. 
The meats have been strongly clustered, with the red cluster in the top left. 
The blue and orange clusters, that were distinct in the previous projection, have been mixed, indicating that the blue and orange clusters have largely similar protein and vitamin components. Notably, there are two blue/orange clusters. One sits closer to the meats cluster and contains dairy products like milk, infant formula, and cheese. The second cluster contains other meatless foods from both the orange and blue clusters.
Note again that ENS-t-SNE provides a more meaningful embedding of both individual datapoints and clusters. In particular, the 
 similarity  between the blue and orange clusters is missing from the standard t-SNE view, as the distance between clusters in t-SNE is often arbitrary.

\begin{figure*}[h]
    \begin{centering}
    
    \includegraphics[width=0.3\linewidth]{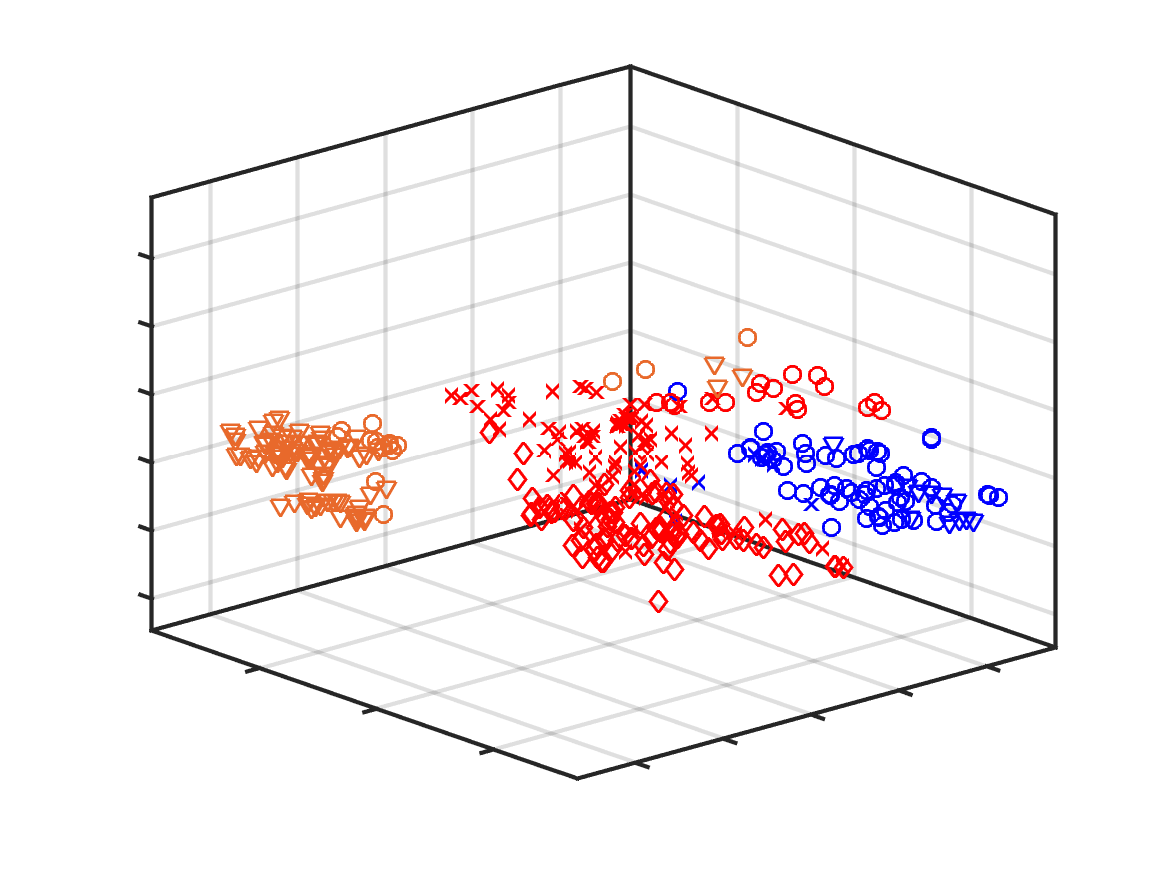}
     \includegraphics[width=0.3\linewidth]{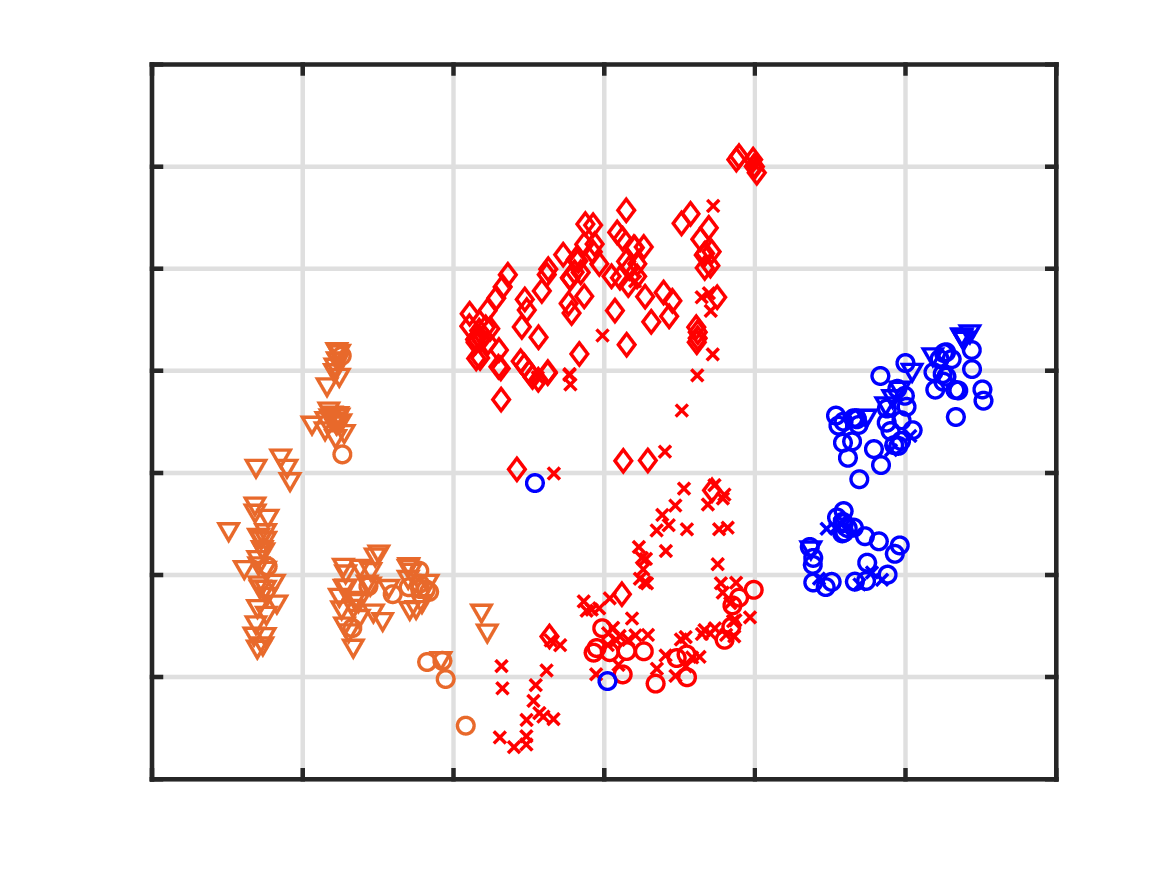}
    \includegraphics[width=0.3\linewidth]{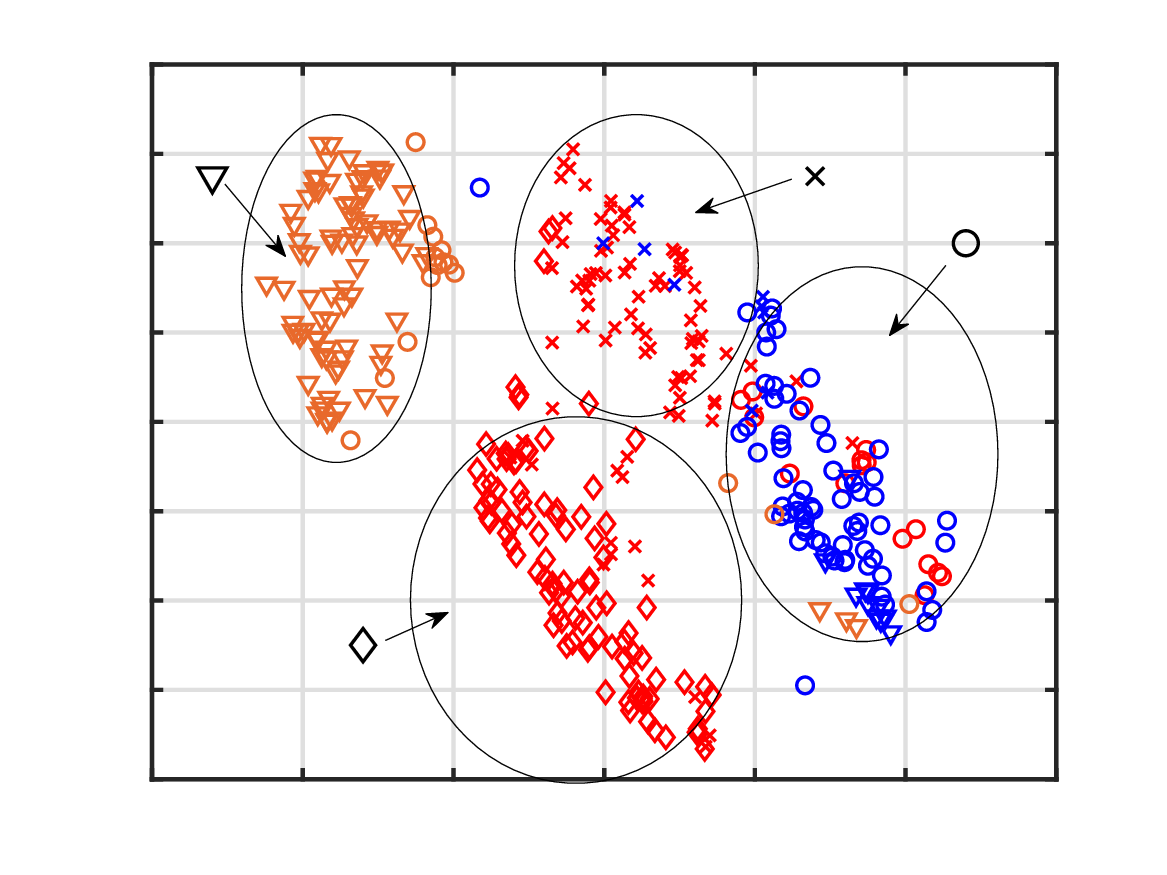}
    \end{centering}
    
    \parbox[c]{0.32\linewidth}{\centering (a)}
    \parbox[c]{0.32\linewidth}{\centering (b)}
    \parbox[c]{0.32\linewidth}{\centering (c)}
    \vspace{-.2cm}
    \caption{Demonstration of ENS-t-SNE  on auto-mpg dataset. Fig.~\ref{fig:car_dataset}(a) demonstrates a glimpse of the 3D ENS-t-SNE embedding of the dataset, Fig.~\ref{fig:car_dataset}(b) and Fig.~\ref{fig:car_dataset}(c) demonstrate the corresponding 2D projections of 3D ENS-t-SNE embedding. 
    Colors indicate number of cylinders: red (4 or less), blue (5-6), orange (7 or more). Shapes indicate weight diamonds (0-25\%), crosses (25-50\%), circles (50-75\%) and triangles (75-100\%). 
    }
\label{fig:car_dataset}
\end{figure*}

\subsection{Auto-MPG Dataset}
\label{sec:auto_mpg}

The auto-mpg dataset from the UCI machine learning repository~\cite{dua2019uci}, provides data for $398$ cars, each with the following 8 attributes: mpg, cylinders, displacement, horsepower, weight, acceleration, model year, origin. The CLIQUE subspace clustering algorithm to the data. We select two ``interesting" subspaces, measured with the Fowlkes-Mallows score~\cite{fowlkes1983method}: the first includes (mpg, cylinders, displacement), and the second includes (horsepower, weight, acceleration).

We apply ENS-t-SNE algorithm for the two subspaces 
using perplexity value 30. The corresponding 3D embedding by ENS-t-SNE is demonstrated in Figure~\ref{fig:car_dataset}. In order to show the clusters in the obtained embedding, we use colors (red, blue, and orange) and shapes (diamond, triangle, square, and crosses).

To show the clusters in embedded dataset, we use the cylinders and the weights. We partition the total data into three groups based on the number of cylinders as follows: In the first group we place all the cars with 4 or fewer cylinders, in the second group are cars with 5 or 6 cylinders, and in the third group we put the cars with more than 6 cylinders. In Figure~\ref{fig:car_dataset} the first group is colored in red, the second group is colored in blue and the third group is colored in orange.
We further partition the dataset into four groups based on the weights according to 25, 50 and 75 quantiles. In Figure~\ref{fig:car_dataset} the datapoints corresponding to the first group are shown in diamond shapes, the second group in crosses, the third group in circles and the fourth group in triangles.

Figure~\ref{fig:car_dataset} shows that ENS-t-SNE was able to find an embedding of the dataset in 3D separating the data into several clusters. The first perspective groups together datapoints with the same colors, i.e., cars with similar numbers of cylinders are grouped together; see the second subfigure of Figure~\ref{fig:car_dataset}. The second perspective groups together datapoints with the same shapes, i.e., cars with similar weights are grouped together; see the third subfigure of Figure~\ref{fig:car_dataset}.

In Figure~\ref{fig:car_dataset} we observe that although in the two perspectives cars are clustered according to corresponding dimensions (number of cylinders and weight), there are some exceptions. For example, the blue outliers in the second (and also third) subfigure  correspond to two exceptional cars which have low weights but higher number of cylinders (5 or 6). 
Consider, for comparison, the standard t-SNE visualization of the same dataset in 2D; see Figure~\ref{fig:tsne-default}(b). The dominant factor for the embedding is the number of cylinders, resulting in three well-separated clusters in the embedding. Note, however, that the t-SNE embedding completely missed the weight information, as there is no pattern between the shapes. Contrast this with the ENS-t-SNE embedding, where both relationships (number of cylinders and weight) can be seen from the corresponding directions; see Fig.~\ref{fig:car_dataset}.

Applying ENS-t-SNE to similar datasets (with multiple interpretations) makes it possible to find a visualization that respects all interpretations. Furthermore, datapoints on the periphery of the clusters and outliers can be interpreted as datapoints that are very similar in one interpretation but completely different in others.


\section{Quantitative Evaluation}

The real-world and synthetic examples above show that  ENS-t-SNE can provide  meaningful 3D datapoint positions and cluster positions. Here we provide some quantitative data. 

While MDS, t-SNE and UMap can produce 3D embeddings, they cannot optimize per-projection views, as MPSE and ENS-t-SNE can. We could use single projection techniques to obtain subspace embeddings but the resulting plots will be largely unrelated. Thus  the only available technique that can be directly  compared to ENS-t-SNE is MPSE.
With this in mind, we quantitatively evaluate MPSE and ENS-t-SNE, using the metrics \textit{trustworthiness}, \textit{continuity}, \textit{neighborhood hit} and \textit{stress}, as recommended in a recent survey~\cite{DBLP:journals/tvcg/EspadotoMKHT21}.

For each metric, we compute the 3D values with respect to all distances, as well as the per-perspective values with respect to each subspace.
For the following definitions, let $N$ be the size of a dataset (number of points), and $K$ be a parameter for the size of a neighborhood. For trustworthiness, continuity, and neighborhood hit we use $K=7$ for all evaluations, as in~\cite{DBLP:journals/tvcg/EspadotoMKHT21}.
We use the same perplexity value for each dataset as described in their figures: perplexity of 40 for the Palmer's Penguins and value of 30 for the Auto-MPG and Food Composition dataset. These values were chosen by investigating a range in (20, 100) and selecting the best quantitative embedding. 

\textbf{Trustworthiness} measures how well neighbors in the embedding match the neighbors in high dimensional space, with large errors penalized heavily. As its name implies, a high trustworthiness is a good indication that one can trust the local patterns in the embedding. This is a measure of \textit{precision} with respect to clusters:
\begin{equation}
    \textstyle
    \label{eq:trustworthiness}
    1 - \frac{2}{NK(2N - 3K - 1)}\sum_{i=1}^N \sum_{j\in U_i^K} r(i,j) - K
\end{equation}
where $U_i^K$ is the set of points among the $K$ nearest neighbors of point $i$ in the embedded space but not in the high dimensional space and $r(i,j)$ is the rank of the embedded point $j$ with respect to the embedded nearest neighobors of point $i$.

\textbf{Continuity} is related to trustworthiness, but measures how many neighors are \textit{missing} in the embedding but are present in high dimensional space. A high continuity score means that most of the neighborhood around a given point is nearby (rather than far away). This is a measure of \textit{recall} with respect to clusters:
\begin{equation}
    \label{eq:continuity}
    \textstyle
    1-\frac{2}{NK(2N-3K-1)} \sum_{i=1}^N \sum_{j\in V_i^K} \hat{r}(i,j) - K
\end{equation}
where $V_i^K$ is the set of points among the $K$ nearest neighbors of point $i$ in the high dimensional space but not in the embedding, and $\hat{r}(i,j)$ is the rank of point $j$ with respect to $i$ in the high dimensional space.  

\textbf{Neighborhood Hit (NH)} measures the proportion of nearest neighbors of a point in the embedding which have the same label, similar to a $k$-nearest neighbor classifier. 
In order to use this measure, we need labels which we have (or generate) for each of our datasets. Since our datasets have two or more sets of different labels, when we compute NH on the three dimensional embedding we take the Cartesian product of the two labelings. For instance, if a penguin is both an Adelie penguin and female, then we assign it a label of (Adelie, female). Formally, NH is defined as follows:
\begin{equation}
    \label{eq:nh}
    \textstyle
    \sum_{i=1}^N \frac{1}{KN} |j\in Ne_i^K :l_j = l_i|
\end{equation}
where $Ne_i^K$ is the $K$ nearest neighborhood of point $i$ in the embedding, and $l_i$ is the label of point $i$.
We note that the food composition dataset does not have ground truth labels, but there is good evidence that our clustering is accurate. The clustering has a high silhouette score and is visually validated in Fig~\ref{fig:tsne-default} and Fig.~\ref{fig:food-comp-ens-tsne}.

\textbf{Stress} is the sum of squared differences in the distances between the embedding and high dimensional space:
\begin{equation}
    \label{eq:stress}
    \textstyle
    \sum_{i<j} (||X_i - X_j|| - d_{i,j})^2
\end{equation}
where $X_i$ is the embedded position of point $i$ and $d_{i,j}$ is the distance between points $i,j$ in the high dimensional space.
Low stress indicates good distance preservation. Note that MPSE directly optimizes stress for each view, so we expect it to often outperform ENS-t-SNE.
Table~\ref{tab:quant-table} shows the results of these metrics, averaged over 10 trials each.
Note how ENS-t-SNE tends to outperform MPSE on the continuity, trustworthiness, and NH scores for the 2d views indicating that as expected ENS-t-SNE is more reliably capturing local structures in the projected views. Low stress is not the goal of this algorithm, so having high stress compared to MPSE is acceptable.
ENS-t-SNE can achieve lower stress in one view than MPSE, since MPSE minimizes the average stress per projection, meaning that ENS-t-SNE may produce unbalanced projections with respect to stress where MPSE will produce more balanced projections.

\begin{figure}[t]
    \centering
    \includegraphics[width=0.49\linewidth]{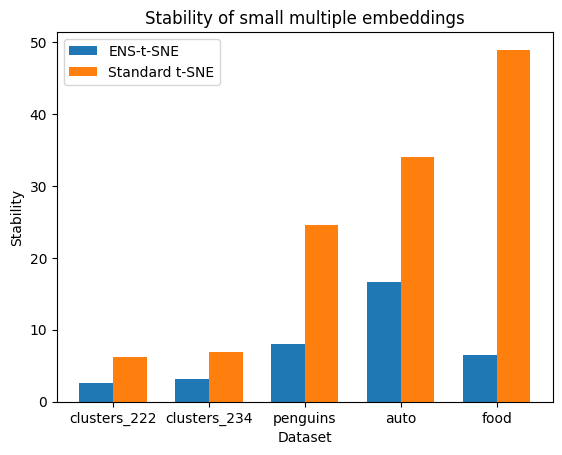} 
    \includegraphics[width=0.49\linewidth]{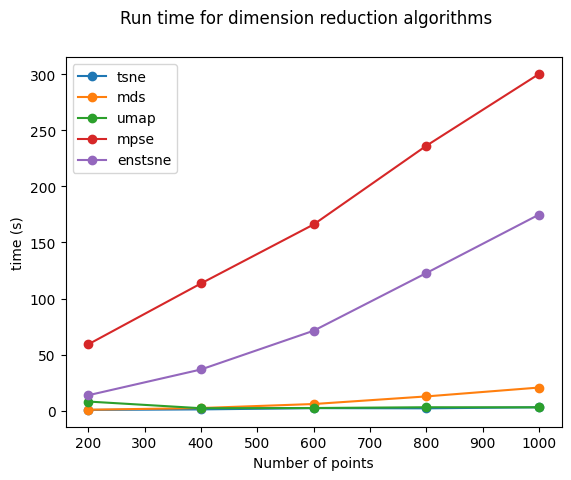}

    \parbox[c]{0.49\linewidth}{\centering (a)}
    \parbox[c]{0.49\linewidth}{\centering (b)}
    
    \caption{(a) The stability score of ENS-t-SNE (blue) and standard t-SNE (orange). A low score means there was little movement between subspace embeddings and a high score means there was large movement. Note that ENS-t-SNE has a much better average score for movement between subspace embeddings.
    (b) How related dimension reduction algorithms scale in the number of datapoints.}
    \label{fig:movement}
    \vspace{-0.5cm}
\end{figure}

\textbf{Stability}. 
Although we cannot directly quantitatively compare to t-SNE (or other similar dimension reduction embeddings), we can measure how similar a set of projections are. It is known that for many comparative tasks it is desirable to have as little change as possible while still being faithful to the data. This notion is often referred to as stability or preservation of the mental map~\cite{archambault2010animation,eades1991preserving}.
We show that ENS-t-SNE is more stable when computing a set of projections of subspaces than standard t-SNE on the same set of subspaces.

We adapt the notion of  \textit{stability} to a pair of embeddings, $stability(E_1, E_2)$ computed by first aligning $E_2$ to $E_1$ as close as possible using affine transformations and then computing the average distance between corresponding points in the two embeddings. 
For a series of embeddings $E_1, \dots E_n$ we compute the average pairwise stability, i.e., $\sum_{i<j}\text{stability}(E_i,E_j)$.

In all datasets under consideration, ENS-t-SNE produces better stability between small multiple projections. 
The results for an average of 30 trials are shown in Fig.~\ref{fig:movement}. Note that ENS-t-SNE has consistently better stability than t-SNE, which is not unexpected, as t-SNE has no correspondence between small multiple projections.

\begin{table}[t]
    \centering
    \resizebox{\linewidth}{!}{
    \begin{tabular}{|c|c c || c c || c c |}
    \hline
     & \multicolumn{6}{|c|}{Palmer's Penguins} \\
     \cline{2-7}
     & \multicolumn{2}{c}{3d} & \multicolumn{2}{c}{view 1} & \multicolumn{2}{c|}{view 2} \\ \cline{2-7} 
     & ENS-t-SNE & MPSE & ENS-t-SNE & MPSE & ENS-t-SNE & MPSE\\ \hline
cont. $\uparrow$ & \textbf{0.9858} & 0.9842 & \textbf{0.9871} & 0.9370 & \textbf{0.7848} & 0.6862\\ \hline 
trust. $\uparrow$ & 0.9846 & \textbf{0.9853} & \textbf{0.9909} & 0.9386 & \textbf{0.7649} & 0.7639\\ \hline 
NH $\uparrow$ & \textbf{0.9871} & 0.9472 & \textbf{0.9785} & 0.9232 & 1.0000 & 1.0000\\ \hline 
stress $\downarrow$ & 0.7720 & \textbf{0.6415} & \textbf{0.2659} & 0.2997 & \textbf{0.0000} & 0.0569\\ \hline

    \hline \hline 
     & \multicolumn{6}{|c|}{Auto-MPG} \\
     \cline{2-7}
     & \multicolumn{2}{c}{3d} & \multicolumn{2}{c}{view 1} & \multicolumn{2}{c|}{view 2} \\ \cline{2-7}
     & ENS-t-SNE & MPSE & ENS-t-SNE & MPSE & ENS-t-SNE & MPSE\\ \hline
cont. $\uparrow$ & 0.9636 & \textbf{0.9797} & \textbf{0.9897} & 0.9493 & \textbf{0.9864} & 0.9859\\ \hline 
trust. $\uparrow$ & 0.9753 & \textbf{0.9846} & \textbf{0.9914} & 0.9625 & \textbf{0.9952} & 0.9860\\ \hline 
NH $\uparrow$ & 0.8451 & \textbf{0.8550} & 0.9913 & 0.9913 & \textbf{0.8958} & 0.8859\\ \hline 
stress $\downarrow$ & \textbf{0.5099} & 1.9999 & \textbf{1.9974} & 1.9974 & 0.2837 & \textbf{0.1888}\\ \hline

    \hline \hline 
     & \multicolumn{6}{|c|}{USDA Food Composition} \\
     \cline{2-7}
     & \multicolumn{2}{c}{3d} & \multicolumn{2}{c}{view 1} & \multicolumn{2}{c|}{view 2} \\ \cline{2-7}
     & ENS-t-SNE & MPSE & ENS-t-SNE & MPSE & ENS-t-SNE & MPSE\\ \hline
cont. $\uparrow$ & 0.9573 & \textbf{0.9653} & \textbf{0.9800} & 0.9449 & \textbf{0.9920} & 0.8987\\ \hline 
trust. $\uparrow$ & 0.9596 & \textbf{0.9852} & \textbf{0.9850} & 0.9390 & \textbf{0.9971} & 0.8936\\ \hline 
NH $\uparrow$ & \textbf{0.8852} & 0.8834 & \textbf{0.8822} & 0.8318 & 0.7985 & \textbf{0.8336}\\ \hline 
stress $\downarrow$ & \textbf{0.5971} & 1.9999 & \textbf{0.6904} & 1.9999 & 1.9981 & \textbf{1.9981}\\ \hline  \hline 
    
    \end{tabular}}
    \caption{Quantitative comparison between ENS-t-SNE and MPSE. Each pair of columns denotes what view is measured; either the full three dimensional embedding with respect to the full set of distance matrices or for the two dimensional projections with respect to the corresponding distance matrix. 
    For each view, the best score is bolded. }
    \label{tab:quant-table}
    \vspace{-0.5cm}
\end{table}

\subsection{Scalability}

In this section we consider the scalability of ENS-t-SNE with respect to the number of perspectives, the number of clusters per perspective, and the number of datapoints. In particular, the goal is to evaluate how the accuracy or the speed of ENS-t-SNE is affected as these parameters increase in value.
The results indicate that the runtime of ENS-t-SNE scales reasonably well as these parameters increase. 
The accuracy decreases when the number of perspectives and clusters grows. 
However, if the number of perspectives is two the accuracy does not decrease as the number of clusters increase.

In order to measure the accuracy of the embedding for a dataset containing several clusters, we define the separation error as follows: For a 2D image containing two labels, the best linear classifier is found and the proportion of errors in this classification is returned. For a 2D image containing more than two labels, a linear classifier for every possible combination of two labels is computed, and the average proportion of errors between all combinations is returned. For a 3D embedding with multiple perspectives, with each image having two or more labels, the separation error in each image is computed and the average is returned.

\begin{figure}[!ht]
    \begin{centering}
        \includegraphics[width=0.49\linewidth]{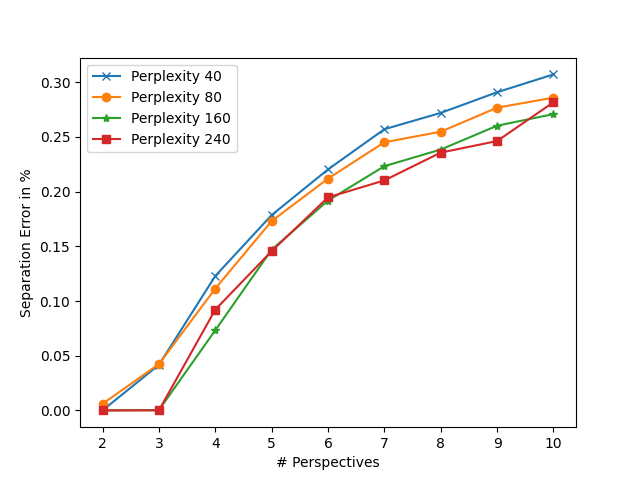}
        \includegraphics[width=0.49\linewidth]{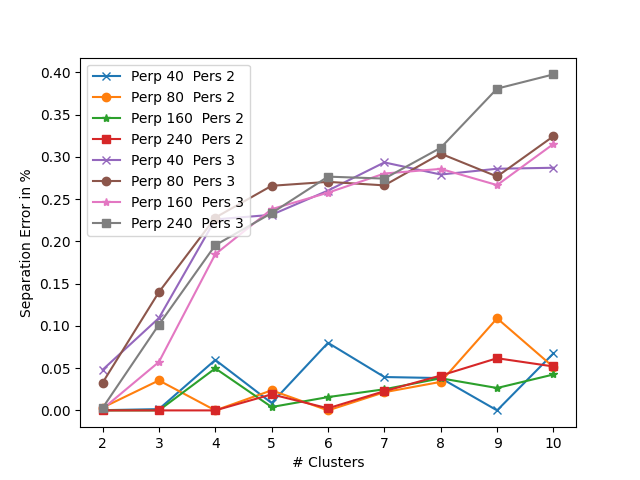}
        
        \parbox[c]{0.49\linewidth}{\centering (a)}
        \parbox[c]{0.49\linewidth}{\centering (b)}
    
        \includegraphics[width=0.49\linewidth]{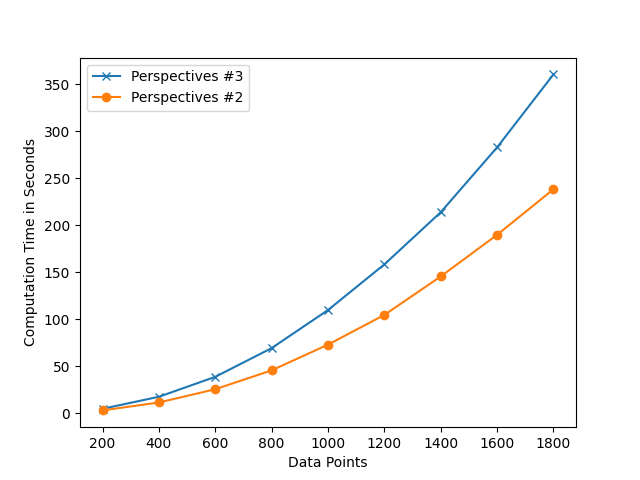}
        \includegraphics[width=0.49\linewidth]{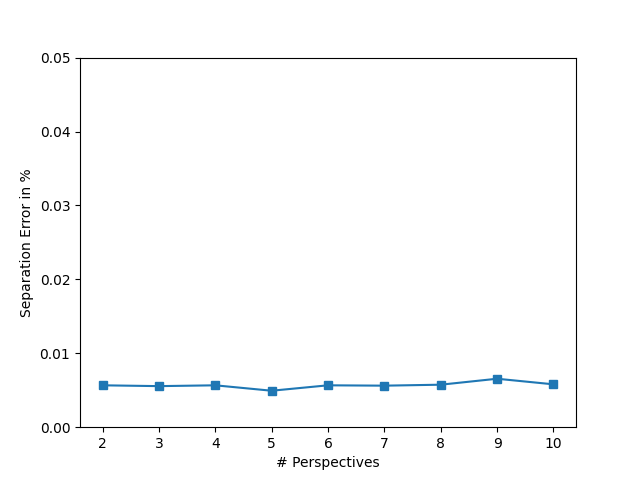}

        \parbox[c]{0.49\linewidth}{\centering (c)}
        \parbox[c]{0.49\linewidth}{\centering (d)}
        
    \end{centering}
        \caption{Demonstration of the scalability of the ENS-t-SNE algorithm. Fig.~\ref{fig:scalability_tests}(a) shows the separation error for 400 samples with two clusters in each perspective.
        Fig.~\ref{fig:scalability_tests}(b) shows the separation error for 400 samples with multiple clusters in each perspective.
        Fig.~\ref{fig:scalability_tests}(c) shows the running time for 2 and 3 perspectives, for 2 and 3 clusters, while varying the sample sizes.
        Fig.~\ref{fig:scalability_tests}(d) shows the separation error for 1000 samples uniformly distributed in a solid 3D ball. Perspectives chosen at random, labels chosen depending on which side of each 2D image each point falls into. The perplexity is fixed to $600$, the number of perspectives vary.}
        \label{fig:scalability_tests}
        \vspace{-0.5cm}
\end{figure}

To check the scalability of the proposed ENS-t-SNE algorithm we create datasets as described in Section~\ref{sec:cluster_construction}. We consider $N = 400$ datapoints and vary the number of perspectives $M = 2, 3, \dots, 10$,  making sure that each perspective has $NC_m = 2$ identifiable clusters in it. 
We then apply ENS-t-SNE  for each of these datasets using perplexity values $40, 80, 160, 240$ and report the results in Fig.~\ref{fig:scalability_tests}(a).
The x-axis of Figure~\ref{fig:scalability_tests}(a)
shows the number of perspectives and the y-axis shows the separation error.
The results indicate that for a small number of perspectives (2,3) the separation error is small and as the number of perspective increases the error grows.

Next, we test the scalability of the algorithm as the number of clusters per perspectives increases. We create the datasets in a similar fashion setting $N = 400$, $M = 2, 3$, and varying the number of clusters per perspective $NC_m = 2, 3, \dots, 10$. We run  ENS-t-SNE with perplexity values $40, 80, 160, 240$ and report the separation error in 
Fig.~\ref{fig:scalability_tests}(b): 
as the number of clusters increase for 3 perspectives, the separation error grows, while for 2 perspectives the separation error is stable. 

We continue by analyzing the influence of the number of datapoints on the running time of the algorithm. 
For this purpose, we create datasets containing clusters according to Section~\ref{sec:cluster_construction}. We set the number of perspectives $M = 2, 3$ and the number of clusters per perspective $NC_m = 2$, while varying the total number of datapoints $N$ from $200$ to $1800$ in increments of $200$. We run ENS-t-SNE for these datasets with perplexity value $0.2 * N$ and report the running time as a function of $N$ in 
Figure~\ref{fig:scalability_tests}(c).
The results indicate a steady increase in  running time as the number of datapoints and the number of perspectives grow.

The final experiment tests the effect of the number of perspectives on the accuracy of the algorithm. We generate data as follows: We uniformly distribute $1000$ points in a solid 3D ball, and randomly select several perspectives. 
We label each point in each perspective according to on which side of the perspective the points fall into. 
The perplexity is fixed to $600$ and the number of perspectives vary from $2$ to $10$.
We report the separation error vs the number of perspectives in 
Fig.~\ref{fig:scalability_tests}(d).
The results indicate that when there are no forced clusters, the algorithm has more freedom to separate the data into parts that respect the original label assignment, providing more stable separation error.


\section{Limitations}
Naturally, there are many limitations to ENS-t-SNE. Here we consider only a partial list, starting with scalability. 
It is known that t-SNE is computationally expensive and in this prototype we have not yet considered applying ideas for speeding it up, such as those in~\cite{maaten2014accelerating,tang2016visualizing}.

While MPSE~\cite{hossain2021multi} focuses on simultaneously capturing global distances between objects and ENS-t-SNE aims to capture local  neighborhoods, other approaches for  dimension reduction, such as UMAP~\cite{mcionnes2018umap}, optimize both at the same time. It would be worthwhile to quantitatively verify the extend to which these goals can be realized by the different approaches.  
The utility of ENS-t-SNE depends on finding interesting subspaces/subdimensions and combinations thereof. We have not yet considered automating the process by using approaches such as those for subspace clustering as part of the ENS-t-SNE pipeline.
Setting up the distance matrices that ENS-t-SNE needs can be done in different ways, as illustrated in the different examples in the paper. Evaluating different approaches and automating the process remains to be done.

While we expect that 3D ENS-t-SNE embeddings might be easier to interpret and work with, compared to small multiple type visualizations (with or without linked views), we are not aware of human-subject studies to validate this intuition. 
Although we were inspired by 3D physicalizations one can ``walk around" and interact with, a thorough human subjects study is required to verify that this method supports tasks effectively.

\section{Conclusions and Future Work}

We described ENS-t-SNE, a generalization of t-SNE, which computes a 3D embedding of a dataset along with a set of 2D projections that optimize subspace clustering information. 
We note that while our paper describes ENS-t-SNE in 3D, the technique can be applied to higher dimensions (lower than the number of input dimensions).

As the main part of the paper  describes the proposed ENS-t-SNE algorithm (from the idea to the implementation), the quantitative and qualitative evaluation is just sketched out here. 
Nevertheless, several different types of experiments, on synthetic and real-world datasets, indicate that ENS-t-SNE can indeed simultaneously capture multiple different types of relationships defined on the same set of high-dimensional objects. All source code, experimental data, and analysis described in this paper are available on github (along with a video explanation) at \href{https://github.com/enggiqbal/MPSE-TSNE}{https://github.com/enggiqbal/MPSE-TSNE}.

An interesting direction that we began to explore is to extend the objective function such that each perspective shows the t-SNE embedding for different values of perplexities; see the supplemental material.
Another possible application is using ENS-t-SNE to visualize image datasets, based on different parts of the input images. We include some preliminary results for the MNIST dataset in the supplemental material.

ENS-t-SNE generalizes t-SNE to multiple perspectives. Generalizing other dimensionality reduction techniques, such as UMAP~\cite{mcionnes2018umap} might be of interest. Combining local and global perspective at the same time, for example by combining ENS-t-SNE and MPSE~\cite{hossain2021multi}, might provide embeddings that allow us to balance local and global distance preservation.



\bibliographystyle{abbrv-doi-hyperref-narrow}
\bibliography{multiview}        



\onecolumn

\setcounter{page}{1}

{\LARGE  \textbf{Embedding Neighborhoods Simultaneously t-SNE: Supplemental Materials}}
\appendix             

\text{ }
\newline
\text{ }

In Appendix~\ref{sec:pseudocode} we provide a practical pseudocode implementation of ENS-t-SNE. In Appendix~\ref{sec:mnist} we discuss an application of ENS-t-SNE to MNIST dataset to find a 3D visualization of it and in Appendix~\ref{sec:perplexity_effect} we discuss an extendion of ENS-t-SNE to visualize the effect of perplexity parameter.
\section{ENS-t-SNE Pseudocode}
\label{sec:pseudocode}

\begin{algorithm}
\caption{Practical implementation of ENS-T-SNE}
\label{algo:ens-t-sne}
\begin{algorithmic}
    \Require $N\times N$ pairwise distance matrices: $\mD^1, \mD^2,\dots, \mD^M$, perplexity parameter $\textrm{Perp}$.
    \For{perspectives $m=1,2,\dots,M$}
        \State Compute $P^m = f(\mD^m, \textrm{Perp})$
    \EndFor
    \State Assign initial values to $Y$, $\Pi = [\Pi^1,\Pi^2,\dots,\Pi^M]$, $\mu, \nu$.
    \For{batch size = n/4, n/2}
        \For{$t=1,2,T$}
            \State $\mY = \mY + \mu \nabla_{\mY}^{ss} \widetilde{C}(\mY,\Pi)$
            \State $\mu = \mu + 1$
        \EndFor
    \EndFor
    \For{batch size = n/4, n/2}
        \For{$t=1, \dots,T$}
            \State $\mY = \mY + \mu \nabla_{\mY}^{ss} \widetilde{C}(\mY,\Pi)$
            \State $\Pi = \Pi + Q \left( \nu \nabla_\Pi^{ss} \widetilde{C}(\mY, \Pi) \right)$
            \State $\mu = \mu + 1$
            \State $\nu = \nu + 1$
        \EndFor
    \EndFor
    \Ensure $\mY$, $\Pi^1,\Pi^2,\dots,\Pi^M$.
\end{algorithmic}
\end{algorithm}

\section{Visualizing MNIST Dataset by ENS-t-SNE}
\label{sec:mnist}

\begin{figure}[h!]
    \centering
    \includegraphics[width=0.19\linewidth]{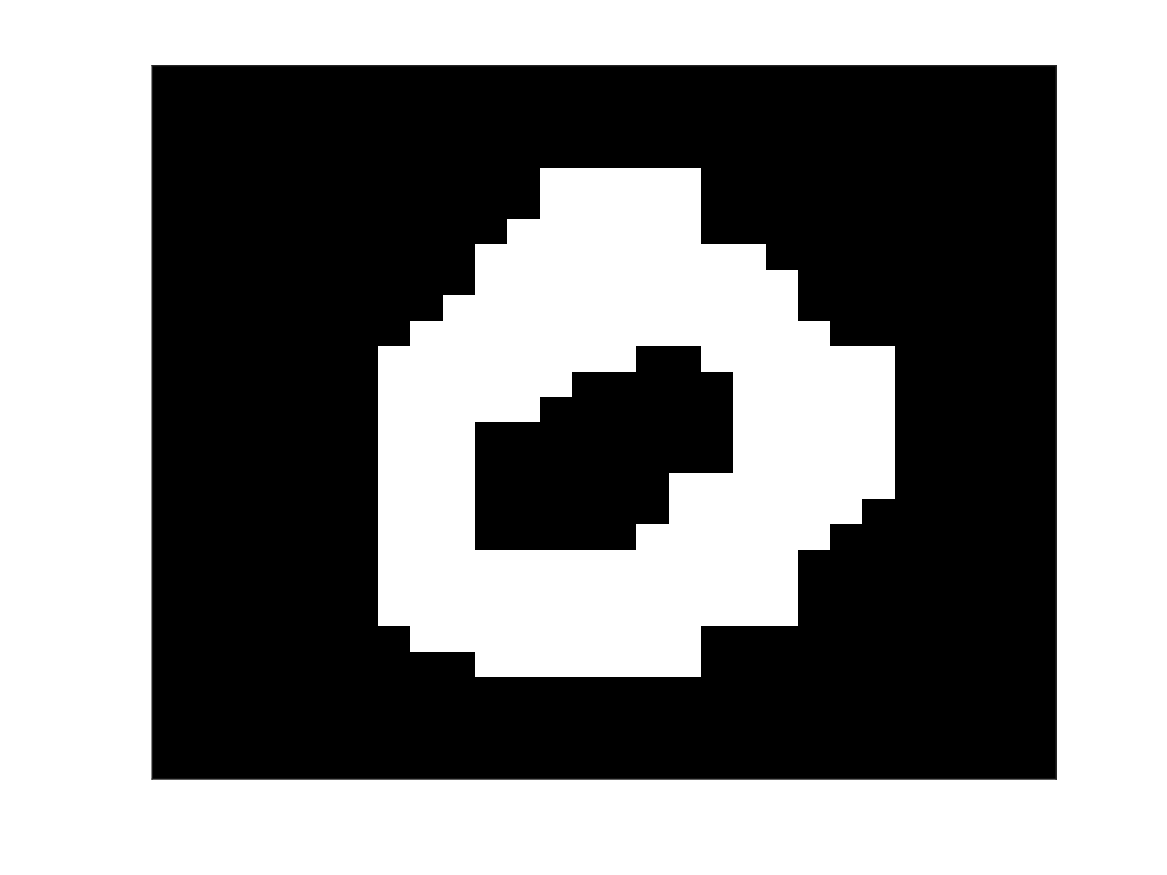}
    \includegraphics[width=0.19\linewidth]{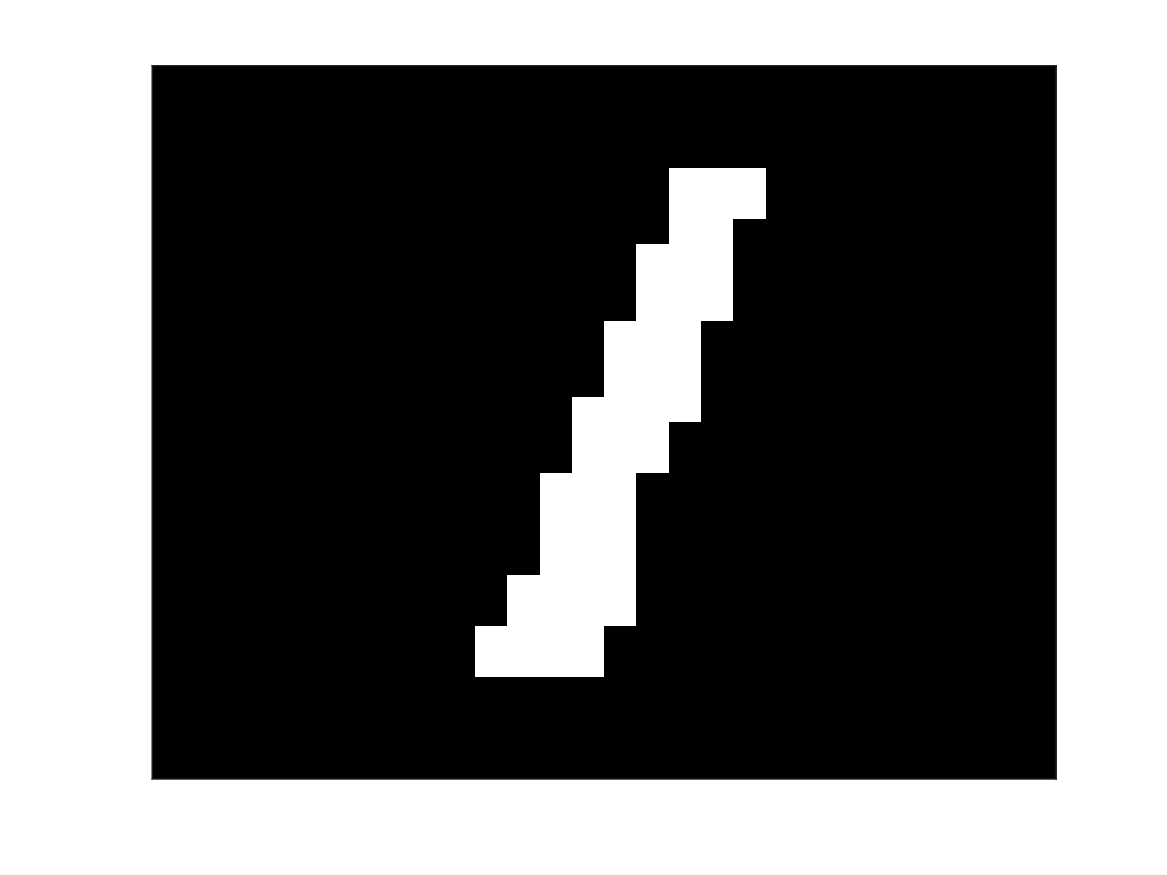}
    \includegraphics[width=0.19\linewidth]{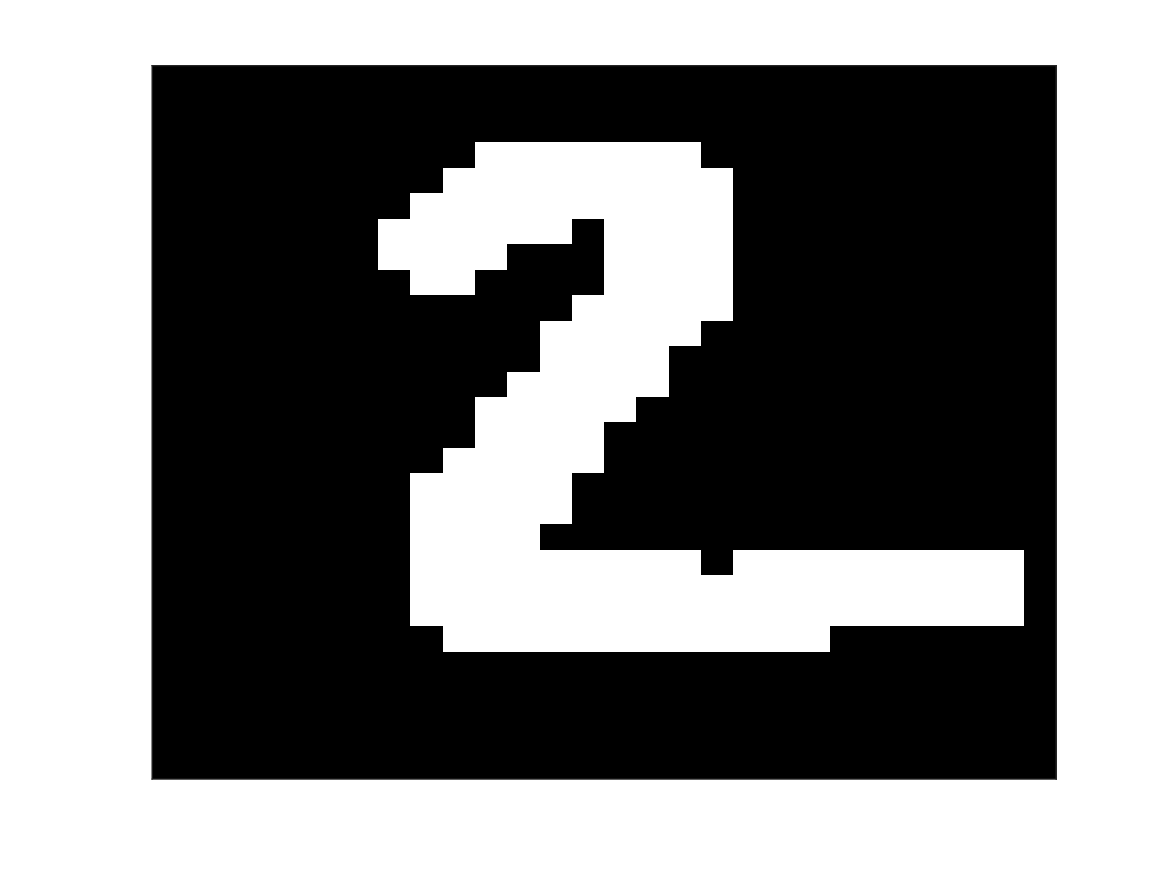}
    \includegraphics[width=0.19\linewidth]{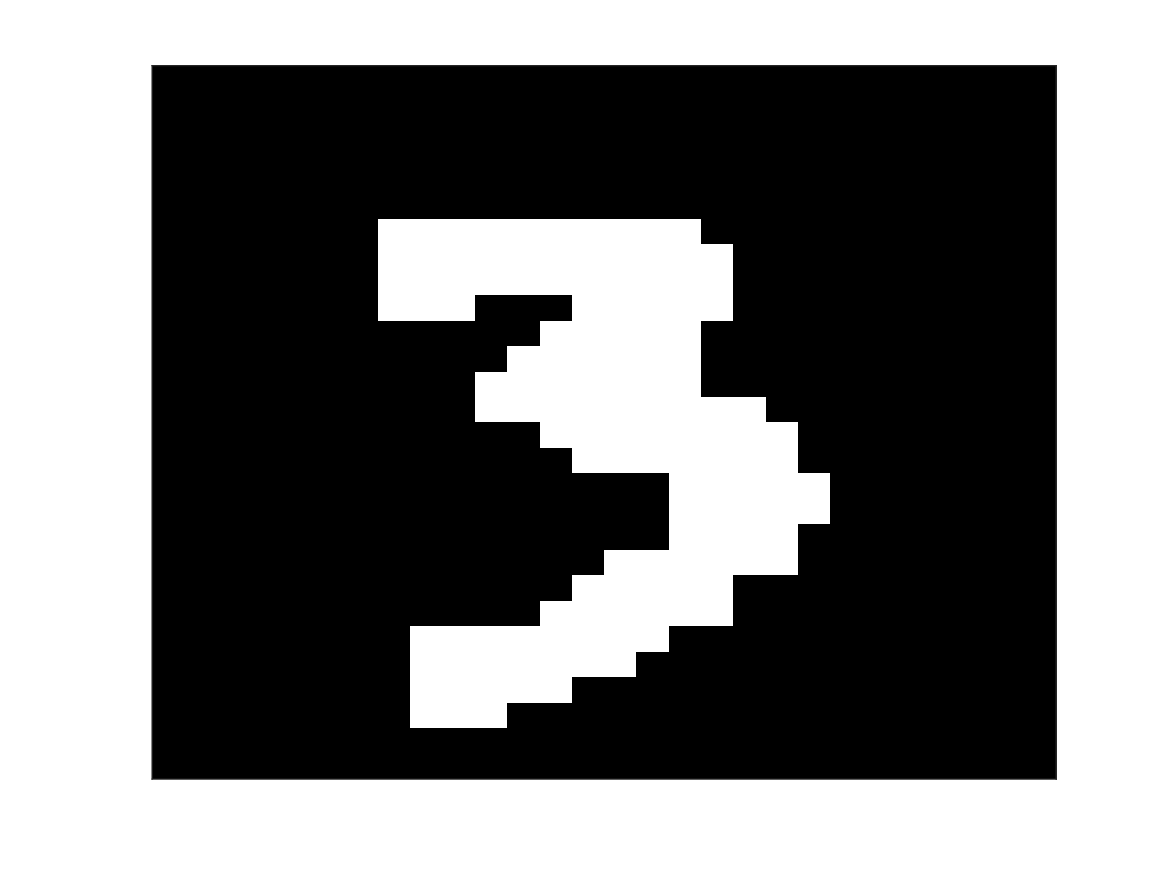}
    \includegraphics[width=0.19\linewidth]{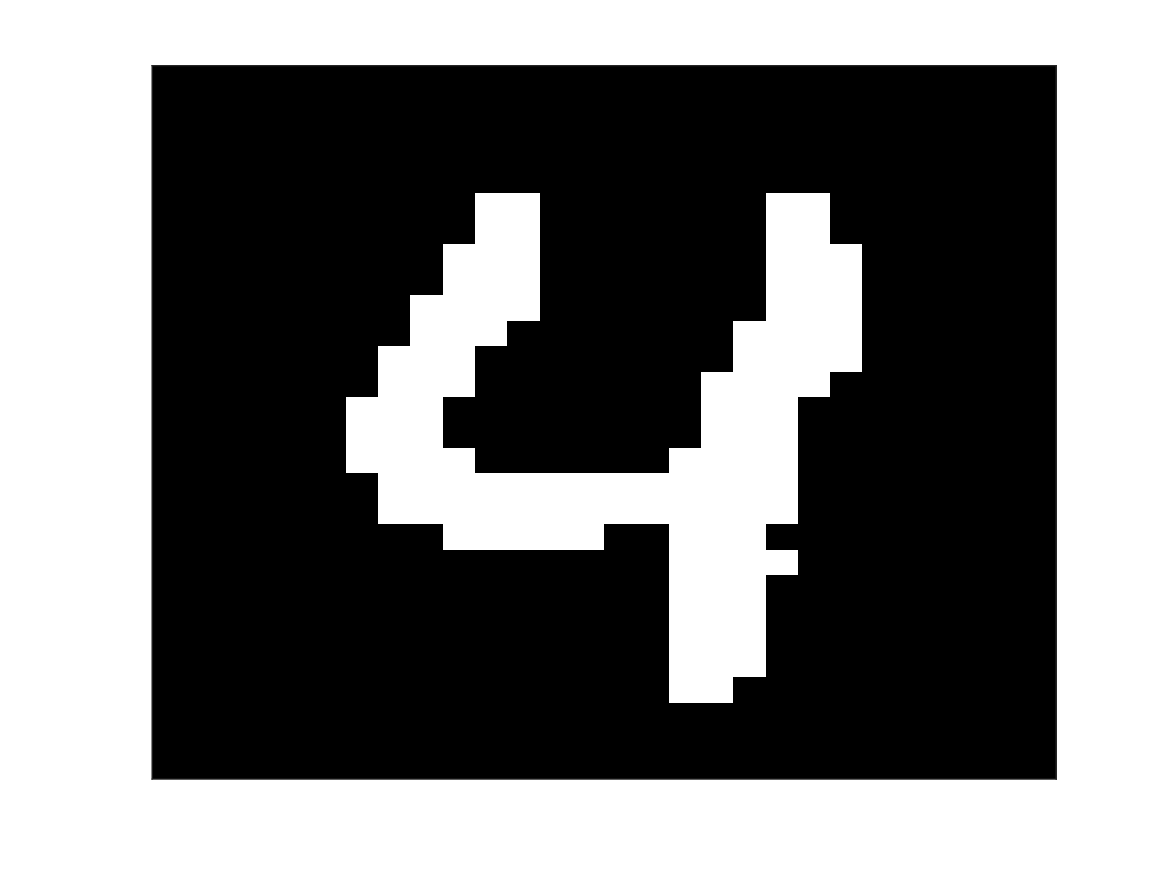}
    
    \includegraphics[width=0.19\linewidth]{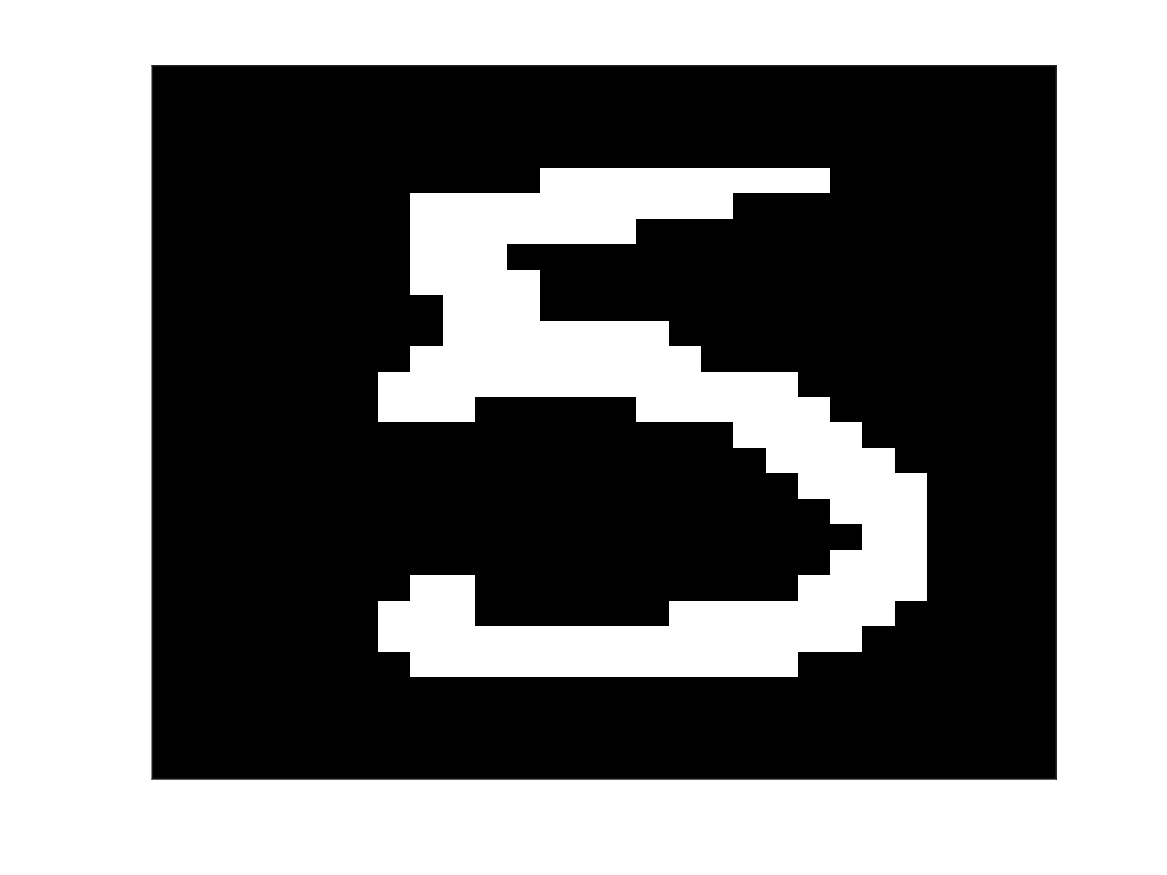}
    \includegraphics[width=0.19\linewidth]{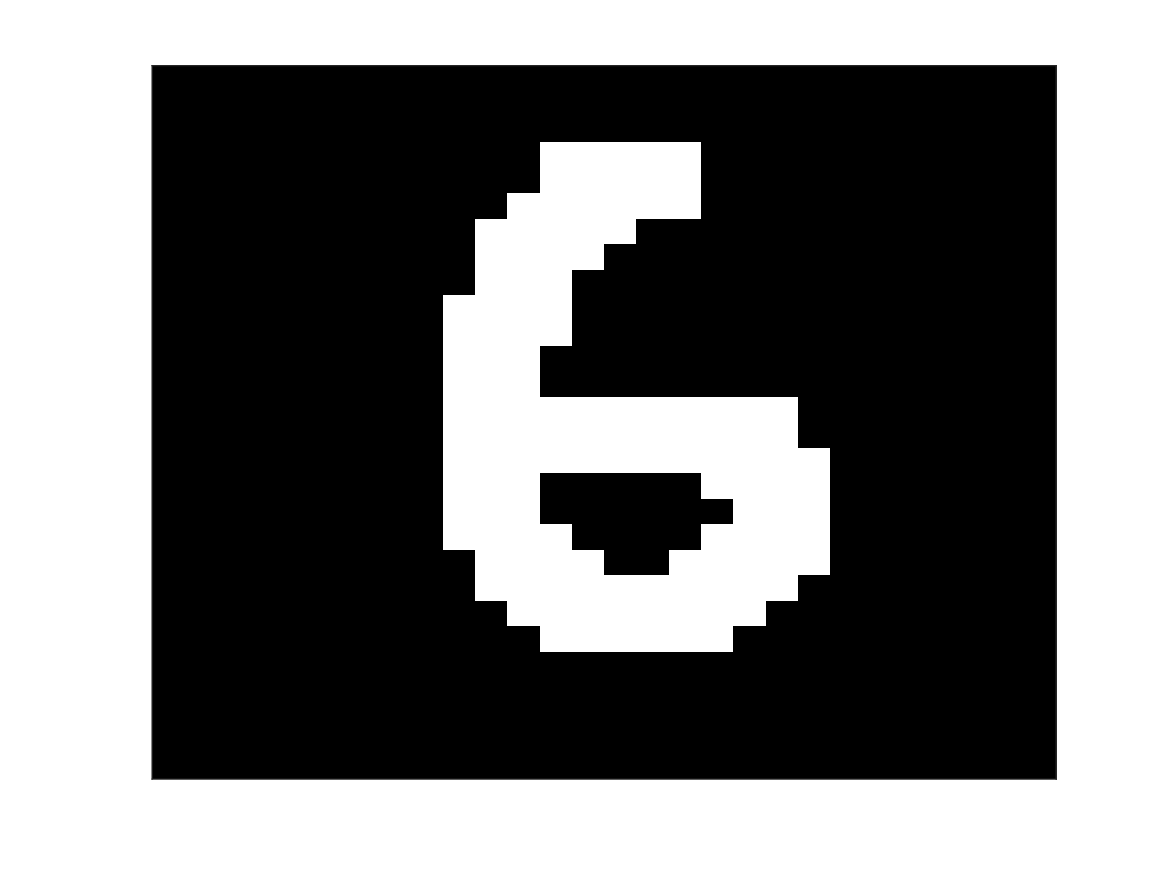}
    \includegraphics[width=0.19\linewidth]{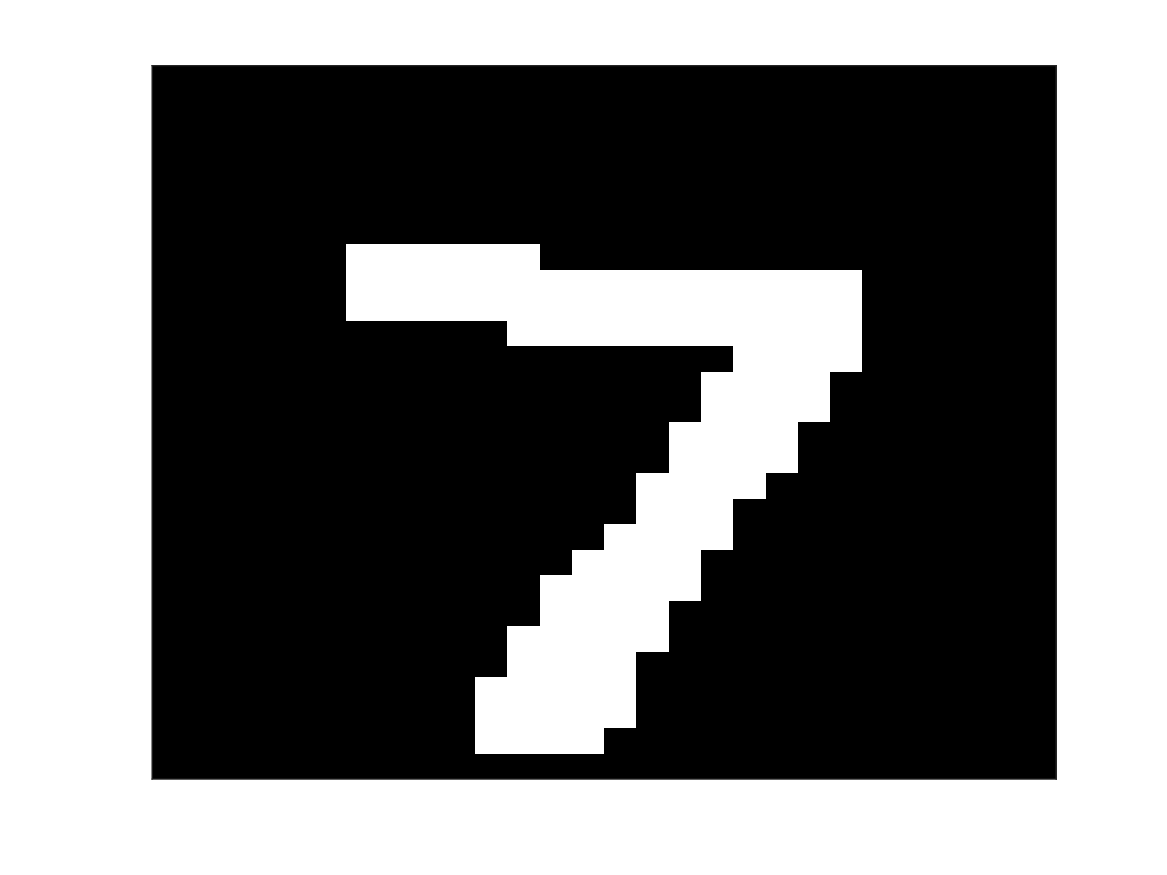}
    \includegraphics[width=0.19\linewidth]{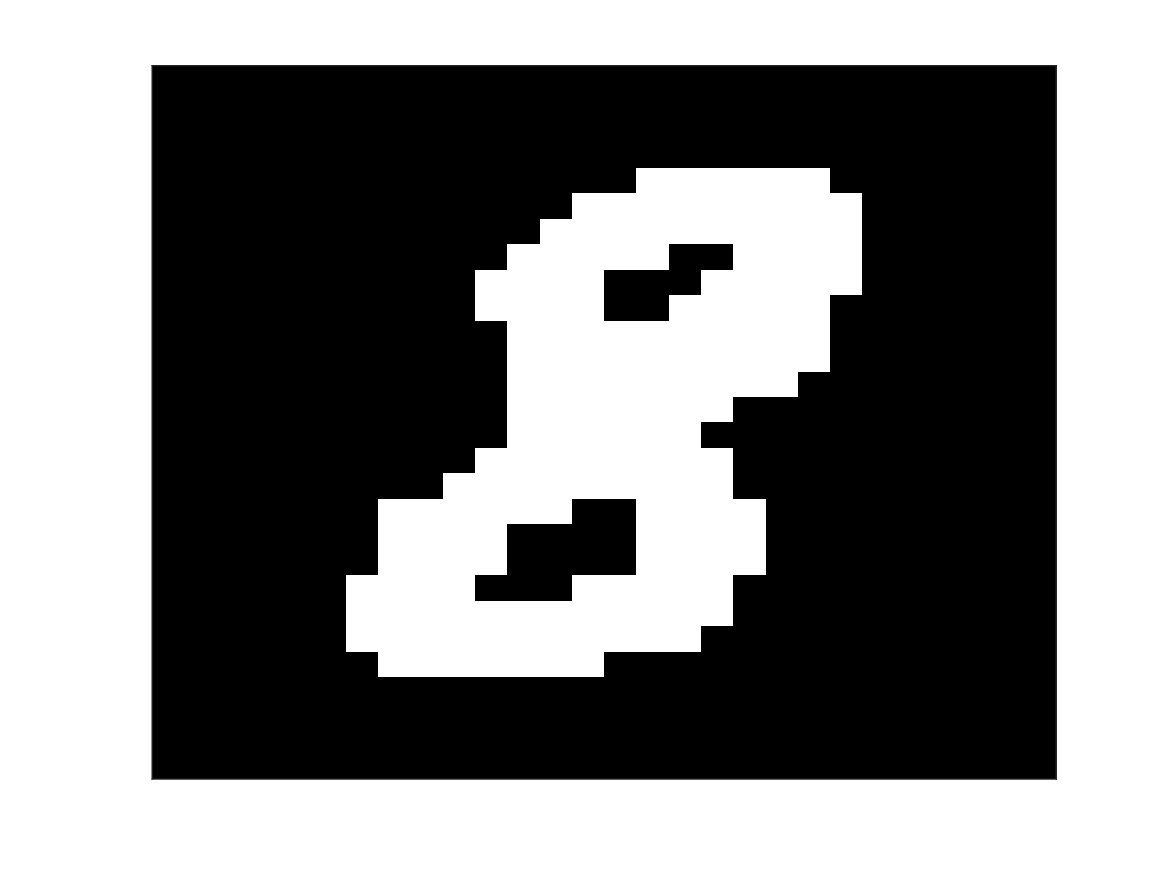}
    \includegraphics[width=0.19\linewidth]{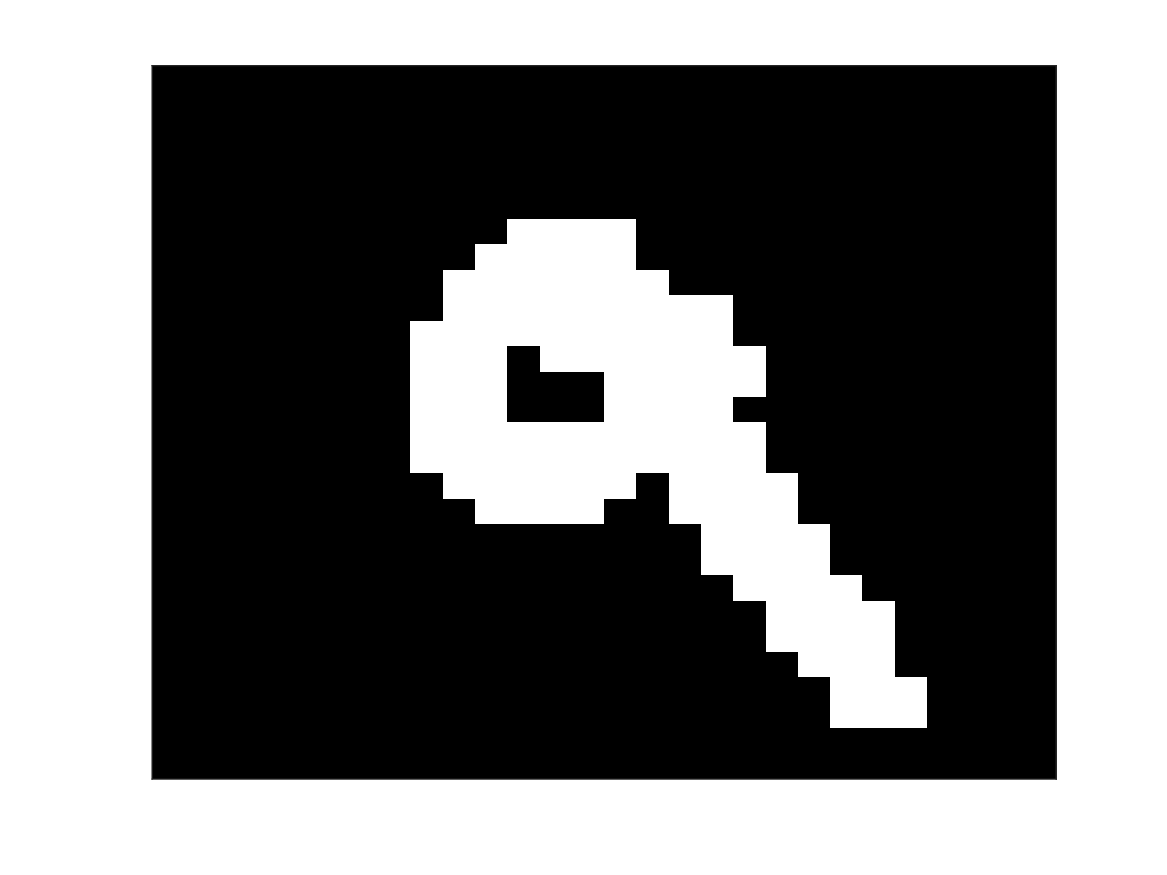}
    \caption{Sample images from the MNIST dataset.}
    \label{fig:mnist_sample}
\end{figure}

In this section we apply ENS-t-SNE to visualize the 
MNIST handwritten digit database~\cite{lecun-mnisthandwrittendigit-2010}.
The dataset contains $70,000$ handwritten digits in greyscale, each of sizes $28 \times 28$; see examples from the dataset in Figure~\ref{fig:mnist_sample}.
The standard way of applying  machine learning algorithms to the MNIST dataset is to vectorize the matrices corresponding to the greyscale pixel values of each digit and obtain a vector of size $28 \times 28 = 784$. Thus, each instance of the dataset can be viewed as a single datapoint in $\mathbb{R}^{784}$, with the idea being that points corresponding to the same digit should be close to each other. Dimensionality reduction techniques that aim to capture global distances (e.g., Principal Component Analysis  and Multi-Dimensional Scaling) are known to perform poorly when embedding such data into 2D or 3D. Non-linear dimensionality reduction techniques that focus on local neighborhoods, such as t-SNE and UMAP, perform much better.

\begin{figure} [h!]
    \centering
    \includegraphics[width=0.19\linewidth]{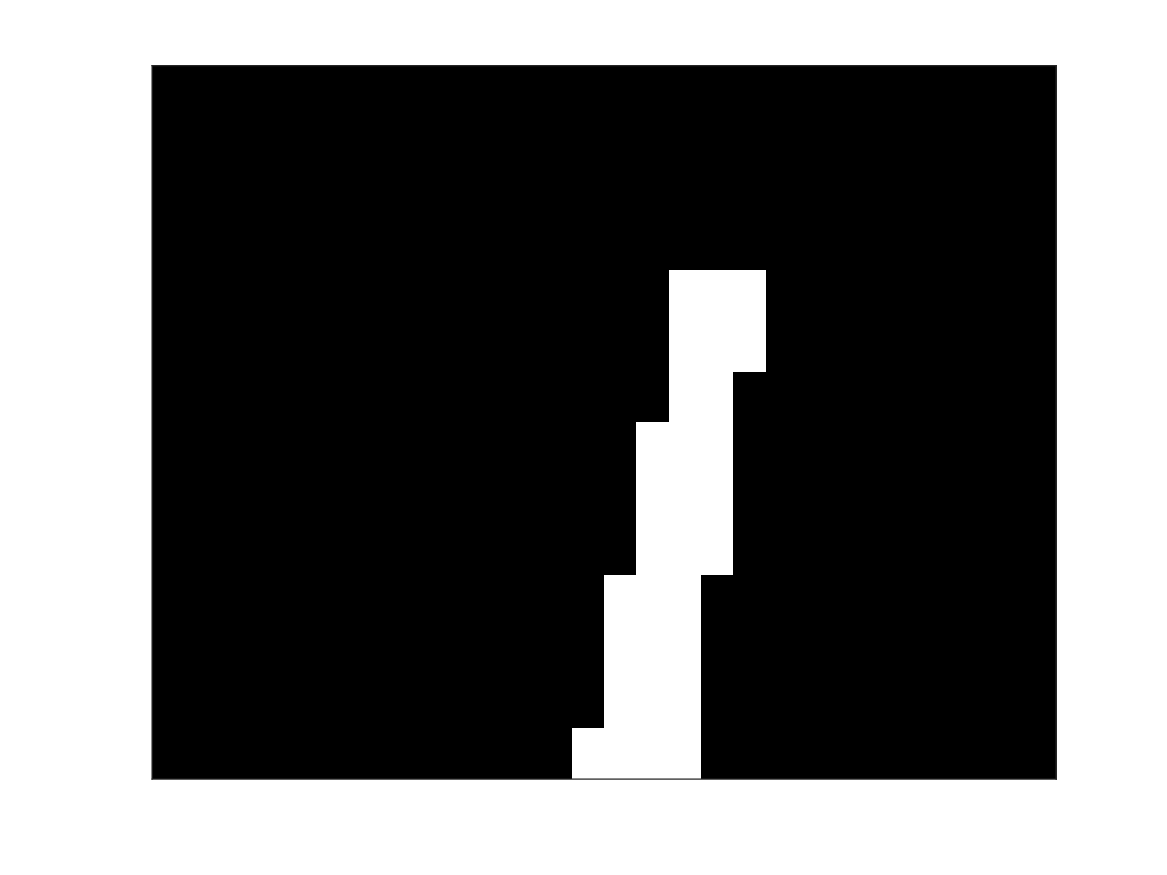}
    \includegraphics[width=0.19\linewidth]{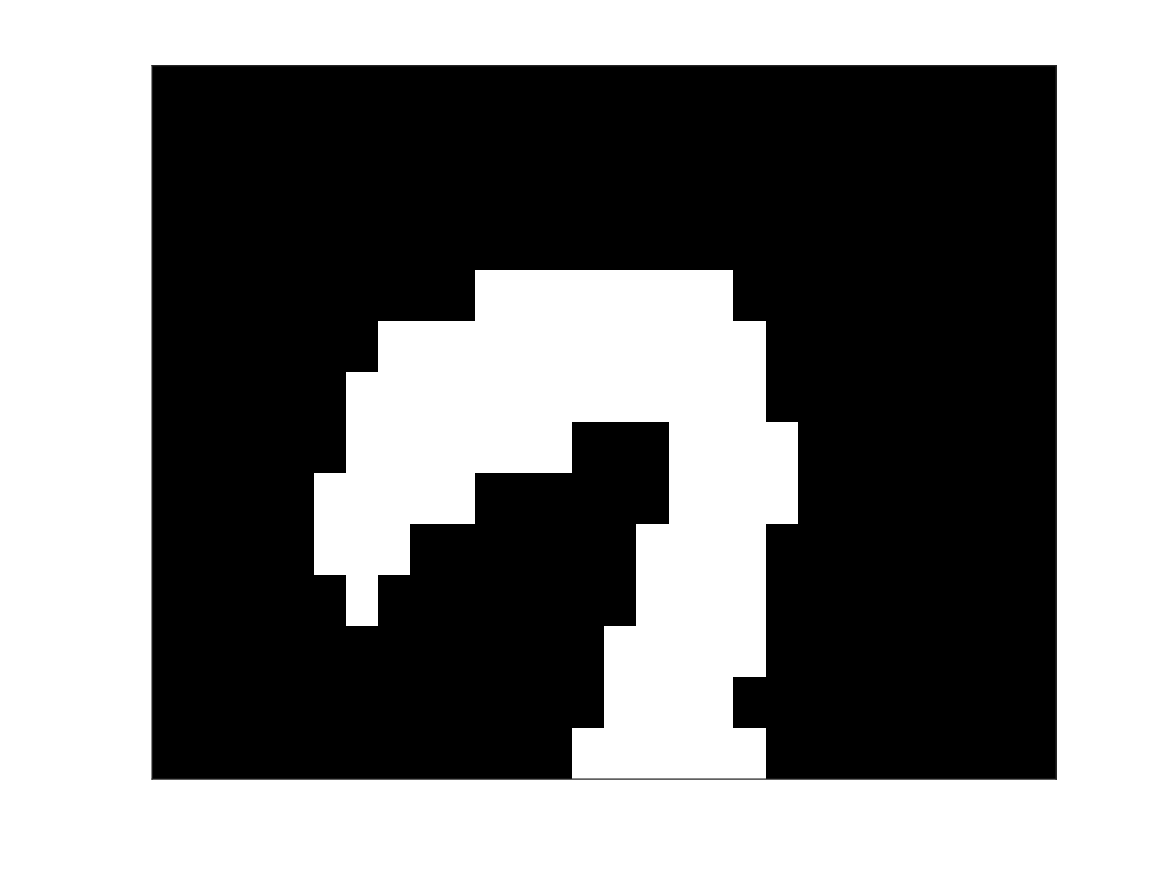}
    \includegraphics[width=0.19\linewidth]{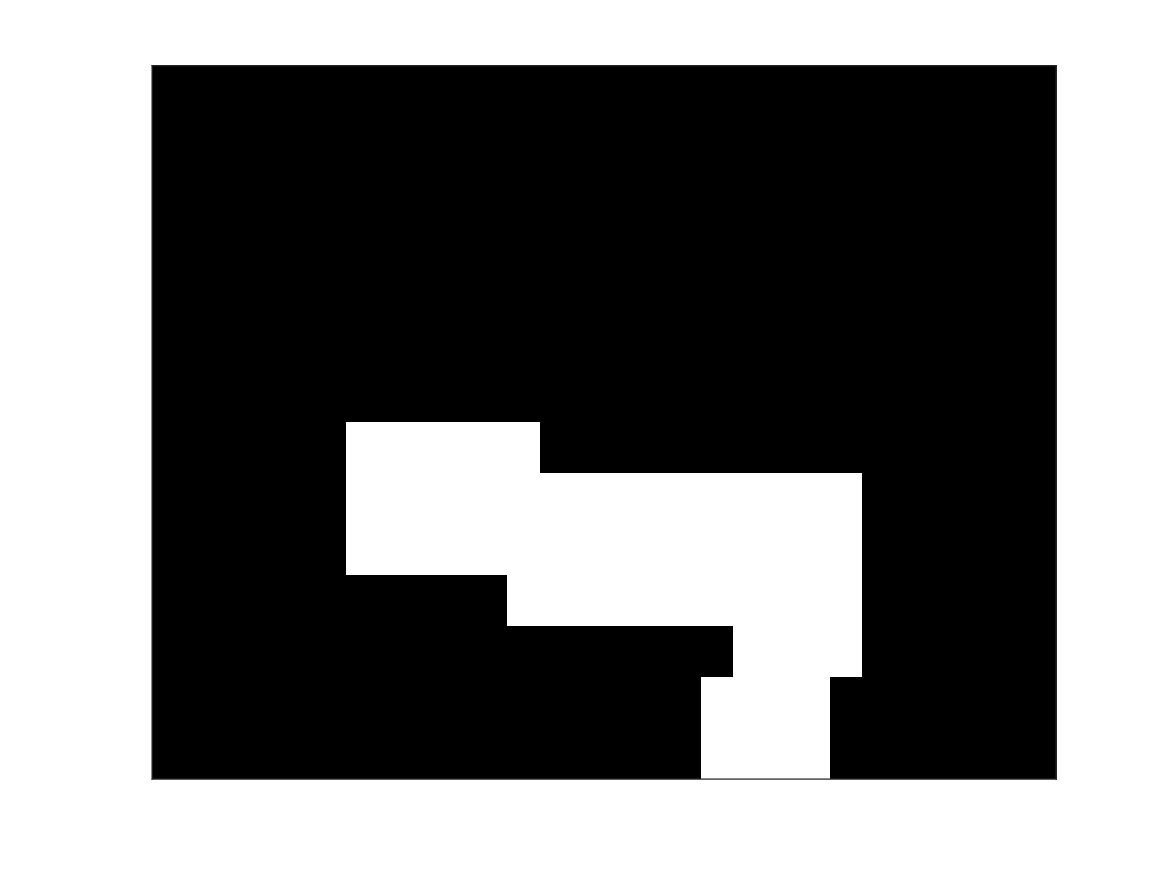}

    \includegraphics[width=0.19\linewidth]{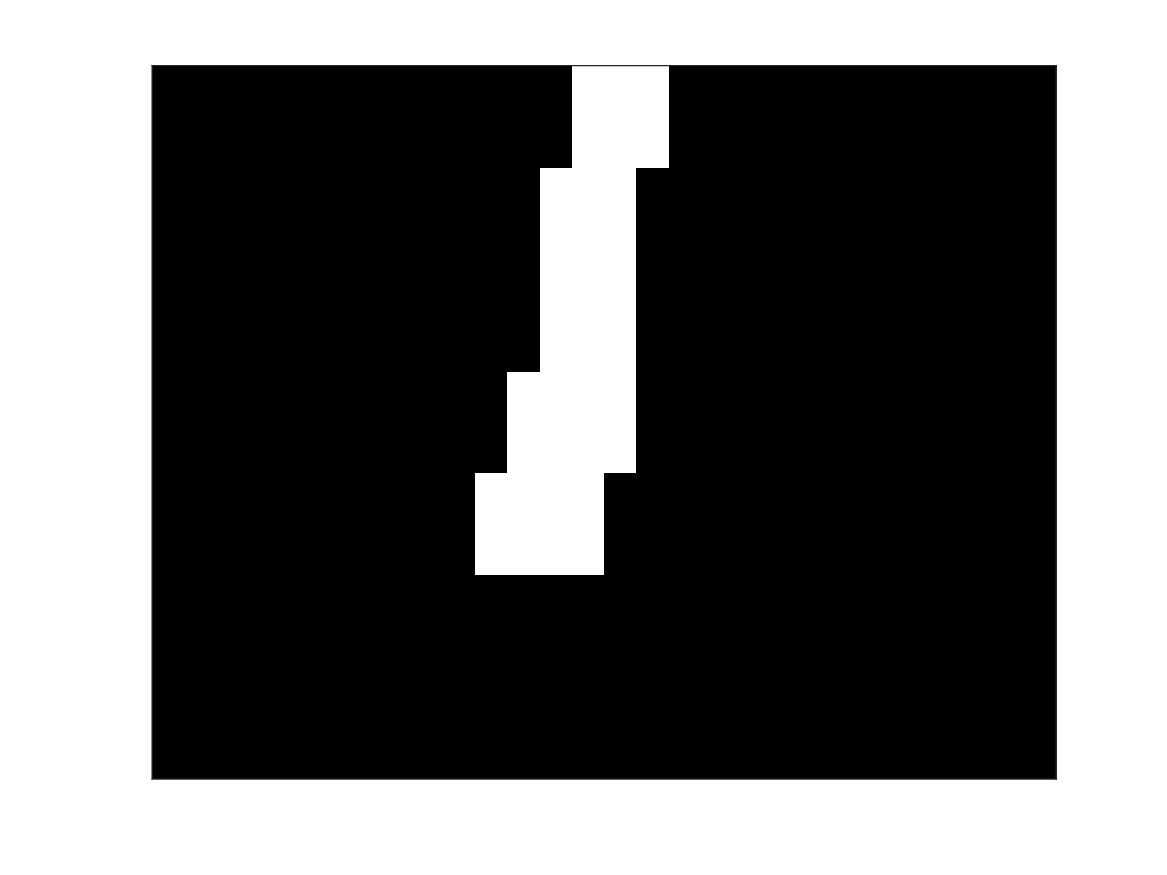}
    \includegraphics[width=0.19\linewidth]{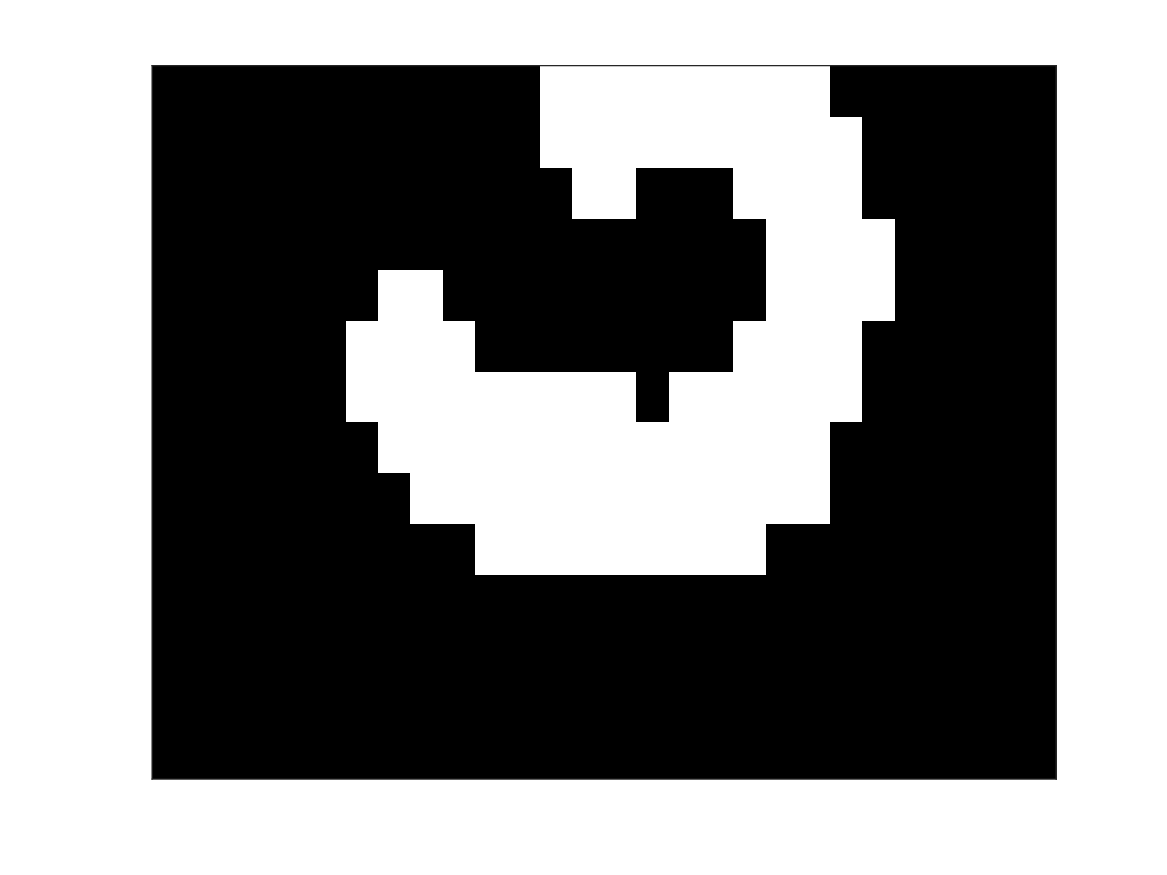}
    \includegraphics[width=0.19\linewidth]{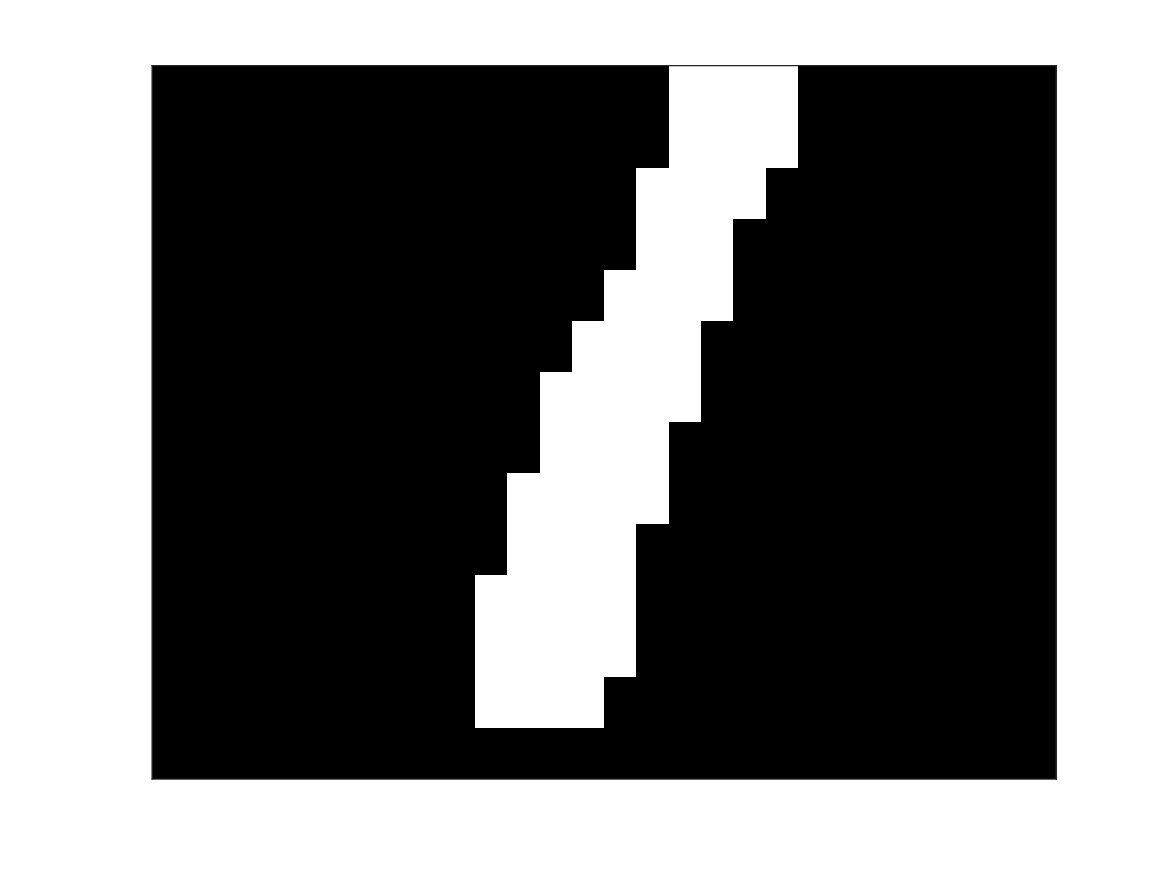}
    
    \caption{ MNIST dataset digits split into top and bottom parts.}
    \label{fig:mnist_top_bottom}
\end{figure}

We apply ENS-t-SNE to embed the MNIST dataset (and similar datasets which require local neighborhood preservation) in 3D to capture the clusters corresponding to each digit. 
The idea is to define multiple distance/similarity measures between pairs of datapoints that would capture different properties/characteristics. As a simple initial example, each image can be divided into two parts: top and bottom; see Figure~\ref{fig:mnist_top_bottom}. 
We then compute the Frobenius norm, $\Vert A \Vert_F$ between the matrices corresponding to the greyscale pixel values of each image. For a given matrix $A$, $\Vert A \Vert_F$ is defined as the square root of the sum of the
squares of its elements and can be viewed as the vector $L_2$ norm of the vector of all elements of the matrix.
Formally, for a matrix $A = (a_{i, j}))_{i = 1, j = 1}^{m, n}$, its Frobenius norm is defined as:
$$\Vert A \Vert_F = \sqrt{\sum_{i = 1}^m \sum_{j = 1}^n a^2_{ij}}.$$
We now have two different sets of distance matrices: one corresponds to the top part of each image and the other corresponds to the bottom part of each image.

The idea behind this experiment is to apply ENS-t-SNE to a dataset with different perspectives, where some points are close to each other (in the same cluster) in one perspective, but are far from each other (in different clusters) in the other perspective. The ENS-t-SNE algorithm should place the points in 3D so that the desired properties are satisfied  in the corresponding perspectives.

\begin{figure*}[!]
    \centering
    \includegraphics[width=0.33\linewidth]{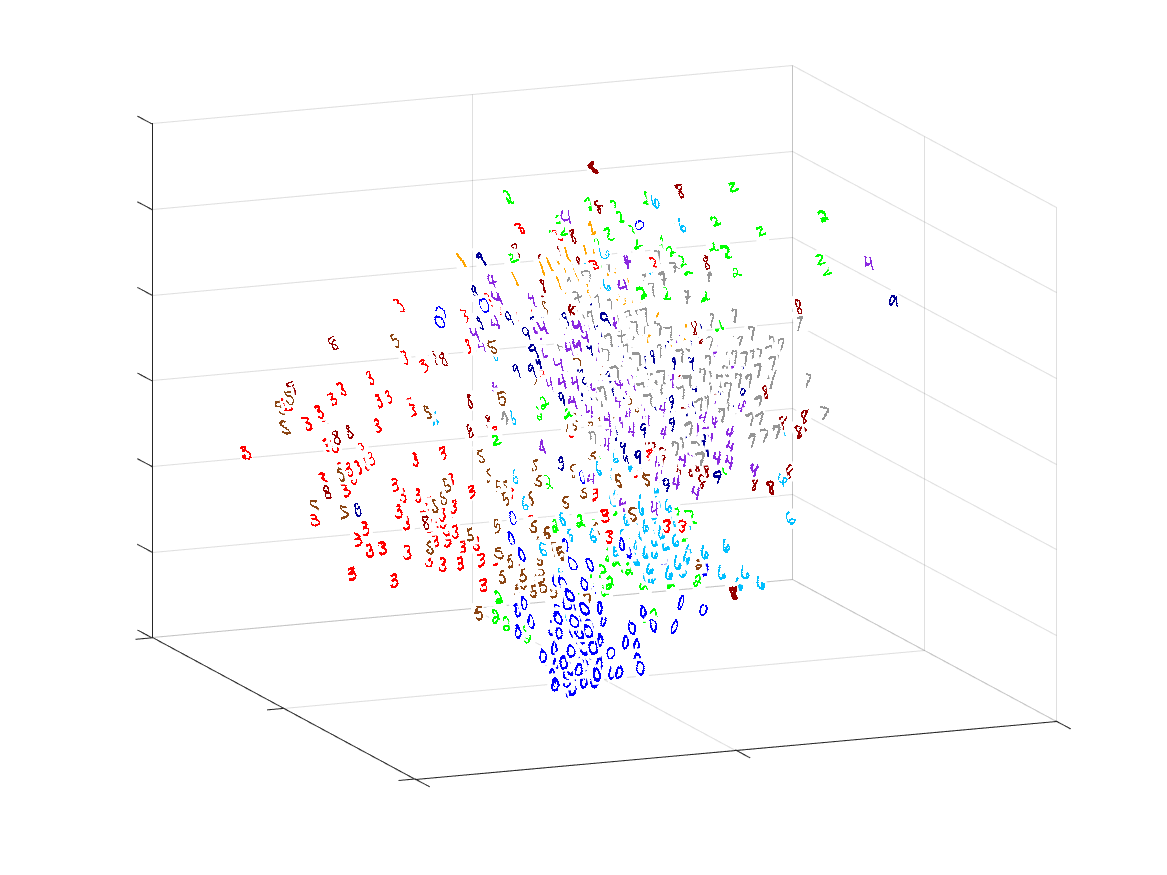}
    \includegraphics[width=0.33\linewidth]{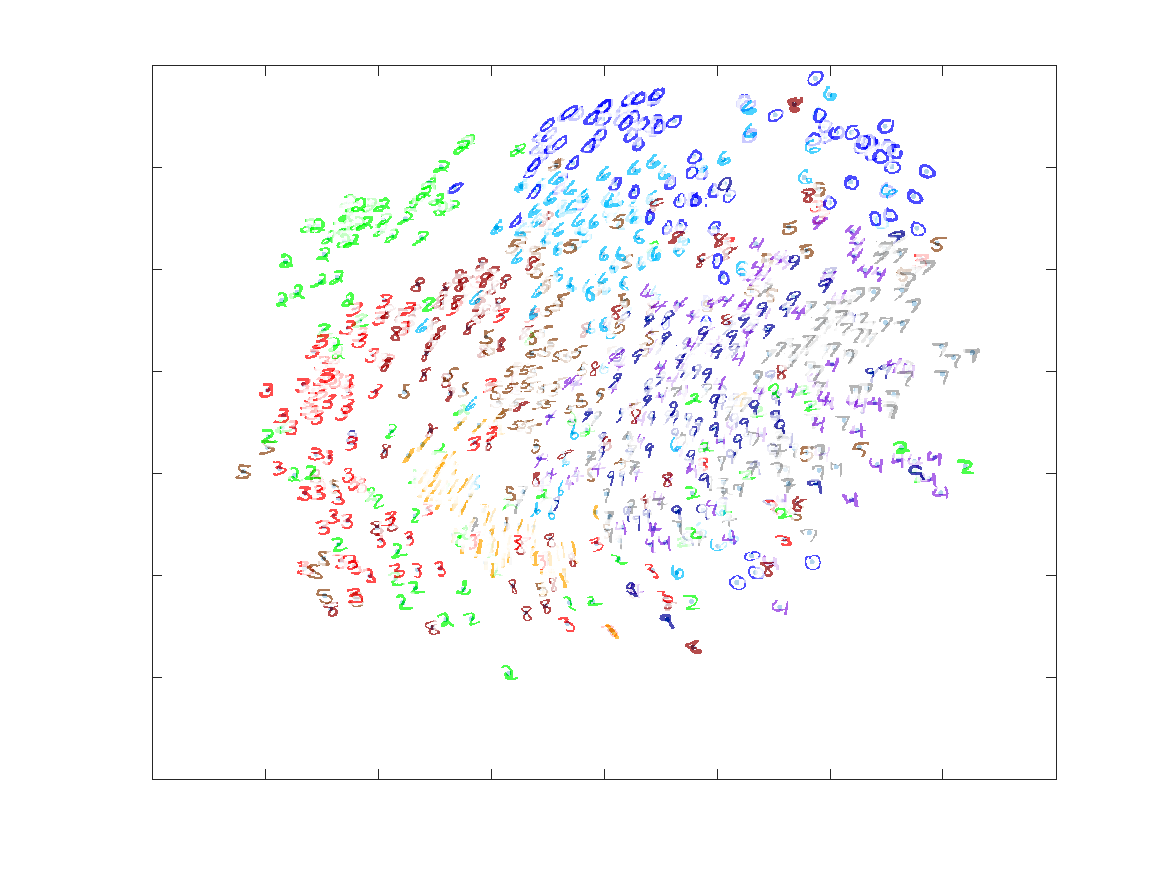}
    \includegraphics[width=0.33\linewidth]{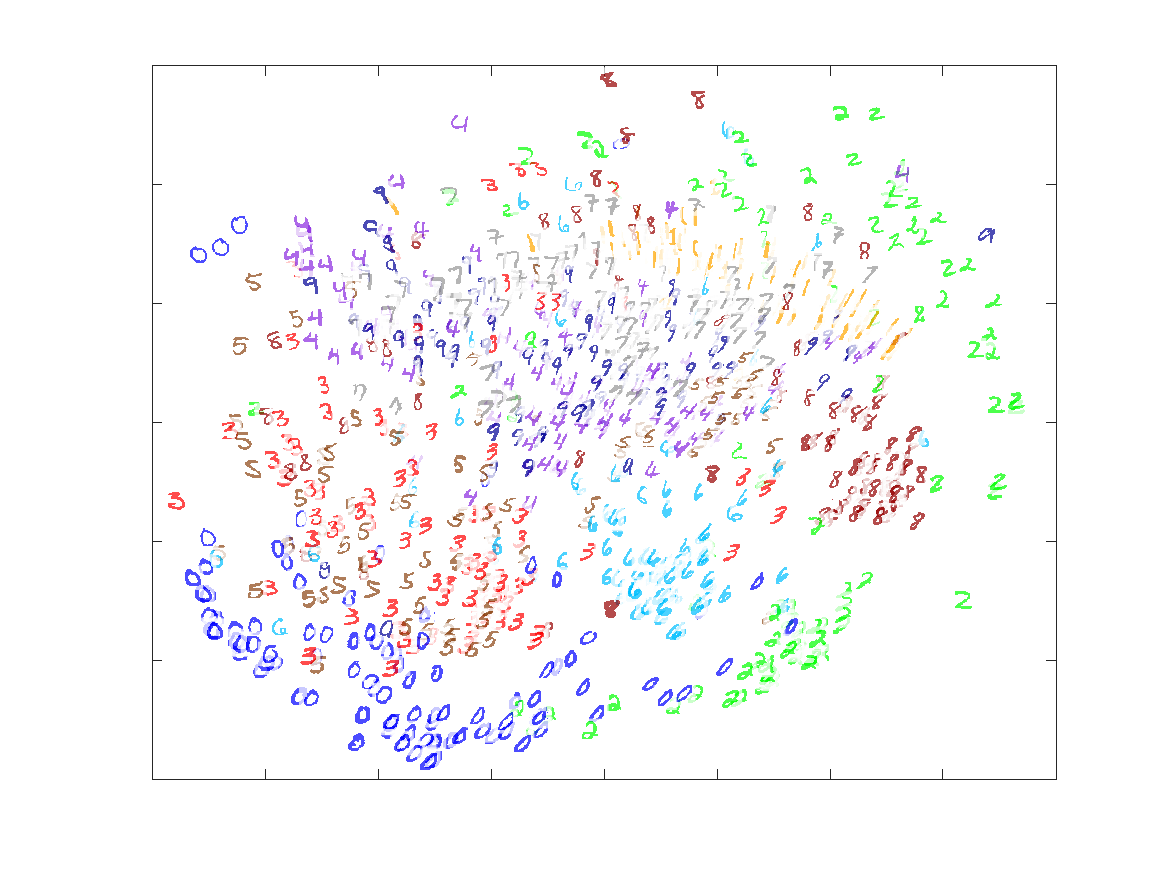}
    \caption{Application of ENS-t-SNE to the 1000 instances from the MNIST dataset with perplexity value 500. The left subfigure demonstrates the 3D embedding computed by ENS-t-SNE. The middle subfigure demonstrates the projection of the 3D embedding onto the view that corresponds to the distance matrix constructed from the upper parts of the digits. The right subfigure demonstrates the projection of the 3D embedding onto the view that corresponds to the distance matrix constructed from the lower parts of the digits.}
    \label{fig:mnist_vis_perp_500}
\end{figure*}

\begin{figure} [h!]
    \centering
    \includegraphics[width=0.49\linewidth]{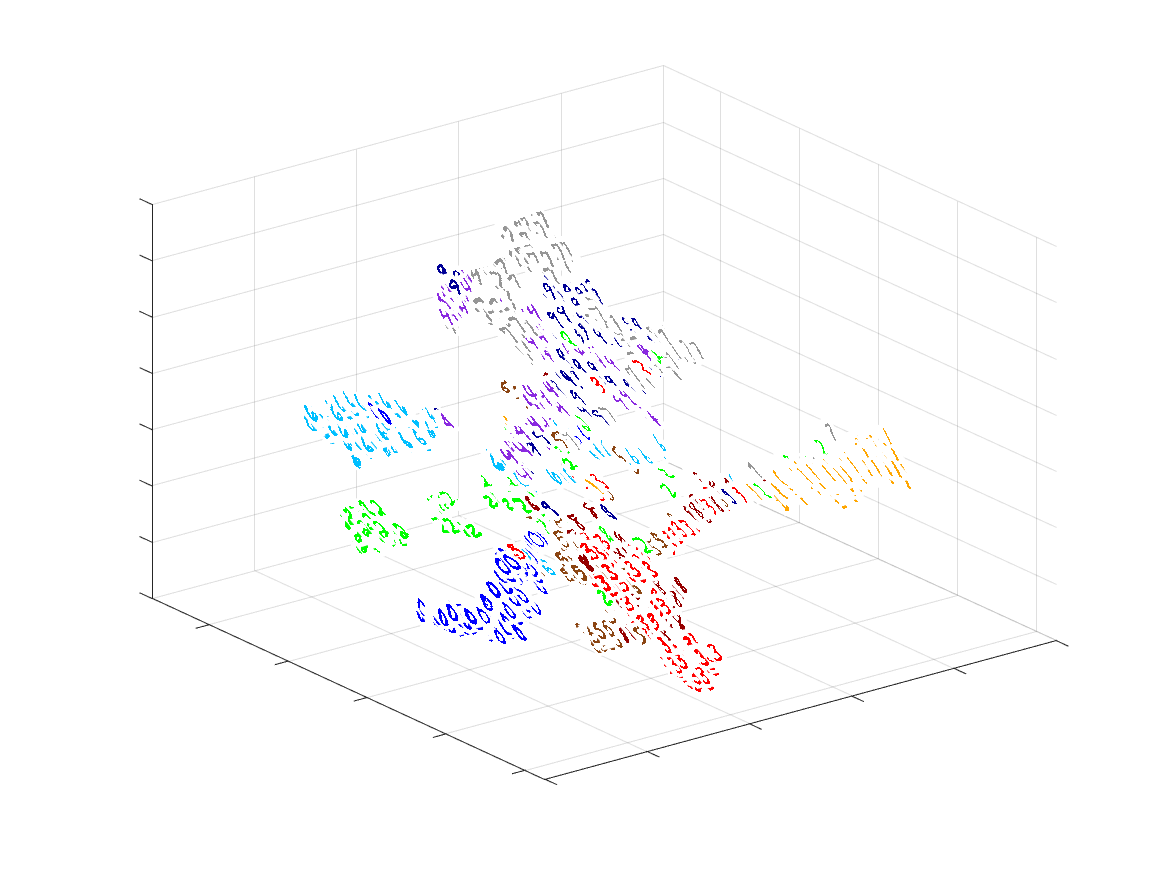}
    \includegraphics[width=0.49\linewidth]{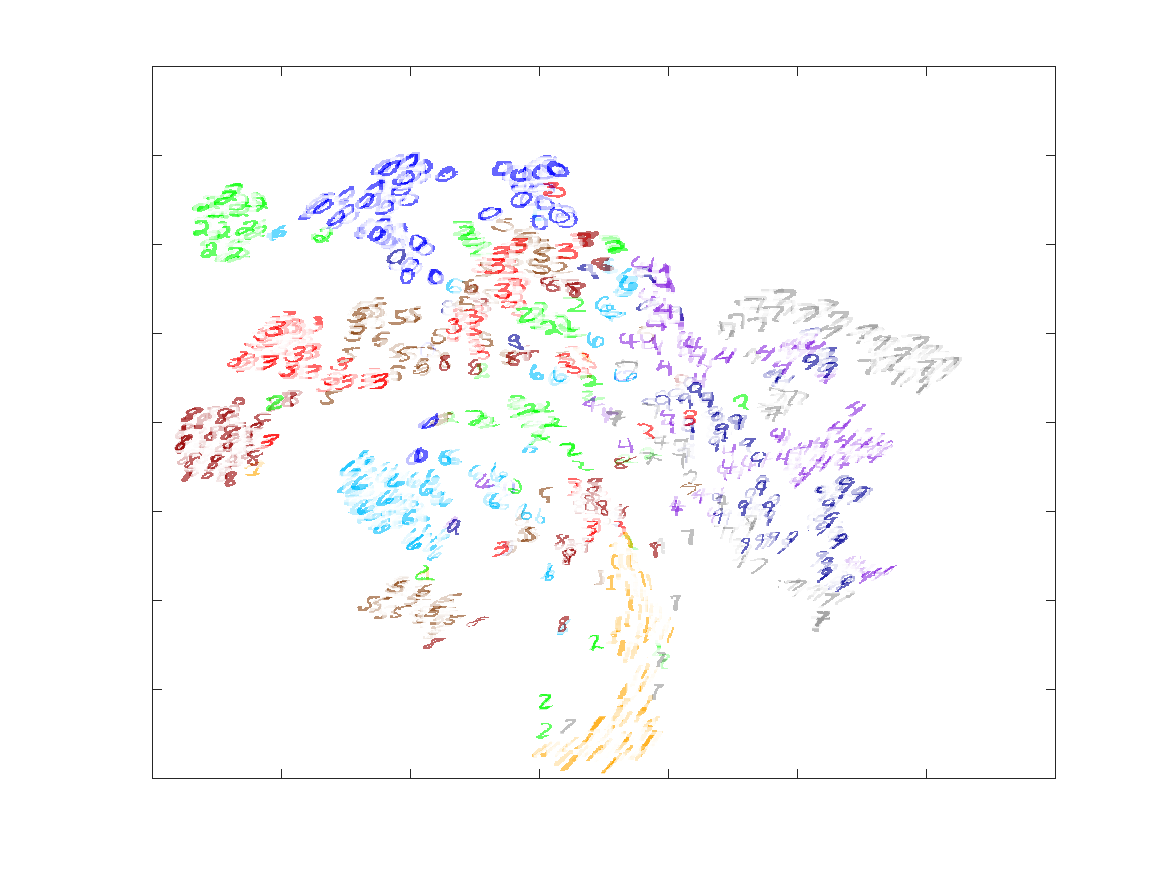}
    \caption{3D (top) and 2D (bottom) embedding obtained by t-SNE of a subset with 1000 datapoints from the MNIST dataset.}
    \label{fig:mnist_tsne}
\end{figure}

The distance matrices for the top and bottom parts of all datapoints are the inputs to the ENS-t-SNE algorithm. The results are different than those from the standard t-SNE applied on the original datapoints, as illustrated in Figures~\ref{fig:mnist_tsne} and \ref{fig:mnist_vis_perp_500}.
The bottom view (the third subfigure of Fig.~\ref{fig:mnist_vis_perp_500}) shows that digits 1, 7 and 4 (sometimes 9, depending on the handwriting) are close to each other. This is expected since for most of these digits the bottom parts are nearly straight line segments oriented roughly the same way. However, since there is a significant difference in the top parts, the clusters are separated in 3D. 
Similarly, the pair of digits 3 and 5 are close in the bottom view and the digits 8 and 9 are close in the top view. 

Comparing the standard t-SNE embedding in Figure~\ref{fig:mnist_tsne}, we can see that the ENS-t-SNE embedding managed to avoid some ``errors". For example in Figure~\ref{fig:mnist_tsne}, there are some 0s that appear within the cluster of 6s while that is not the case for ENS-t-SNE; see the left subfigure of Fig.~\ref{fig:mnist_vis_perp_500}.

In addition to the visual comparison between the MNIST dataset embedding obtained from ENS-t-SNE and the standard t-SNE, we also evaluate them using cluster accuracy~\cite[Section 10.6.3]{han2011data}. 
Specifically, given an embedding (from ENS-t-SNE or t-SNE) we apply $k$-means clustering with 10 clusters and compare to the ground truth 10 clusters. 
For each embedding, finding the correct cluster labels is based on considering all re-orderings of the $k$-means cluster labels and selecting the one that best matches the ground truth. 
After that, count the number of points that are correctly clustered and normalize it by the total number of points.
The results suggest that ENE-t-SNE does as well or better than standard t-SNE:  0.539 for ENS-t-SNE in Fig.~\ref{fig:mnist_vis_perp_500} vs. 0.478 for standard t-SNE in Fig.~\ref{fig:mnist_tsne} (with higher scores corresponding to higher accuracy).
The experiments were run for randomly chosen 1000 instances of MNIST dataset and averaged over 5 runs. In general, the accuracy is comparable, with a slight advantage to ENS-t-SNE.


\section{Visualizing the Effect of Perplexity}
\label{sec:perplexity_effect}

The choice of perplexity parameter in t-SNE greatly affects the quality of achieved embedding; see~\cite{wattenberg2016use}. Usually, smaller perplexity parameters produce visualizations that better reflect local distances between samples, however, when the perplexity is very small the algorithm fails to find a sufficiently good solution. Thus, finding an appropriate value of perplexity for which t-SNE would find the best possible embedding has been of research interest~\cite{cao2017automatic}.

ENS-t-SNE can be used to visualize the differences between various perplexity parameters on the same given distance matrix. In practice, ENS-t-SNE seems to overcome the issue of not producing good results for smaller/larger-than-ideal perplexity value, as long as one of the perplexity values passed into ENS-t-SNE is sufficiently good. Specifically, we apply ENS-t-SNE to the same set of distance matrices but with different perplexity values. The goal is to find an embedding of the dataset in 3D so that on the different projections it solves the t-SNE optimization with different perplexity values (but with the same distance matrix).

The problem formulation is as follows. Let $D$ be an $N \times N$ distance matrix and $\mathrm{perp}_1, \mathrm{perp}_2, \dots, \mathrm{perp}_M > 0$ be a list of perplexity parameters of interest. We wish to minimize the following cost function
\begin{equation}
    \label{eq:cost_different_perplexity}
    \widetilde{C}(X,\Pi;\mathrm{perp}_1, \mathrm{perp}_2, \dots, \mathrm{perp}_M) = \sum_{m=1}^M C(\Pi^m X; \mathrm{perp}_m)
\end{equation}
where $Y \mapsto C(Y; \mathrm{perp}_m)$ is the t-SNE cost function with perplexity parameter $\mathrm{perp}_m$.
Minimizing~\eqref{eq:cost_different_perplexity} is achieved using a momentum based batch or stochastic gradient descent.

\begin{figure*}[!ht]
    \centering
    \includegraphics[width=0.50\linewidth]{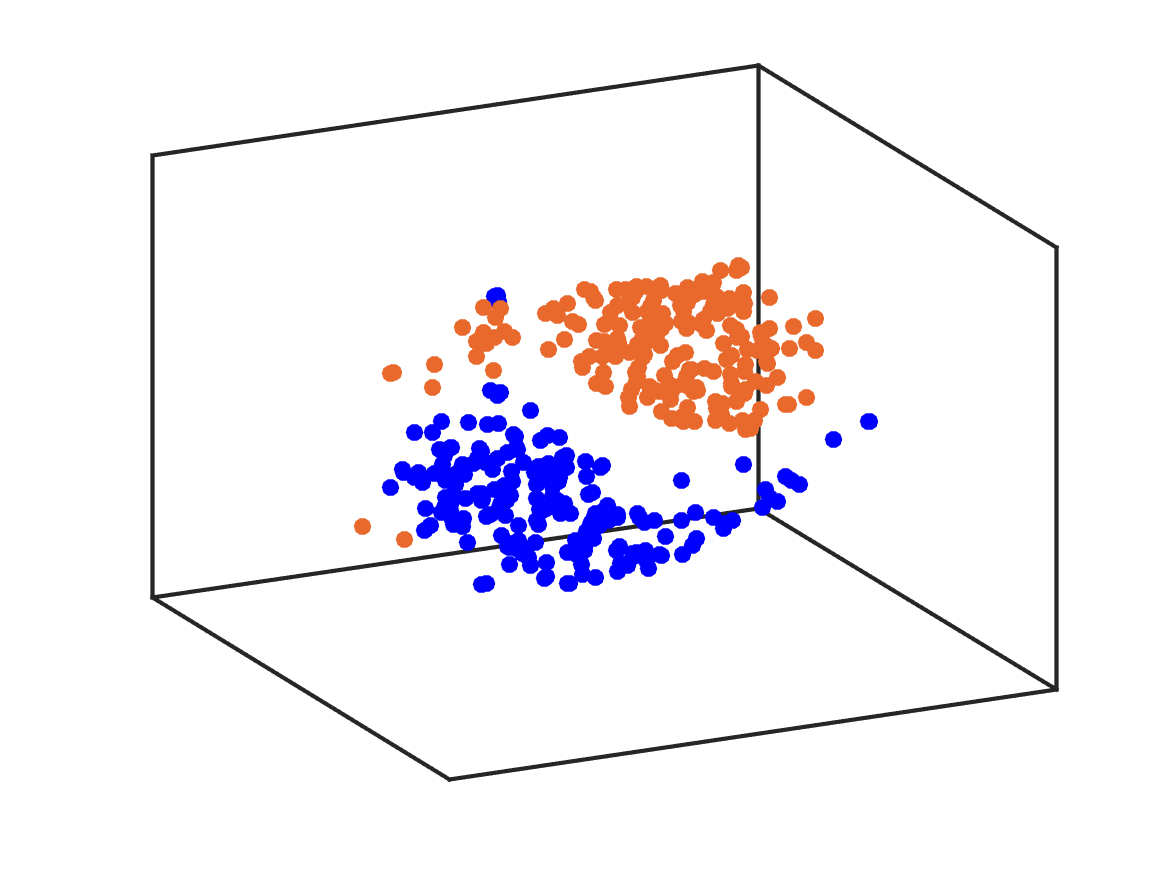}
    
    \includegraphics[width=0.35\linewidth]{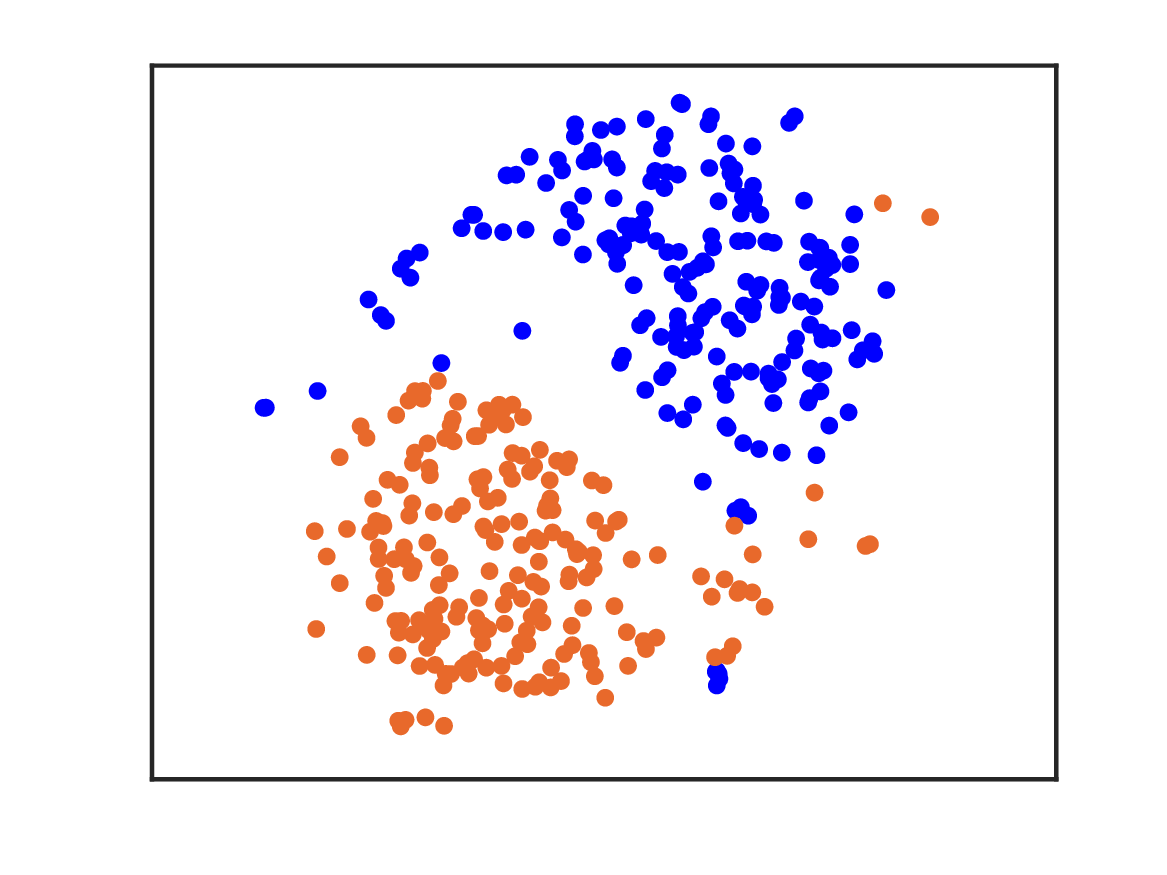}
    \includegraphics[width=0.35\linewidth]{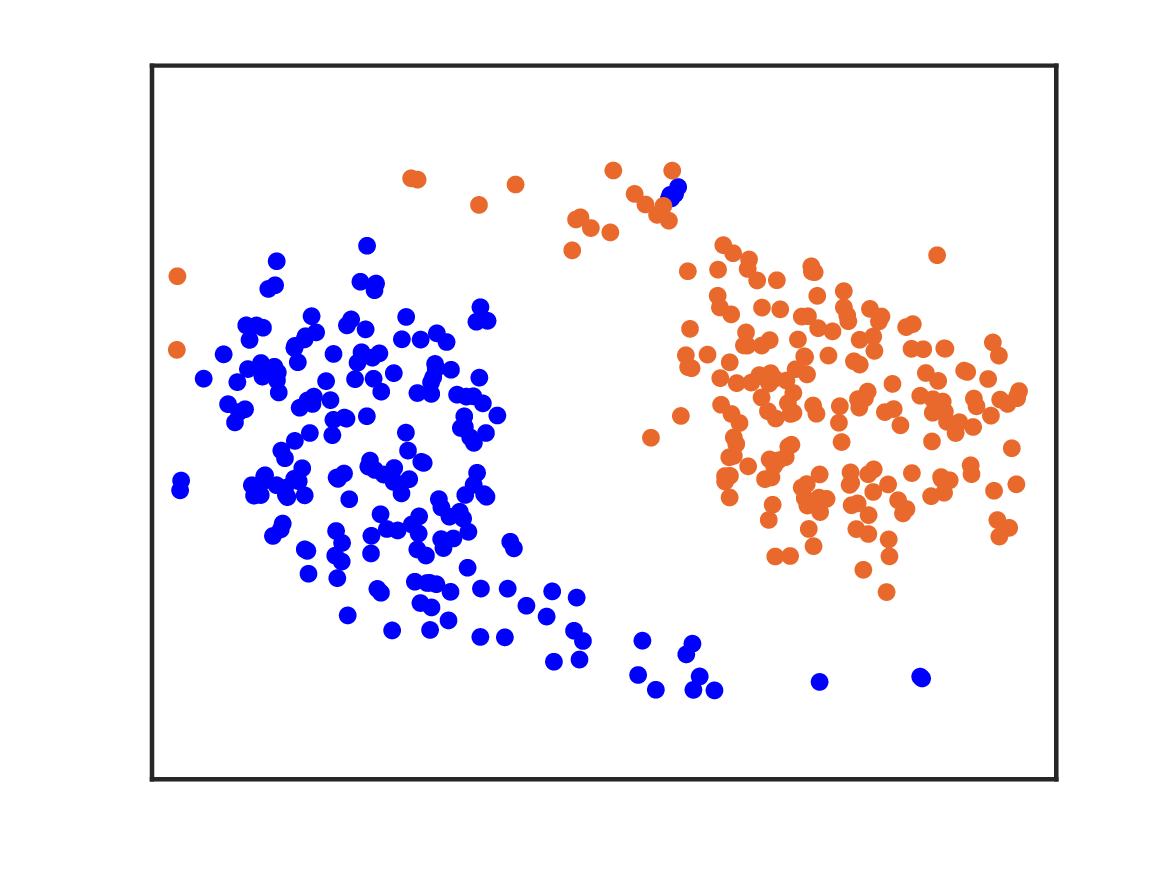}

    \includegraphics[width=0.35\linewidth]{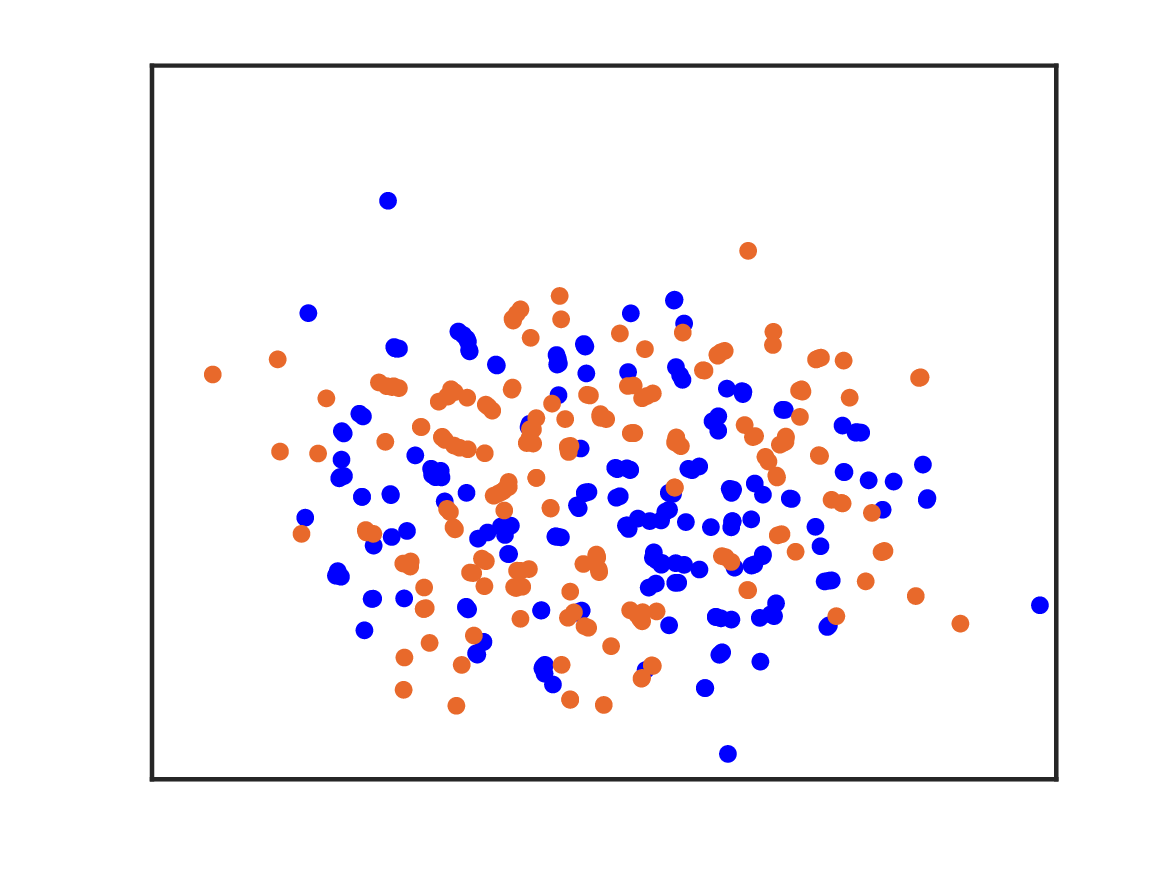}
    \includegraphics[width=0.35\linewidth]{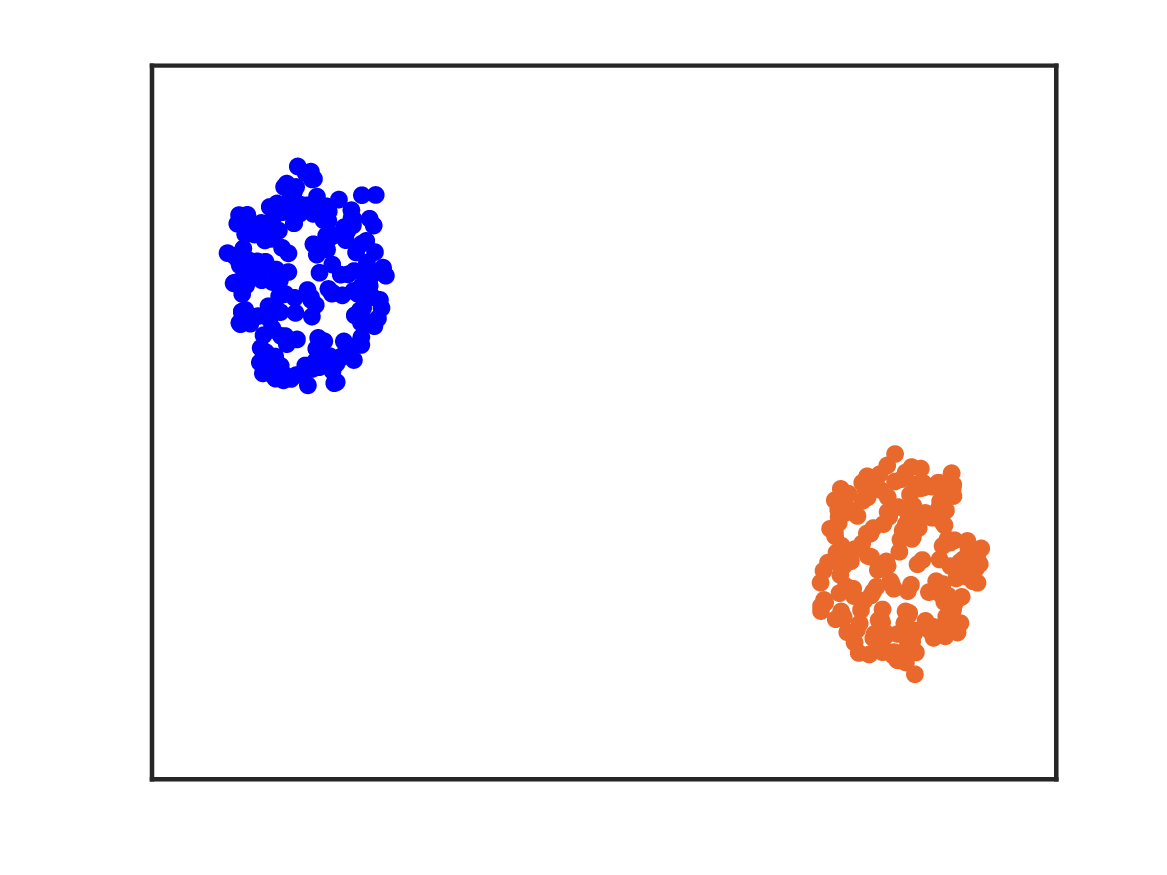}
    
    \caption{Comparison of the ENS-t-SNE visualization of a dataset with perplexities 3 and 100 and the corresponding t-SNE visualizations. The dataset contains 400 samples that form two clusters, as colored. The top image shows a glimpse of the 3D ENS-t-SNE embedding. The middle two images show the two projections of the 3D ENS-t-SNE embedding, the left corresponding to perplexity 3 and the right corresponding to perplexity 100. The bottom two images are t-SNE visualizations of the same distances using perplexity 3 (left) and perplexity 100 (right).}
    \label{fig:perplexity_test_2}
\end{figure*}

In Figure~\ref{fig:perplexity_test_2}, we show an application of ENS-t-SNE for a dataset that contains 2 clusters, constructed according to the model described in Section~\ref{sec:cluster_construction}, with $M = 1$ and with $N = 400$ datapoints, using perplexity parameters equal to 3 and 100. 
The 3D embedding for the corresponding computed distance matrix, shown in the first row of Figure~\ref{fig:perplexity_test_2},  and for the given two values of the perplexity parameter is the solution to problem~\eqref{eq:cost_different_perplexity}. 
The two figures in the second row of Figure~\ref{fig:perplexity_test_2} show the projections of the 3D embedding that best represent the perspectives of this data set with perplexities 3 and 100. 

As a way of comparison, we also compute the corresponding standard t-SNE 2D embeddings for the same distance matrix $D$ with perplexities 3 and 100 (the last row of Figure~\ref{fig:perplexity_test_2}). It is easy to see that standard t-SNE found the clusters when the perplexity was high and failed to find them when the perplexity was low, while ENS-t-SNE captured the clusters for both perplexity values. 
We note that whereas the images produced by ENS-t-SNE are projections of the same 3D embedding, the images produced by t-SNE are obtained  independently.

\begin{figure}
    \centering

    \includegraphics[width=0.45\linewidth]{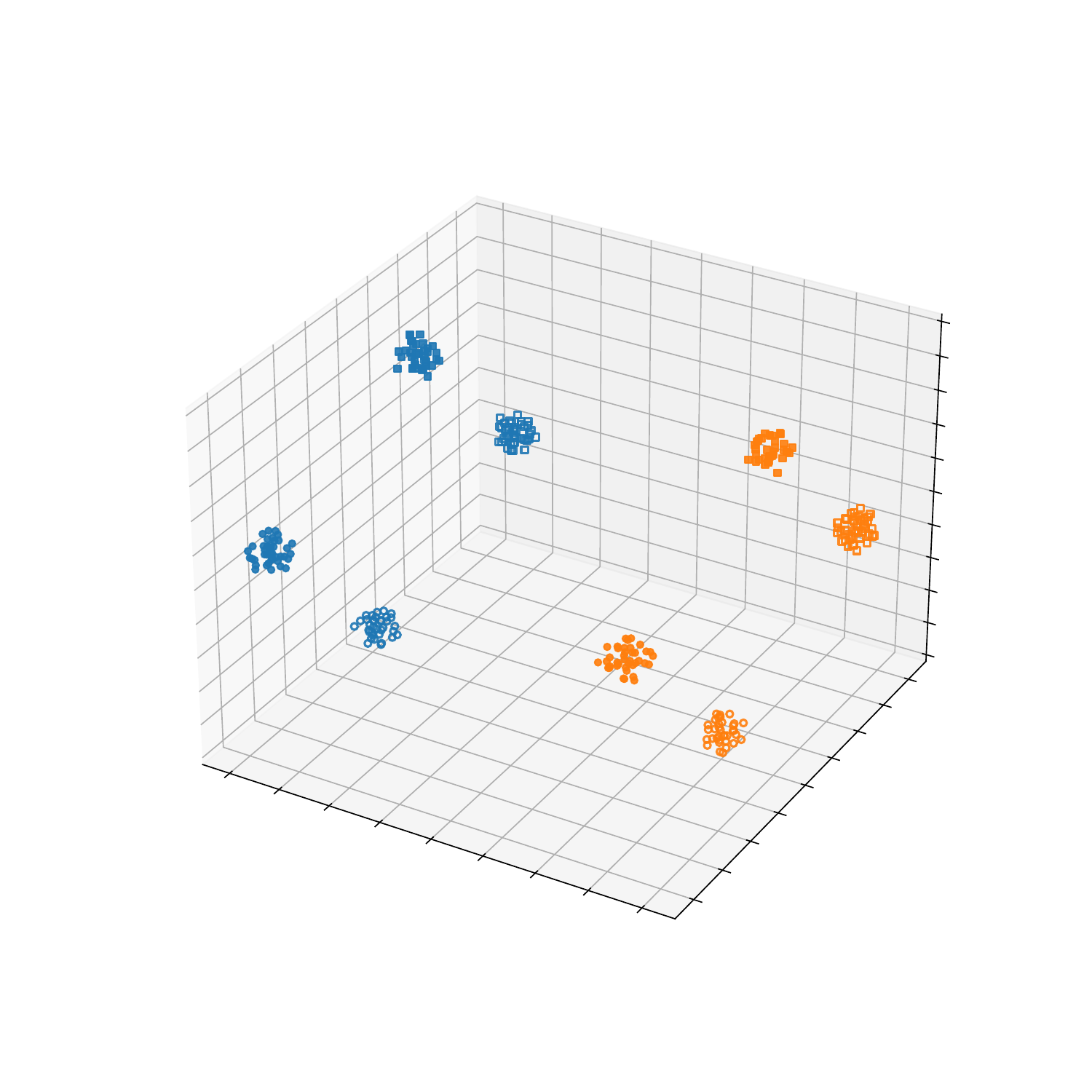}
    \includegraphics[width=0.45\linewidth]{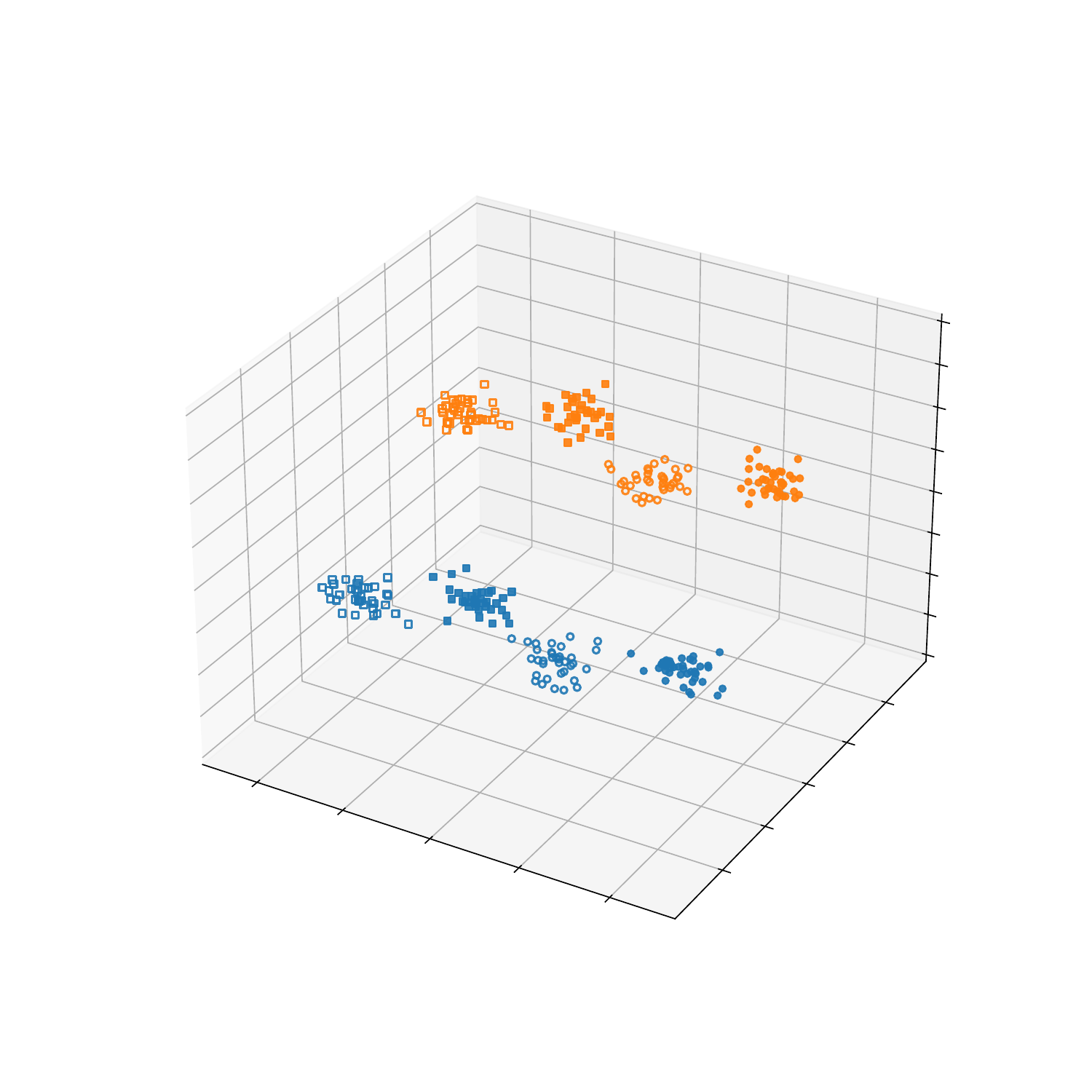}

    \parbox[c]{0.45\linewidth}{\centering ENS-t-SNE}
    \parbox[c]{0.45\linewidth}{\centering MPSE}    
    
    \includegraphics[width=0.32\linewidth]{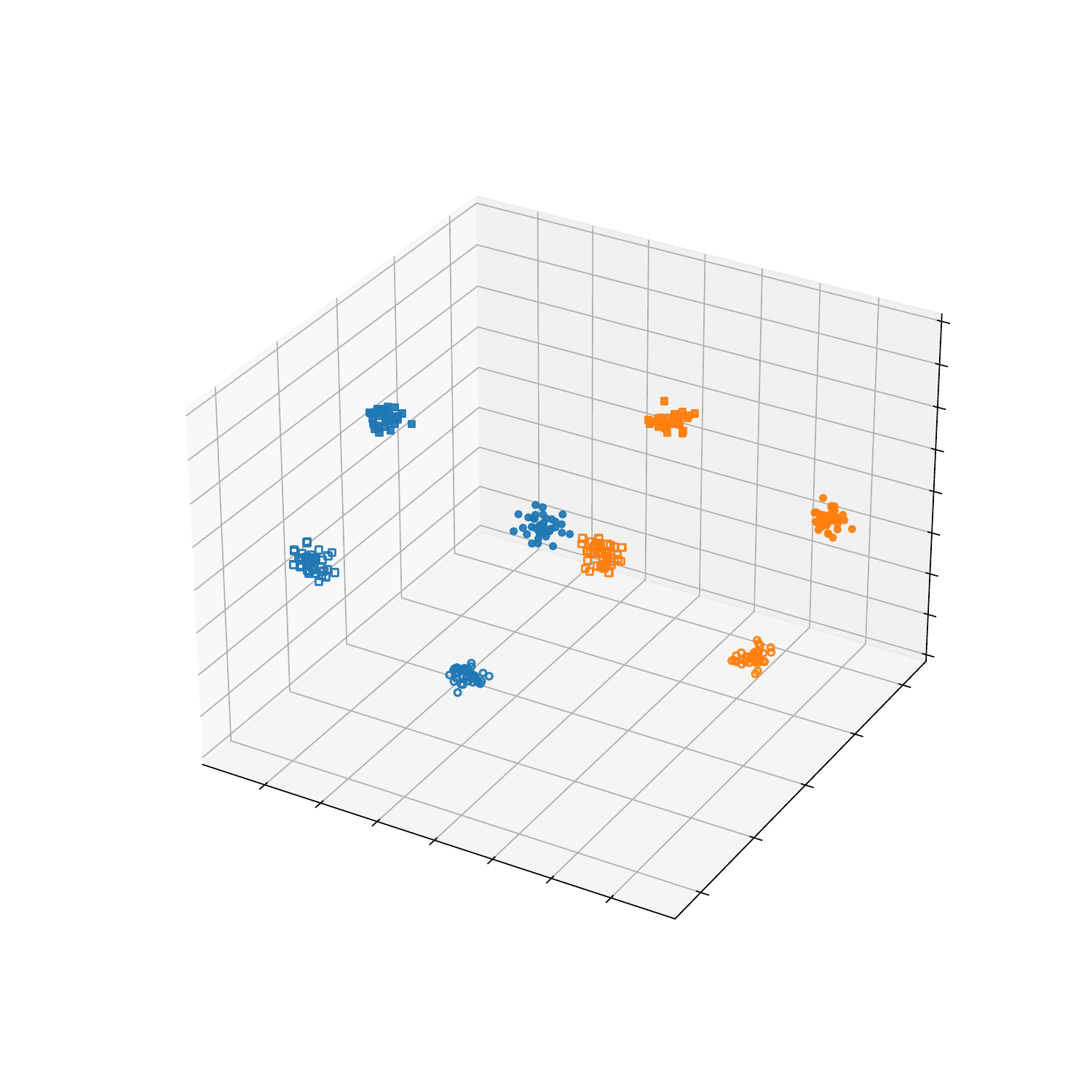}
    \includegraphics[width=0.32\linewidth]{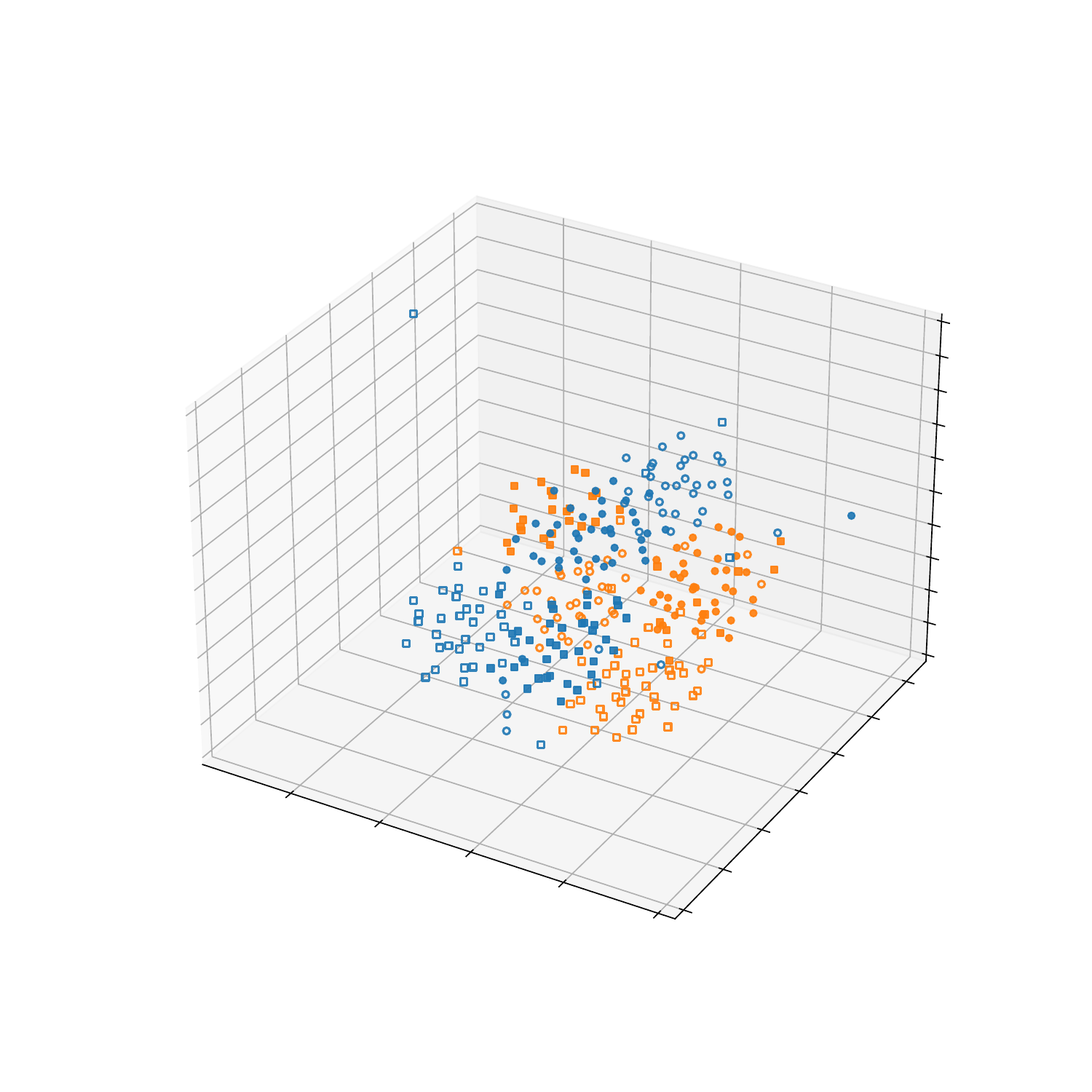}
    \includegraphics[width=0.32\linewidth]{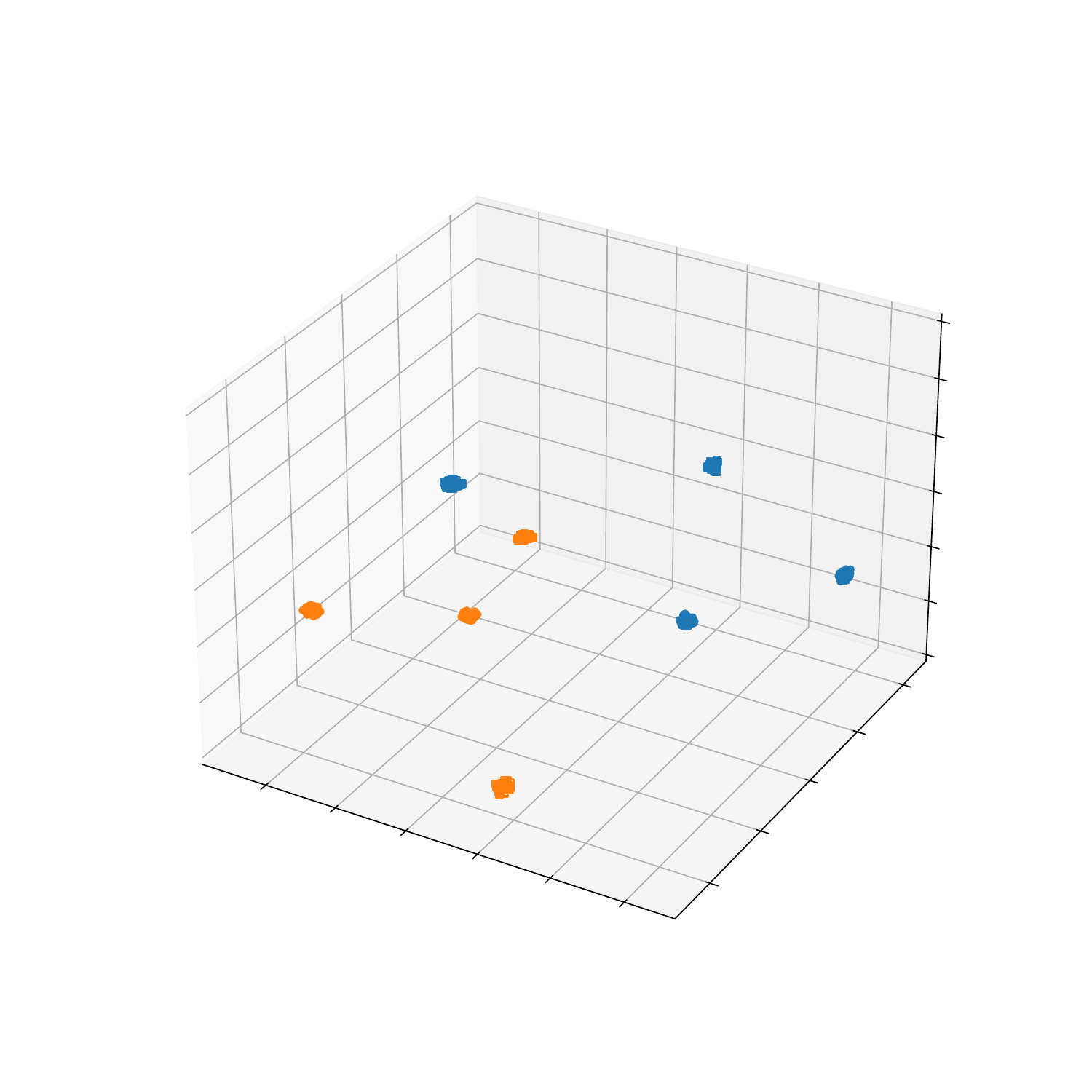}    

    \parbox[c]{0.32\linewidth}{\centering MDS}
    \parbox[c]{0.32\linewidth}{\centering t-SNE}
    \parbox[c]{0.32\linewidth}{\centering UMAP}    
    
    \caption{Synthetic data generated for the quantitative experiment. The two clusters in the first 10 dimensions are separated using color, the next 10 dimensions are denoted by shape, and the final 10 denoted by fill. }
    \label{fig:synthetic-quant-exp}
\end{figure}

\begin{figure}
    \centering
    \includegraphics[width=0.32\linewidth]{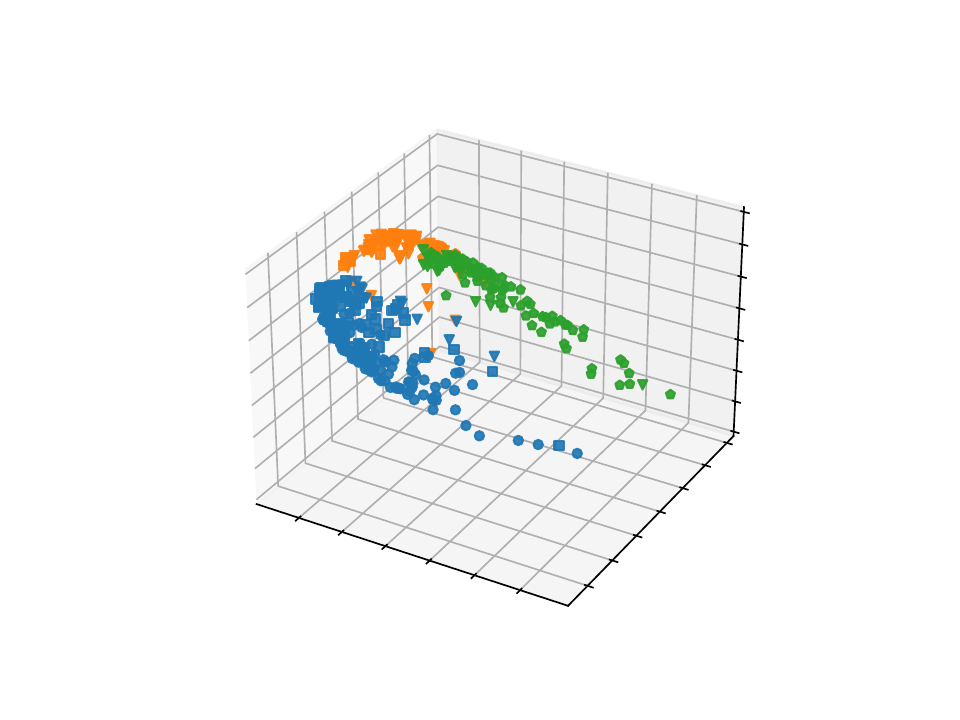}
    \includegraphics[width=0.32\linewidth]{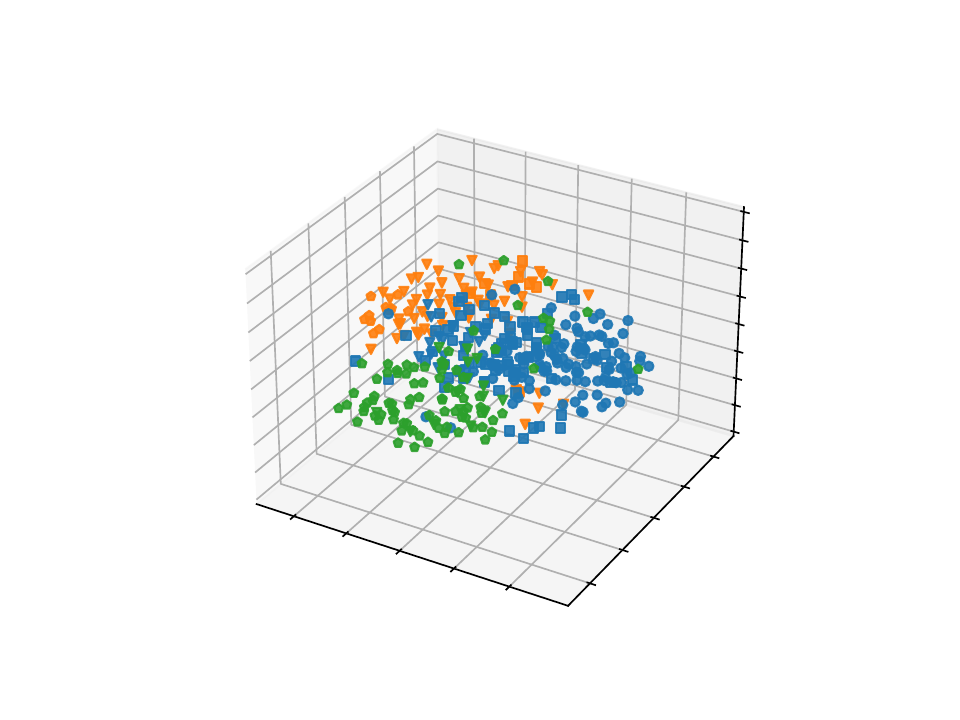}
    \includegraphics[width=0.32\linewidth]{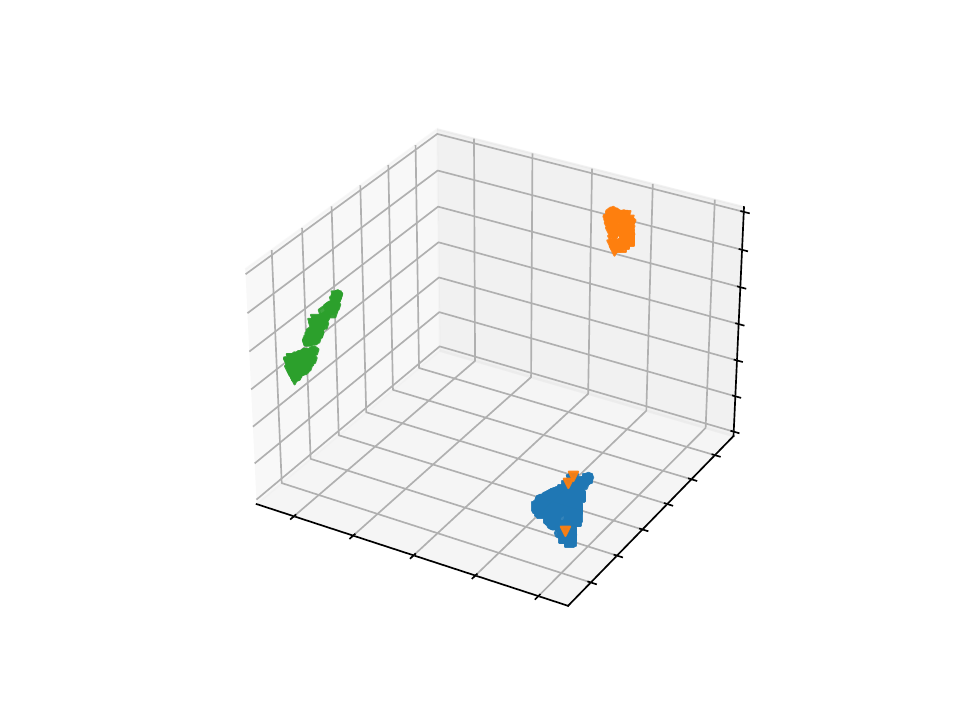}    
    \caption{Auto-MPG data embedded by MDS, t-SNE, and UMAP.}
    \label{fig:auto-other}
\end{figure}

\begin{figure}
    \centering
    \includegraphics[width=0.32\linewidth]{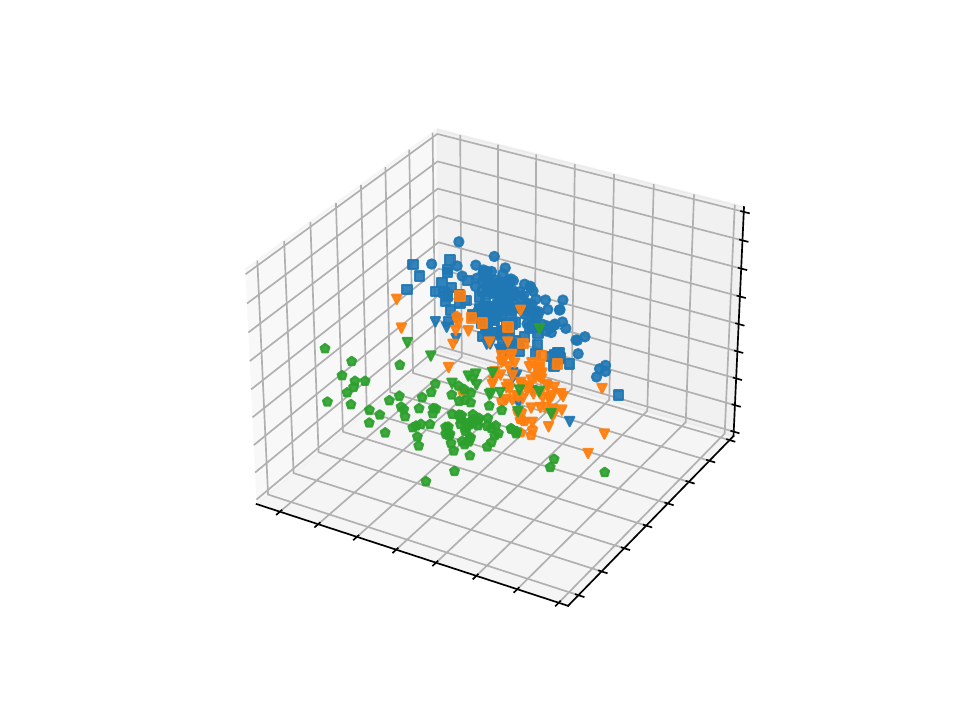}
    \includegraphics[width=0.67\linewidth,height=5cm]{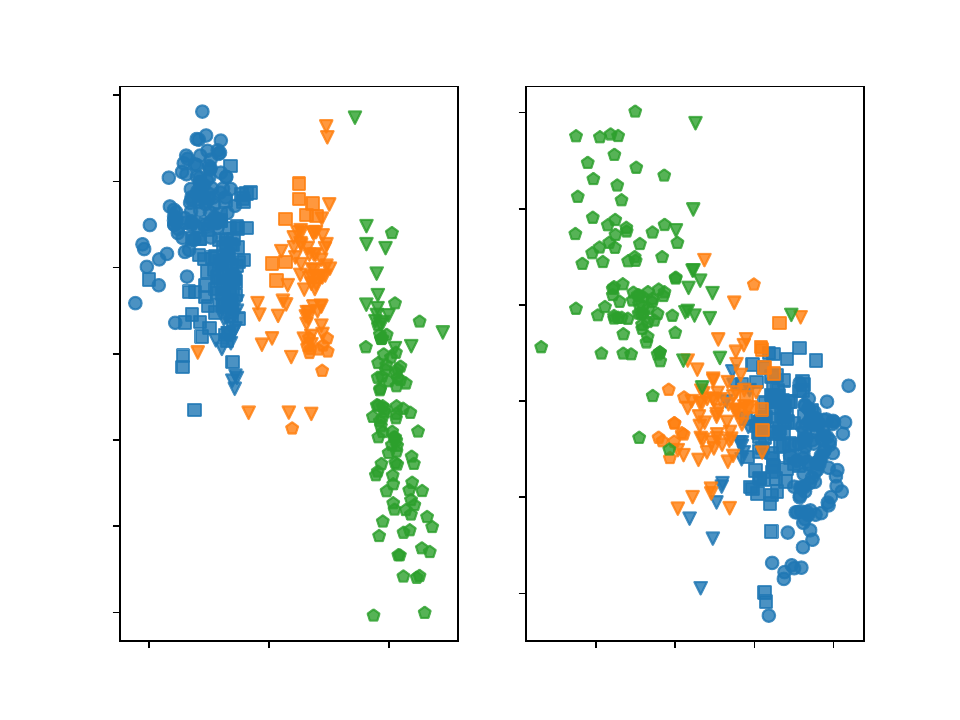}
    \caption{Auto-MPG data embedded by MPSE.}
    \label{fig:auto-mpse}
\end{figure}

\begin{figure}
    \centering
    \includegraphics[width=0.32\linewidth]{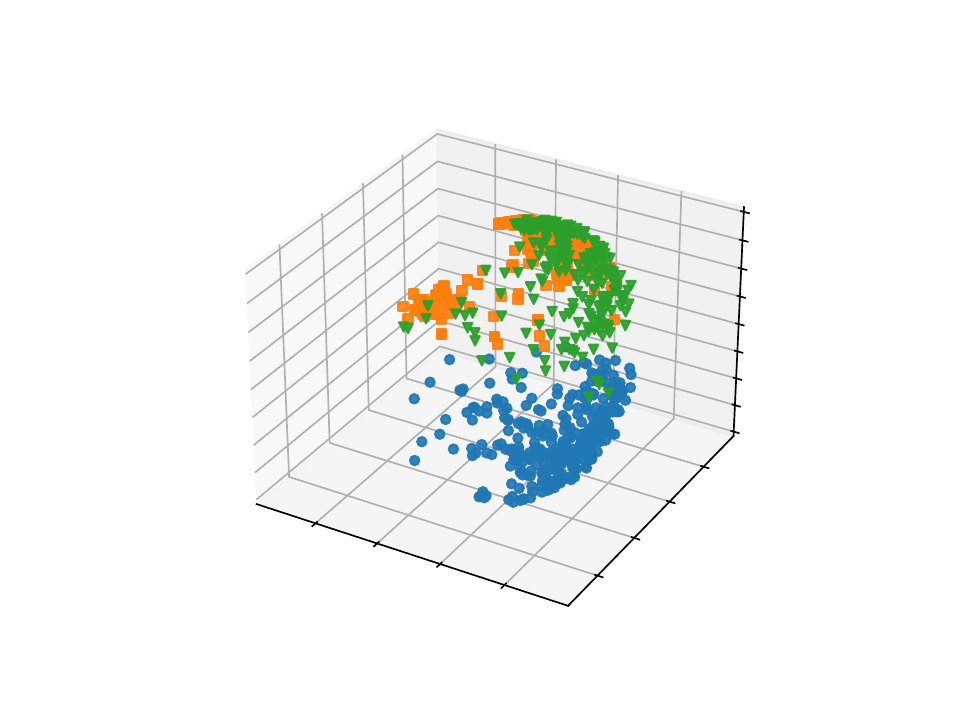}
    \includegraphics[width=0.32\linewidth]{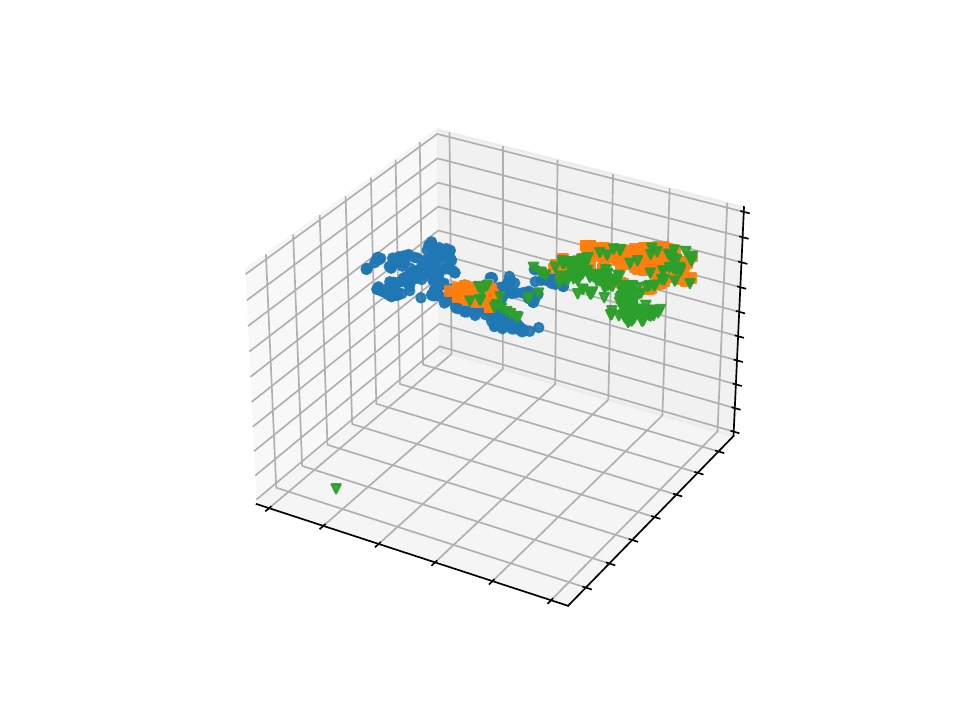}
    \includegraphics[width=0.32\linewidth]{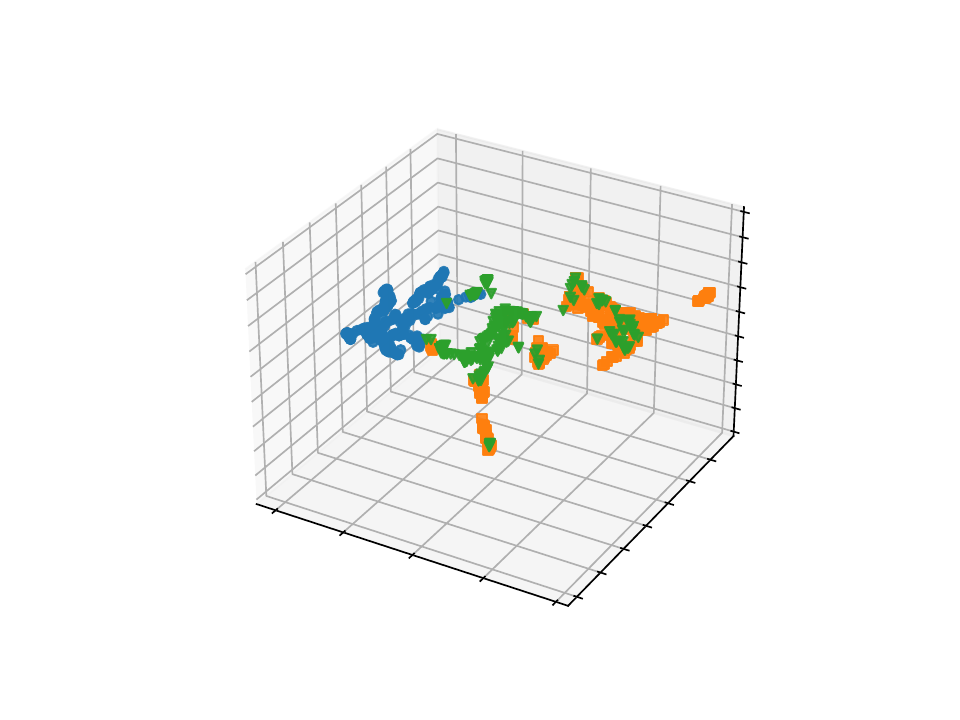}    
    \caption{Food Composition data embedded by MDS, t-SNE, and UMAP.}
    \label{fig:food-other}
\end{figure}

\begin{figure}
    \centering
    \includegraphics[width=0.32\linewidth]{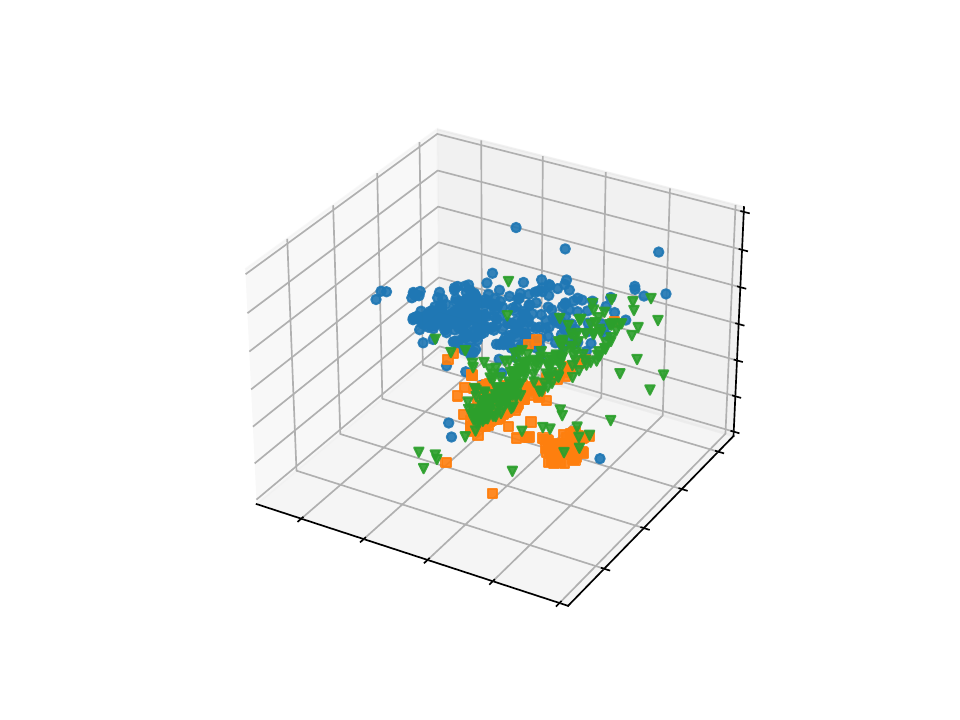}
    \includegraphics[width=0.67\linewidth, height=5cm]{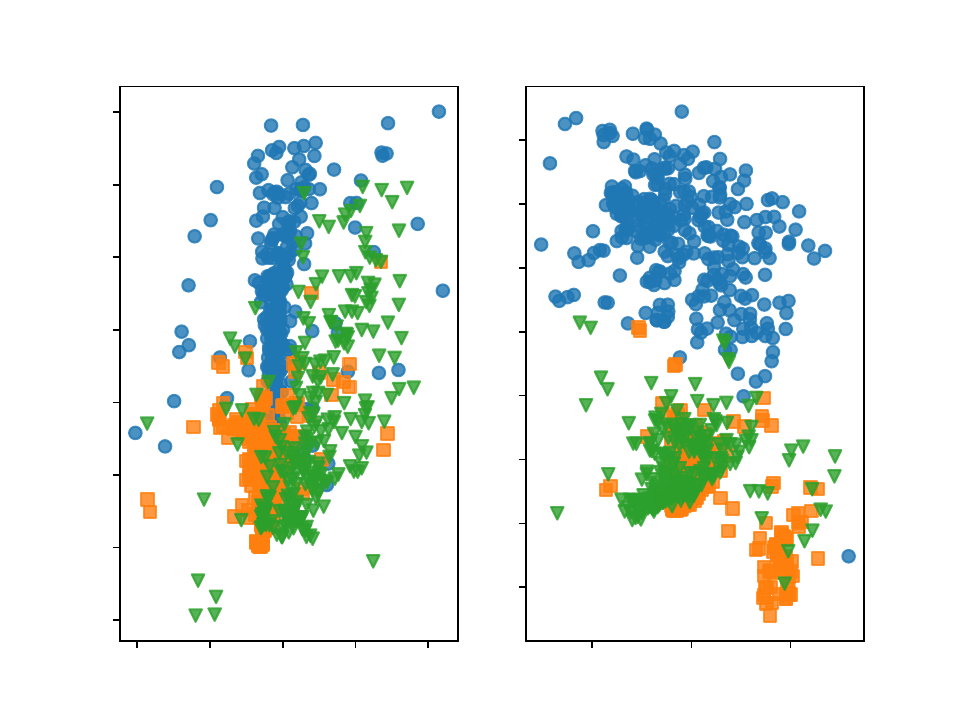}
    \caption{Food Composition data embedded by MPSE.}
    \label{fig:food-mpse}
\end{figure}

\end{document}